\documentclass{article}

 \usepackage[preprint,nonatbib]{neurips_2025}

\usepackage[utf8]{inputenc} % allow utf-8 input
\usepackage[T1]{fontenc}    % use 8-bit T1 fonts
\usepackage{hyperref}       % hyperlinks
\usepackage{url}            % simple URL typesetting
\usepackage{booktabs}       % professional-quality tables
\usepackage{tabularx}
\usepackage{caption}
\usepackage{amsfonts}       % blackboard math symbols
\usepackage{subfig}         % for \subfloat
\usepackage{wrapfig}
\usepackage{nicefrac}       % compact symbols for 1/2, etc.
\usepackage{microtype}      % microtypography

\usepackage{booktabs}
\usepackage{multirow}
\usepackage[table,xcdraw]{xcolor}
\usepackage{amsmath}
\usepackage{graphicx}

\usepackage{enumitem}
\usepackage{bbding}         % check and cross mark

\usepackage{makecell}

\usepackage[most]{tcolorbox}
\usepackage{listings}
\tcbset{
prompt_box_style/.style={
    enhanced,
    colback=white,
    colframe=black,
    colbacktitle=gray!20,
    coltitle=black,
    rounded corners,
    sharp corners=north,
    boxrule=0.5pt,
    drop shadow=black!50!white,
    breakable, % Support text across pages
    attach boxed title to top left={
        xshift=-2mm,
        yshift=-2mm
    },
    boxed title style={
        rounded corners,
        size=small,
        colback=gray!20
    }
},
replystyleg/.style={
    enhanced,
    colback=green!15,
    colframe=black,
    colbacktitle=green!30,
    coltitle=black,
    boxrule=0.5pt,
    drop shadow=black!50!white,
    rounded corners,
    sharp corners=north,
    attach boxed title to top right={
        xshift=-2mm,
        yshift=-2mm
    },
    boxed title style={
        rounded corners,
        size=small,
        colback=green!40
    }
},
replystyler/.style={
    enhanced,
    colback=red!15,
    colframe=black,
    colbacktitle=red!40,
    coltitle=black,
    boxrule=0.5pt,
    drop shadow=black!50!white,
    rounded corners,
    sharp corners=north,
    attach boxed title to top right={
        xshift=-2mm,
        yshift=-2mm
    },
    boxed title style={
        rounded corners,
        size=small,
        colback=red!40
    }
}
}
\newtcolorbox{promptbox}[1][]{
promptstyle,
title=Prompt,
#1
}

\newcommand*\LifelongAgentBench{\textsf{LifelongAgentBench} }
\newcommand*\LifelongAgentBenchNoSpace{\textsf{LifelongAgentBench}}

\title{\centering 
 \includegraphics[scale=0.04, bb=200 100 600 300]{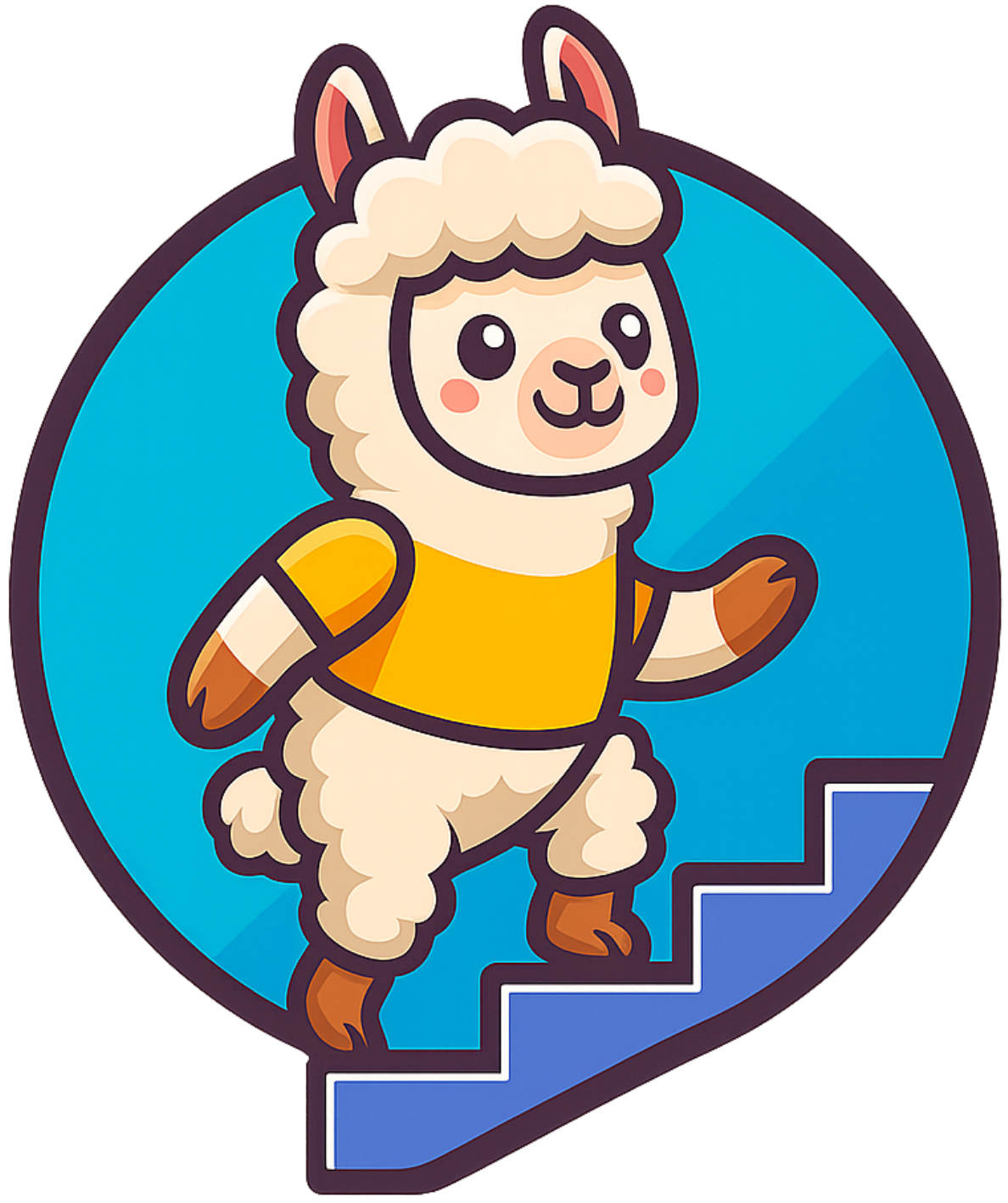} \textsf{LifelongAgentBench}: \\Evaluating LLM Agents as Lifelong Learners}

\author{Junhao~Zheng$^{1}$\thanks{JZ and XC are lead authors that contributed equally. (Email: junhaozheng47@outlook.com, xidicai067@gmail.com)},  Xidi~Cai$^{1*}$,  Qiuke~Li$^{1}$, Duzhen~Zhang$^{2}$, Zhong-Zhi~Li$^{3}$, Yingying~Zhang$^4$, \\\textbf{Le~Song}$^{2}$, \textbf{Qianli~Ma}$^{1}$\thanks{Correspondence author  (Email: qianlima@scut.edu.cn).} \\
$^1$ South China University of Technology, 
$^2$ MBZUAI, 
$^3$ Chinese Academy of Sciences, \\
$^4$ East China Normal University\\
\vspace{-0.5cm}
{\includegraphics[width=0.025\linewidth]{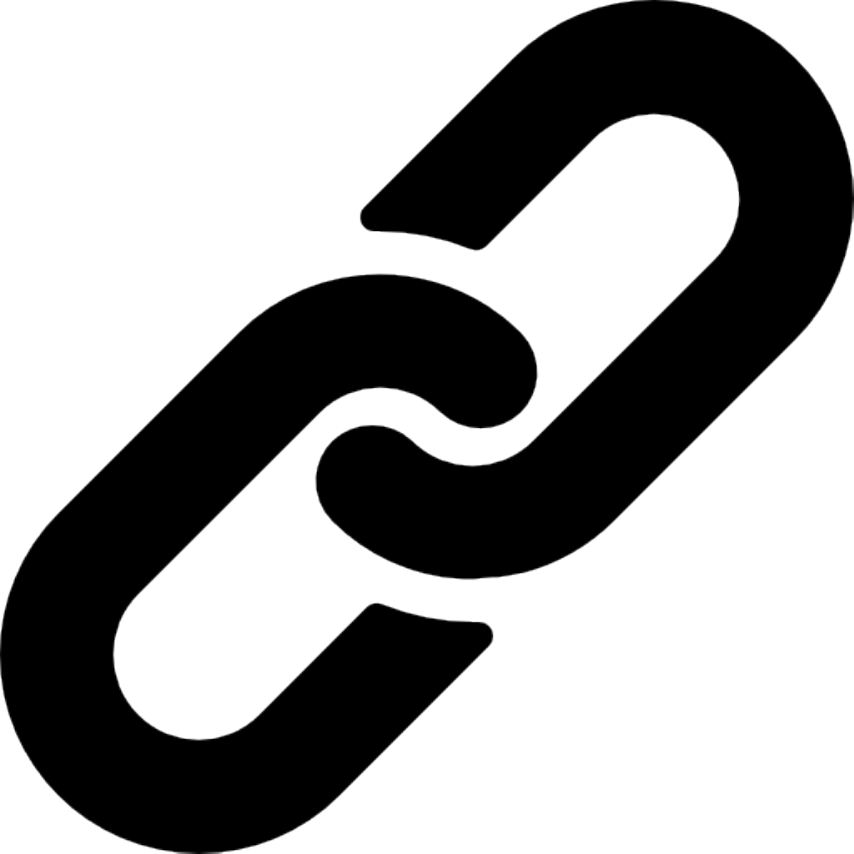} \href{https://caixd-220529.github.io/LifelongAgentBench/}{Project Page}}\quad
{\includegraphics[width=0.033\linewidth]{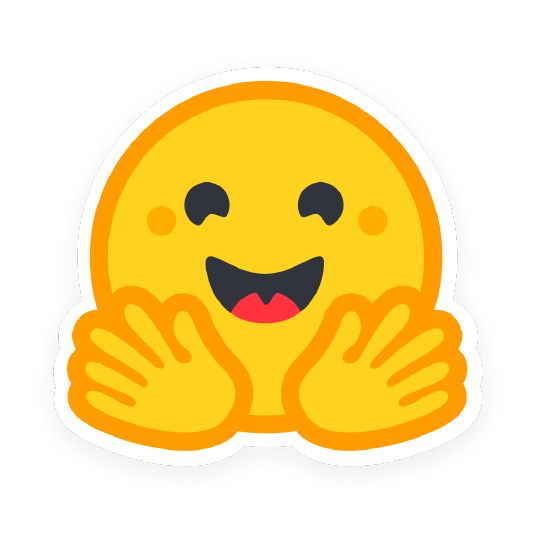} \href{https://huggingface.co/datasets/csyq/LifelongAgentBench}{Datasets}}\quad
{\includegraphics[width=0.025\linewidth]{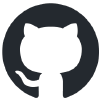} \href{https://github.com/caixd-220529/LifelongAgentBench}{Source Code}}
}

\begin{document}

\maketitle

\begin{abstract}

Lifelong learning is essential for intelligent agents operating in dynamic environments. Current large language model (LLM)-based agents, however, remain stateless and unable to accumulate or transfer knowledge over time. Existing benchmarks treat agents as static systems and fail to evaluate lifelong learning capabilities. We present \LifelongAgentBenchNoSpace, the \textit{first} unified benchmark designed to systematically assess the lifelong learning ability of LLM agents. It provides skill-grounded, interdependent tasks across three interactive environments—Database, Operating System, and Knowledge Graph—with automatic label verification, reproducibility, and modular extensibility. Extensive experiments reveal that conventional experience replay has limited effectiveness for LLM agents due to irrelevant information and context length constraints. We further introduce a \textit{group self-consistency} mechanism that significantly improves lifelong learning performance. We hope \LifelongAgentBench will advance the development of adaptive, memory-capable LLM agents.
% \footnote{Source code will be publicly available: \url{https://anonymous.4open.science/r/continual_agent_bench-4F3B}}
% https://github.com/caixd-220529/continual_agent_bench/tree/review

\end{abstract}

\addtocontents{toc}{\protect\setcounter{tocdepth}{0}}
\section{Introduction}
\label{sec:introduction}

The rapid development of large language models (LLMs) has revolutionized language-based artificial intelligence, achieving state-of-the-art performance across a wide range of natural language processing tasks. Recently, research has shifted from static models to LLM-based agents, designed to interact with dynamic environments, perform complex decision-making, and continuously improve through experience. These agents combine the language understanding and generation capabilities of pretrained LLMs with autonomous action selection and interaction policies.

However, a critical limitation remains: today’s LLM-based agents fundamentally lack memory and the ability to incrementally accumulate knowledge over time. They operate in a \textbf{stateless manner}, treating each task independently without the capacity to remember, adapt, or transfer past experiences. Achieving \textit{general artificial intelligence} demands agents that can \textbf{continuously acquire, retain, and reuse knowledge across diverse environments and long time horizons}. This lifelong learning capability is widely regarded as a cornerstone of human-level intelligence but remains largely unaddressed in current agent research \cite{zheng2025towards,Zheng2025Roadmap,li2025system}.

Existing LLM agent benchmarks \cite{zhouwebarena,Koh2024VisualWebArena,Liu2023AgentBench} have been designed under the static agent paradigm. They focus on isolated tasks, ignoring inter-task dependencies, skill reuse, and the realistic challenges of knowledge retention and catastrophic forgetting. More critically, there is currently \textbf{no standardized benchmark for systematically evaluating lifelong learning in LLM agents}. This absence has severely limited progress toward developing agents capable of lifelong adaptation and memory. Additionally, practical adoption is hindered by label inaccuracies \cite{Yang2024AgentOccam}, lack of verifiability, and poor reproducibility in prior benchmarks.

\begin{figure}[!t]
\centering
\includegraphics[width=0.95\linewidth]{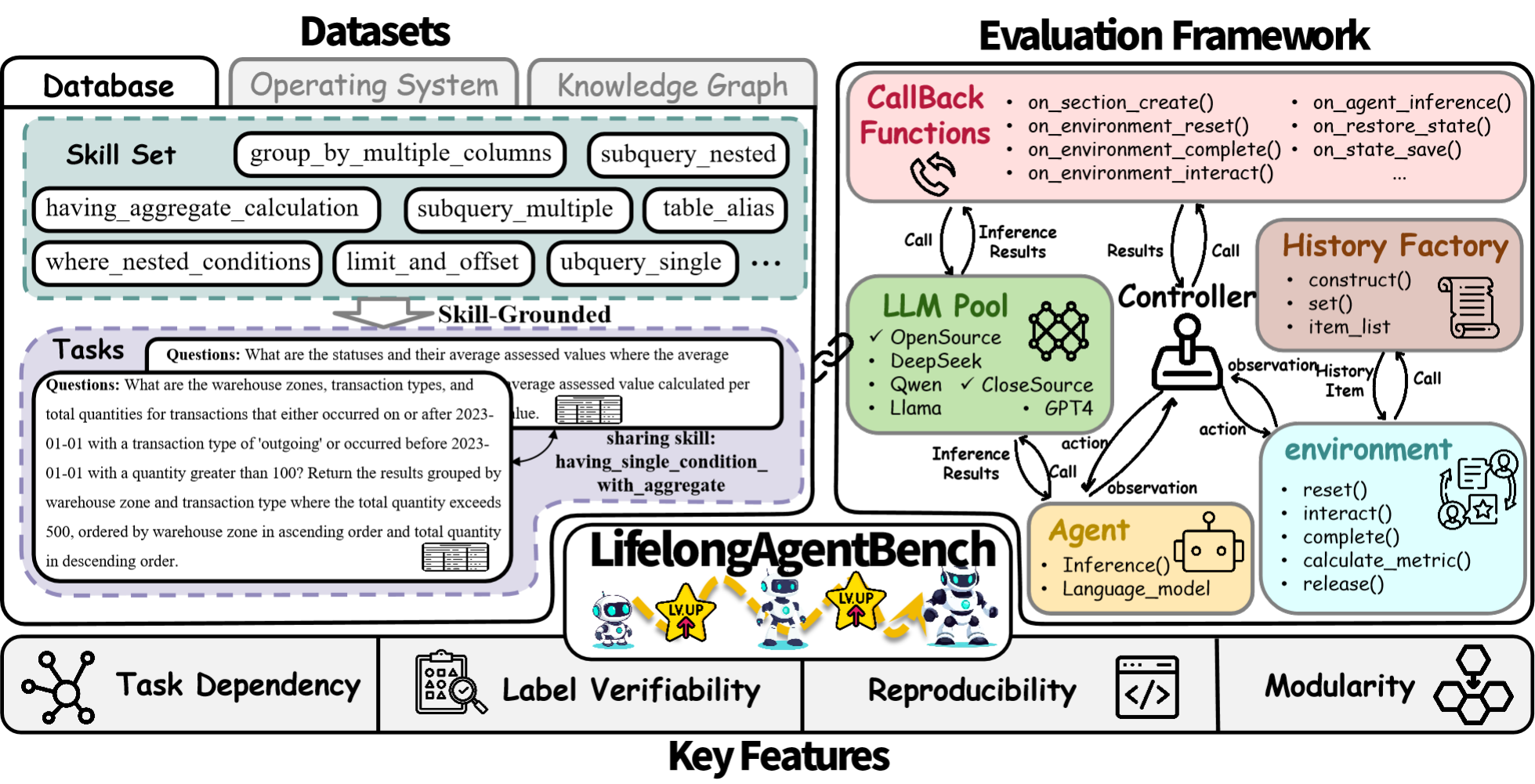}
\caption{Overview of \LifelongAgentBenchNoSpace: a unified dataset and evaluation framework with skill-grounded tasks, modular components, and four key properties: task dependency, label verifiability, reproducibility, and modularity.}
\label{fig:overview}
\end{figure}

To address these critical gaps, we propose \LifelongAgentBenchNoSpace, the first unified benchmark specifically designed to evaluate the lifelong learning capabilities of LLM-based agents across realistic and diverse interactive environments (Figure~\ref{fig:overview}). \LifelongAgentBench systematically tests agents’ abilities to acquire \emph{atomic skills}, transfer them across tasks, and maintain stable performance over long sequences of dependent tasks. It includes three task-rich environments---Database (DB), Operating System (OS), and Knowledge Graph (KG)---to simulate complex, evolving scenarios requiring continuous learning.

\LifelongAgentBench provides four key innovations that set it apart from existing LLM agent benchmarks:
\textbf{Task Dependency:} Tasks are skill-grounded and explicitly designed to quantify inter-task relatedness, enabling rigorous analysis of knowledge transfer and catastrophic forgetting.
\textbf{Label Verifiability:} Each environment includes automatic label verification (e.g., SQL query validation, OS state hashing, SPARQL output verification) to ensure objective and reproducible evaluation.
\textbf{Reproducibility:} The benchmark provides a fully containerized infrastructure and modular design, making it straightforward for researchers to reproduce experiments and extend the framework.
\textbf{Modularity:} The platform offers extensible callback functions and a pluggable LLM agent interface, supporting both open-source and commercial models such as LLaMA \cite{grattafiori2024llama}, DeepSeek \cite{guo2025deepseek}, Qwen \cite{yang2024qwen2}, and GPT-4 \cite{achiam2023gpt}. A detailed comparison between \LifelongAgentBench and prior benchmarks is presented in Table~\ref{tab:benchmark_comparison}.

\begin{table}[!t]
  \centering
  \caption{Comparison between \LifelongAgentBench and existing benchmarks. $\dagger$: \cite{Yang2024AgentOccam} highlights label error issues in WebArena.}
  \resizebox{\linewidth}{!}{
    \begin{tabular}{l|c|c|c|c|c}
        \toprule
    \textbf{Benchmark} & WebArena $\cite{zhouwebarena}$ & VisualWebArena $\cite{Koh2024VisualWebArena}$ & AgentBench $\cite{Liu2023AgentBench}$ & VisualAgentBench $\cite{liu2024visualagentbench}$ & \LifelongAgentBench (Ours) \\
    \midrule
    \textbf{Task Execution Scheme} & \multicolumn{4}{c|}{Parallel execution with undetermined execution order} & Serial execution with fixed order and historical dependency retention \\
    \rowcolor[rgb]{ .949,  .949,  .949} \textbf{Task Dependency} & \multicolumn{4}{c|}{\XSolidBrush} & \Checkmark \\
    \textbf{Knowledge Transfer} & \multicolumn{4}{c|}{\XSolidBrush} & \Checkmark \\
    \rowcolor[rgb]{ .949,  .949,  .949} \textbf{Modular Extension} & \multicolumn{4}{c|}{\XSolidBrush}       & \Checkmark \\
    \midrule
    \textbf{Supported Environment} & \multicolumn{2}{c|}{Web Only} & \multicolumn{2}{c|}{Environment with serializable action and observation} & Environment with serializable action and observation \\
    \rowcolor[rgb]{ .949,  .949,  .949} \textbf{Label Verification Mechanism} & \multicolumn{2}{c|}{Human Annotation $\dagger$} & \multicolumn{2}{c|}{Automatic Label Verification} & Automatic label verification via human review, LLM judgement, and execution results \\
    \textbf{Deployment} & \multicolumn{2}{c|}{Single Machine Deployment} & \multicolumn{2}{c|}{Distributed Deployment} & Both Single Machine and Distributed Deployment \\
    \rowcolor[rgb]{ .949,  .949,  .949} \textbf{Transparent RPC} & \multicolumn{2}{c|}{/} & \multicolumn{2}{c|}{\XSolidBrush} & \Checkmark \\
    \textbf{Code Quality Control} & \multicolumn{2}{c|}{Not declared} & \multicolumn{2}{c|}{Code are partially check by black, mypy, beartype.} & All code are checked by black and mypy (strict). \\
    \midrule
    \rowcolor[rgb]{ .949,  .949,  .949} \# \textbf{Instance} & 812   & 910   & 1091  & 746   & 1396 \\
    \bottomrule
    \end{tabular}%
    }
  \label{tab:benchmark_comparison}%
\end{table}%

We conduct extensive experiments using \LifelongAgentBenchNoSpace, yielding several key insights:
(1) While experience replay is effective in traditional continual learning, its impact in agent settings varies significantly depending on model size, architecture, and task complexity.
(2) Increasing the volume of past experience does not always improve performance and can even degrade it due to irrelevant information and context length limitations.
(3) To mitigate this, we propose a novel \textit{group self-consistency} mechanism, which partitions historical experience into groups and applies voting strategies to improve decision quality. We show that group self-consistency significantly enhances the effectiveness of experience replay across multiple model backbones.

Our contributions are threefold:
(1) We introduce \LifelongAgentBenchNoSpace, the first unified benchmark specifically designed to evaluate the lifelong learning capabilities of LLM-based agents across diverse, realistic interactive environments.
(2) We provide the first systematic analysis of lifelong learning in LLM agents, revealing key limitations of conventional experience replay due to irrelevant information and context length constraints.
(3) We propose a novel \textit{group self-consistency} mechanism that partitions historical experiences and applies voting strategies, significantly enhancing the effectiveness of lifelong learning across multiple LLM backbones.

\section{Related Work}
\label{sec:related}

\textbf{Lifelong Learning.}\quad Lifelong learning, or continual learning, aims to enable AI systems to acquire and retain knowledge across sequential tasks while mitigating catastrophic forgetting \cite{french1999catastrophic}. Prior research has primarily focused on static, non-interactive settings such as image classification or continual instruction tuning, where models are fine-tuned on sequential datasets without interacting with an external environment \cite{Zheng2025Roadmap, zheng2025towards}. In these tasks, both inputs and outputs are fixed, and the model is not required to actively take actions or adapt based on environmental feedback. In contrast, lifelong learning for LLM-based agents interacting with complex environments over long horizons remains largely unexplored.

\textbf{LLM Agent Benchmarks.}\quad Several benchmarks have been proposed to evaluate the capabilities of LLM-based agents. WebArena \cite{zhouwebarena}, AgentBench \cite{Liu2023AgentBench}, and VisualWebArena \cite{Koh2024VisualWebArena} offer valuable evaluation settings but focus on single-episode performance in static environments. These platforms lack mechanisms to model sequential decision making, cumulative learning, or skill transfer across tasks. While recent works have explored LLM agents in interactive scenarios such as game playing \cite{zhu2023minecraft} and tool use \cite{qin2023toolllm}, they do not provide standardized lifelong learning protocols. 

\LifelongAgentBench addresses this critical gap by providing the first benchmark specifically designed to evaluate LLM agents under lifelong learning constraints. It introduces reproducible lifelong evaluation with persistent environment states, explicit task dependencies via skill taxonomies, and scalable experience replay, establishing a foundation for systematic study of generalization, skill transfer, and long-term retention in LLM agents.

\section{Problem Formulation of Lifelong Learning for LLM Agents}
\label{sec:problem}

We model lifelong learning for LLM-based agents as sequential decision making over a series of tasks, each framed as a goal-conditioned partially observable Markov decision process (POMDP) \cite{Zheng2025Roadmap}.
\textbf{Environment:} An environment is $\mathcal{E} = (\mathcal{S}, \mathcal{A}, \mathcal{G}, T, R, \Omega, O)$, where $\mathcal{S}$ is the state space; $\mathcal{A}$, natural language actions; $\mathcal{G}$, task goals; $T$, state transitions; $R$, rewards; $\Omega$, observations; and $O$, the observation function. \LifelongAgentBench provides DB, OS, and KG environments.
\textbf{Agent and Task:} An LLM agent follows a policy $\pi: \Omega \rightarrow \mathcal{A}$ mapping observation $o_t$ to action $a_t$. A task is $\mathcal{T}^{(i)} = \langle \mathcal{E}^{(i)}, o_0^{(i)}, g^{(i)} \rangle$, with $o_0^{(i)}$ as the initial observation and $g^{(i)}$ as the goal. The agent generates a trajectory $\xi^{(i)} = (o_0, a_0, r_0, \dots, o_T, a_T, r_T)$, receiving a single reward upon submitting a final answer (success = 1, failure = 0).
\textbf{Objective:} Given tasks $\mathcal{U} = \{\mathcal{T}^{(1)}, \dots, \mathcal{T}^{(n)}\}$, the goal is to maximize cumulative expected reward:
\(
\max_{\pi} \sum_{i=1}^n \mathbb{E}_{\xi^{(i)} \sim \pi}\left[\sum_{t=0}^{T} R(o_t, a_t, g^{(i)})\right]
\)
\LifelongAgentBench evaluates agents on their ability to leverage past experience to improve current task performance.
\section{Data Construction}
\label{sec:data_construction}

To rigorously evaluate LLM agents in lifelong learning scenarios, we introduce a novel and meticulously constructed benchmark dataset composed of three distinct and challenging environments: Database, Operating System, and Knowledge Graph. Unlike conventional benchmarks that often rely on isolated, simplistic tasks with loosely defined inter-task relationships, our dataset is innovatively designed to reflect complex, realistic lifelong learning contexts. The key contributions of this dataset include the systematic generation of tasks explicitly tied to clearly defined \emph{atomic skills}, sophisticated methodologies for controlling \emph{skill distribution} and \emph{task complexity}, and rigorous noise management to simulate real-world variability. The dataset's construction required extensive validation and curation efforts, underscoring the intricacy and robustness of our approach. Detailed descriptions of the construction procedures and sample data are provided in Appendix \ref{sec:lifelong_learning_environment_design} and Appendix \ref{sec:appendix_dataset_samples}, respectively.

\subsection{Design Principles}

% \begin{figure}
\begin{wrapfigure}{r}{6cm}
    \includegraphics[width=\linewidth]{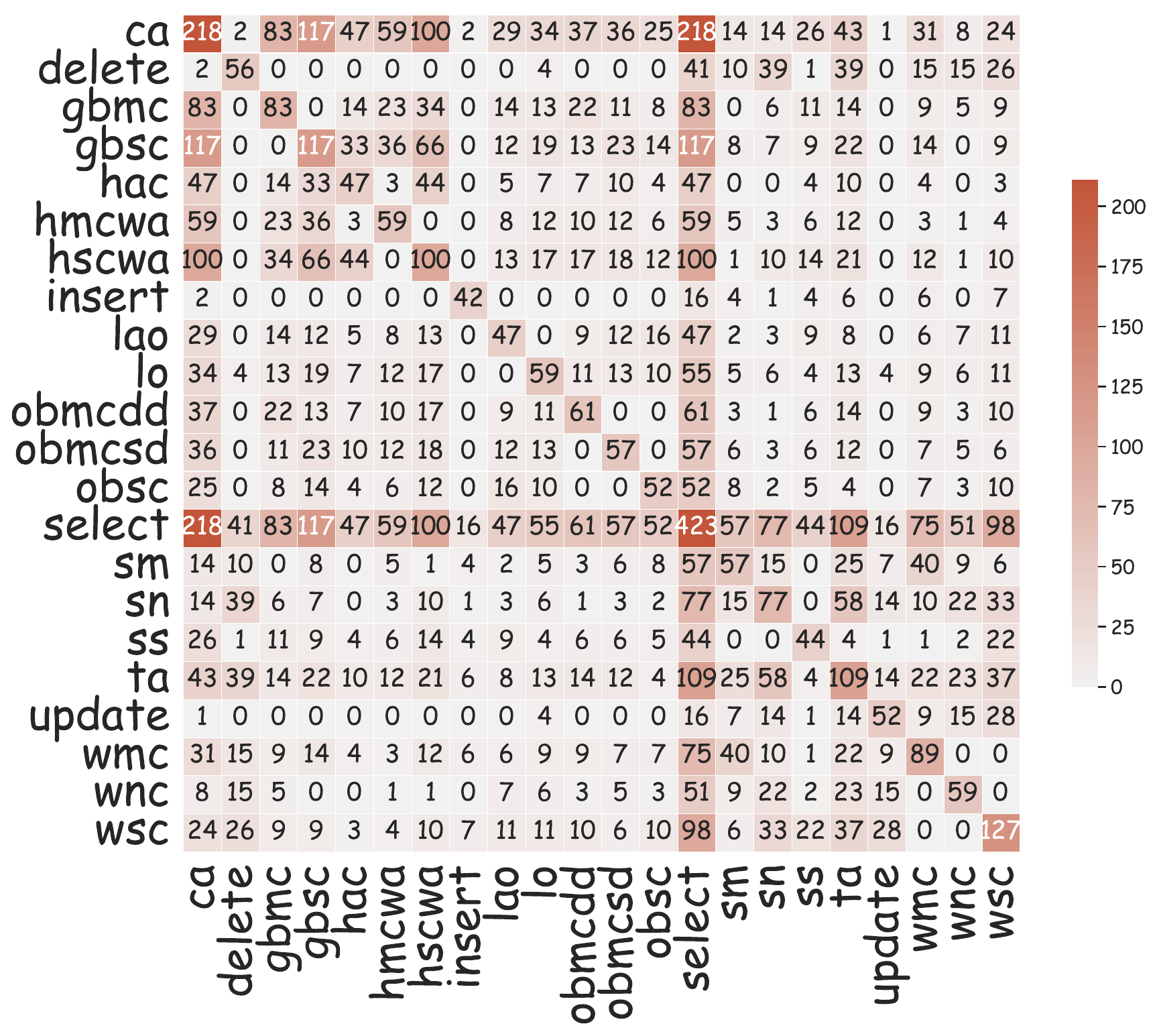}
    \caption{Skill concurrency in the Database environment.}
    \label{fig:skill_concurrency}
\end{wrapfigure}
% \end{figure}

The data construction process follows three core principles. First, we adopt a skill-centric task generation approach. Each environment $\mathcal{E}^{(i)}$ is characterized by a set of atomic skills $\mathcal{SK}_{\mathcal{E}^{(i)}}$, where the number of skills $N_{\mathcal{E}^{(i)}}$ varies with the environment's complexity. Each task $\mathcal{T}_j^{(i)}$ is associated with a subset of these skills $\mathcal{SK}_{\mathcal{E}^{(i)}}^{(j)}$, ensuring consistent competency representation across tasks. The relationship between tasks $m$ and $n$ is quantified by the harmonic mean of shared skill proportions:
\(
as_{\mathcal{E}^{(i)}}^{(m,n)} = 2 as_{\mathcal{E}^{(i)}}^{(m)}  as_{\mathcal{E}^{(i)}}^{(n)}/(as_{\mathcal{E}^{(i)}}^{(m)} + as_{\mathcal{E}^{(i)}}^{(n)})
\)
where $as^{(m)}$ and $as^{(n)}$ denote the proportion of shared skills relative to each task's total skills. This formulation captures both commonality and uniqueness across tasks.

To mitigate skill isolation, we employ a probabilistic sampling strategy where infrequent skills have higher sampling probabilities, ensuring balanced representation across the dataset. Noise levels are controlled by regulating the proportion of tasks containing rare skills, facilitating robustness analysis. Tasks span simple, intermediate, and complex configurations to mimic real-world variability and allow evaluation across progressive difficulty levels. As shown in Figure~\ref{fig:skill_concurrency}, extensive connections exist between skills across tasks. A summarization of the skill set in each enviroment is in Table~\ref{tab:skill_set}.

\subsection{Environment Implementations}
\label{sec:environment_implementation}

\textbf{Database Environment:} We implement this environment using Docker-containerized MySQL instances to ensure reproducibility. A fresh MySQL container is created for each experimental run to maintain task isolation. Tasks are initialized by generating a database table with predefined attributes, which is deleted upon task completion. We identify 22 SQL-related skills, including column aliasing, complex filtering with \texttt{WHERE} and \texttt{HAVING} clauses, multi-column grouping, data manipulation (\texttt{INSERT}, \texttt{UPDATE}, \texttt{DELETE}), and nested subqueries. Detailed description of each skill is provided in Appendix \ref{sec:lifelong_learning_environment_design:database:environment_introduction}.

Task construction begins by sampling skills, with infrequent skills prioritized. SQL queries corresponding to sampled skills are generated using the DeepSeek-R1 model. Each query is executed on a synthetic database instance, and invalid or inconsistent tasks are discarded. To prevent skill imbalance, we require a minimum of 20 occurrences per skill across 500 selected tasks. Task correctness is verified both automatically (e.g., result matching, MD5 hashing of database states) and manually by inspecting 10\% of randomly sampled tasks for syntax and logical coherence.

\textbf{Operating System Environment:} This environment leverages disposable Docker containers running Ubuntu to isolate tasks. Containers are destroyed and re-instantiated after each task. We define 29 Bash command skills, including file manipulation (\texttt{cp}, \texttt{mv}, \texttt{rm}), user management (\texttt{useradd}, \texttt{groupadd}), text processing (\texttt{awk}, \texttt{grep}, \texttt{sed}), and system monitoring (\texttt{ps}, \texttt{top}). Tasks are grouped by complexity: simple (1–4 commands), intermediate (5–8), and complex (9–12). Detailed description of each skill is provided in Appendix \ref{sec:lifelong_learning_environment_design:operating_system:environment_introduction}.

Command sequences are generated using DeepSeek-R1, ensuring logical consistency across multiple steps. Validation scripts automatically compare command outputs against expected results, with file changes verified via checksums. Preliminary experiments revealed that simpler tasks provided limited lifelong learning value; therefore, the final dataset focuses primarily on complex tasks to capture inter-skill dependencies.

\textbf{Knowledge Graph Environment:} This environment is based on a SPARQL query system. Tasks involve querying structured data through operations such as relation extraction and intersection. Tasks were curated from the GrailQA dataset \cite{gu2021beyond} by mapping S-expressions to logical action sequences. These sequences range from 2 to 9 steps to ensure uniform distribution across task lengths. Each query is validated on a synthetic knowledge graph to confirm result correctness. Complex queries (7–9 steps) received additional manual validation to ensure semantic accuracy. Detailed description of each skill is provided in Appendix \ref{sec:lifelong_learning_environment_design:knowledge_graph:environment_introduction}.

\subsection{Quality Control}

\textbf{Label Validation:} We employ automated validation mechanisms, including result comparison for SQL queries (Appendix \ref{sec:lifelong_learning_environment_design:database:data_construction}, Figure~\ref{fig:data_construction_procedure_db}), exit code checking for Bash commands (Appendix \ref{sec:lifelong_learning_environment_design:operating_system:data_construction}, Figure~\ref{fig:data_construction_procedure_os}), and output verification for SPARQL queries (Appendix \ref{sec:lifelong_learning_environment_design:knowledge_graph:data_construction}). Additionally, 10\% of tasks from each environment were manually reviewed for logical consistency and practical relevance. Pilot testing informed the final configuration to optimize task complexity and skill coverage. This multi-stage validation ensures that datasets are both challenging and representative of real-world scenarios.

\begin{figure}[!t]
    \centering
    \subfloat[Database\label{fig:skills_distribution_db_a}]{
        \raisebox{1.5mm}{\includegraphics[width=0.48\linewidth]{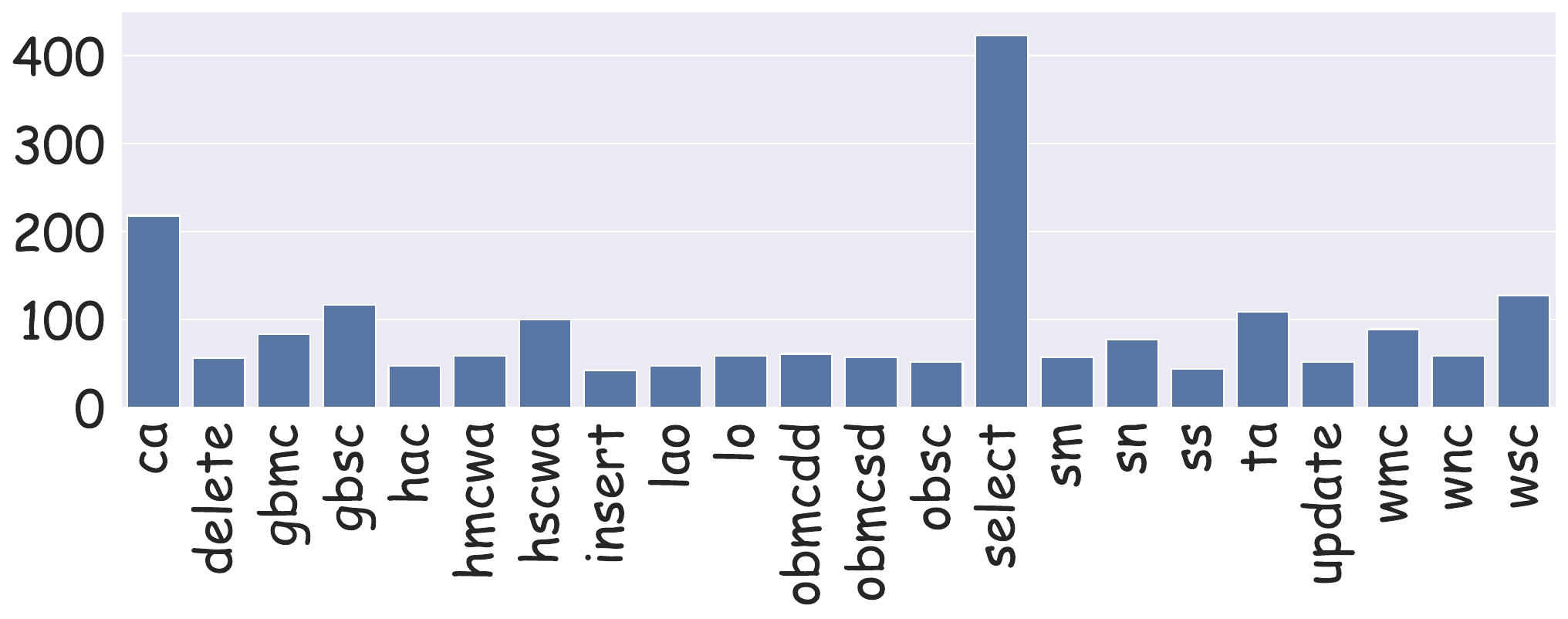}}
    }
    \hfill
    \subfloat[Operating System\label{fig:skills_distribution_db_b}]{
        \raisebox{0mm}{\includegraphics[width=0.48\linewidth]{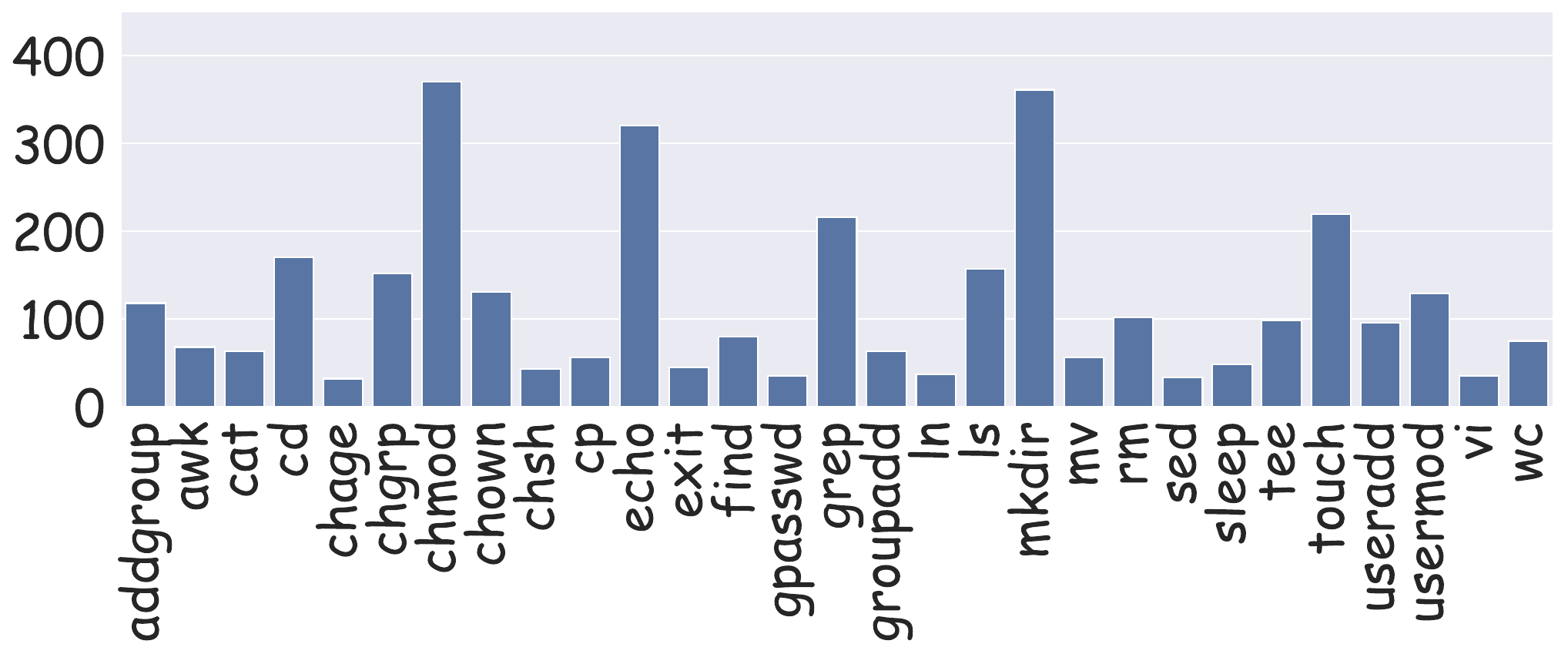}}
    }
    \caption{Skill distributions across tasks. Diverse and balanced skill coverage is maintained.}
    \label{fig:skill_distribution}
\end{figure}

\textbf{Balanced Skills:} In the Database environment, we generated 1,306 tasks with DeepSeek-R1 and selected 500 high-quality samples. Tasks cover 22 SQL skills with balanced distributions ensured by stratified sampling (Appendix \ref{sec:lifelong_learning_environment_design:database:data_construction}, Figure~\ref{fig:skills_distribution_db_a}). In the Operating System environment, 500 complex tasks were curated, with command sequences ranging from 9 to 12 steps to maximize inter-task skill overlap. Lower complexity tasks (1–8 steps) were excluded after preliminary tests showed minimal replay benefits (Figure~\ref{fig:skills_distribution_db_b}). In the Knowledge Graph environment, 396 tasks were extracted from GrailQA, mapped to atomic action sequences ranging from 2 to 9 steps. Replay effects were observed to diminish in sequences exceeding six steps.

\section{Evaluation Framework}
\label{sec:framework}

\textbf{LifelongAgentBench} is designed as a unified evaluation framework that integrates datasets and APIs to benchmark LLM-based agents under lifelong learning settings. In contrast to prior benchmarks that focus on static or single-task evaluation, our framework emphasizes sequential task execution and experience accumulation, offering a realistic simulation of continual learning scenarios. The detailed description is provided in Section~\ref{sec:appendix_evaluation_framework}.

\subsection{System Architecture}

The framework comprises six loosely coupled components: \textbf{model pool} (Appendix \ref{sec:appendix_evaluation_framework:framework_architecture:model_pool}), \textbf{agent} (\ref{sec:appendix_evaluation_framework:framework_architecture:agent}), \textbf{environment} (\ref{sec:appendix_evaluation_framework:framework_architecture:environment}), \textbf{chat history factory} (\ref{sec:appendix_evaluation_framework:framework_architecture:history_item_factory}), \textbf{controller} (\ref{sec:appendix_evaluation_framework:framework_architecture:controller}), and \textbf{callbacks} (\ref{sec:appendix_evaluation_framework:framework_architecture:callback}). Each component can be deployed independently across different servers and communicates via a custom remote procedure call (RPC) toolkit (\ref{sec:appendix_evaluation_framework:remote_procedure_call_toolkit}), enabling flexible distributed or local deployment (\ref{sec:appendix_evaluation_framework:distributed_deployment_framework}).

The \textbf{model pool} maintains mappings between model names and instances, supporting both open-source and proprietary LLM backends. The \textbf{agent} module translates environment observations and dialogue history into formatted inputs, queries the LLM, and parses outputs into executable actions. The \textbf{environment} component executes these actions and returns updated observations to the controller. It also implements standardized methods such as \texttt{reset}, \texttt{interact}, \texttt{complete}, \texttt{calculate\_metric}, and \texttt{release} to ensure consistency across environments.

The \textbf{controller} manages the interaction loop, oversees task scheduling, and relays agent actions to the environment. The \textbf{callback} system provides extensible hooks for monitoring internal events, facilitating reproducibility and experimental customization. 

\subsection{Reproducibility and Modularity}

Two core design principles of \textbf{LifelongAgentBench} are \textbf{reproducibility} and \textbf{modularity}. The framework guarantees deterministic behavior under fixed random seeds and uses containerized environment snapshots to ensure identical task conditions across experimental runs. Additionally, it exposes modular APIs for integrating new environments, task generators, custom agent architectures, or evaluation metrics with minimal engineering overhead. This flexibility allows researchers to experiment with various lifelong learning strategies while maintaining consistency and comparability.

\subsection{Differences from Prior Benchmarks}

Existing LLM agent benchmarks such as WebArena and AgentBench either rely on parallel task execution to reduce evaluation time or use process pools to manage multiple task sequences concurrently. These designs are incompatible with lifelong learning evaluation, where the strict order of task execution directly impacts the agent’s accumulated knowledge and performance.

In contrast, \LifelongAgentBench enforces strict sequential execution to preserve the integrity of experience accumulation and transfer learning assessments. Moreover, while prior frameworks tightly coupled agents, controllers, and environments into complex multi-process architectures, our design promotes developer-friendly single-process debugging with optional distributed scalability. This architecture substantially lowers the barrier to conducting lifelong learning research with LLM agents. The difference is summarized in Table \ref{tab:benchmark_comparison}.

\section{Evaluation of \LifelongAgentBench}
\label{sec:experiments}

We comprehensively evaluate the lifelong learning abilities of LLM-based agents using \LifelongAgentBench across three environments: Database (DB), Operating System (OS), and Knowledge Graph (KG). Our experiments systematically investigate experience replay and group self-consistency under unified protocols.

\subsection{Experimental Setup}

\textbf{Models:} We evaluate four LLM-based agents: Llama-3.1-8B-Instruct, Qwen2.5-7B-Instruct, DeepSeek-R1-Distill-Llama-8B, and DeepSeek-R1-Distill-Qwen-7B. All agents share a unified API with reproducible initialization, dynamic experience replay, and optional group self-consistency. 
\textbf{Baselines and Metric:} We evaluate agents under baseline (no replay), experience replay (1, 2, 4, 8, 16 prior trajectories), and experience replay with group self-consistency. Prior experiences are retrieved from recent \textbf{successful} trajectories. The evaluation metric is task success rate, defined as correct action sequences completing the task. 
\textbf{Environments:} Experiments run on Linux servers NVIDIA A800 (80GB). Code is based on Huggingface Transformers and PyTorch, with a distributed RPC-based framework for modular deployment. The system supports automatic checkpointing for recovery from interruptions.

\subsection{Main Results}

\begin{table}[!t]
\tiny
  \centering
  \caption{Main Result of \LifelongAgentBench. The backbone model is Llama-3.1-8B-Instruct. "Exp" represents that the number of recent \textbf{successful} trajectories that are provided to agent. The best result for each environment is bold. "OOM" represents out of memory.}
  \resizebox{0.8\linewidth}{!}{
    \begin{tabular}{ccccccc}
    \toprule
          & \textbf{Exp=0} & \textbf{Exp=1} & \textbf{Exp=4} & \textbf{Exp=16} & \textbf{Exp=32} & \textbf{Exp=64} \\
    \midrule
    \textbf{DB} & 0.19  & 0.41  & 0.73  & 0.75  & 0.77  & \textbf{0.78 } \\
    \rowcolor[rgb]{ .949,  .949,  .949} \textbf{OS} & 0.43  & 0.46  & \textbf{0.50 } & \textbf{0.50 } & 0.42  & 0.44  \\
    \textbf{KG} & 0.28  & \textbf{0.35 } & 0.33  &  \textcolor[rgb]{ .502,  .502,  .502}{OOM} & \textcolor[rgb]{ .502,  .502,  .502}{OOM} & \textcolor[rgb]{ .502,  .502,  .502}{OOM} \\
    \bottomrule
    \end{tabular}%
    }
  \label{tab:main_result}%
\end{table}%

\textbf{Strong performance of open-source models.} The results in Table~\ref{tab:main_result} demonstrate that the open-source Llama-3.1-8B-Instruct model achieves reasonable and stable performance across all environments. This contrasts with prior benchmarks such as AgentBench, where open-source LLMs often underperform, limiting academic reproducibility and accessibility.

\textbf{Experience replay consistently improves performance.} Incorporating past successful trajectories consistently boosts agent performance compared to the baseline (Exp=0). For DB, replay increases accuracy from 19\% to 78\% at 64 examples. For OS environment, accuracy improves from 43\% to 50\% at 4--16 examples. For KG, accuracy improves from 28\% to 35\% with just 1 example.

\textbf{Trade-off between replay benefits and memory limitations.} Increasing replay beyond optimal values leads to diminishing or negative returns due to excessive input length, increased reasoning complexity, and out-of-memory (OOM) failures. DB tasks, which involve shorter trajectories, benefit from large replay buffers. In contrast, OS and KG tasks, with multi-turn and long-form interactions, show peak performance at lower replay sizes before degradation or OOM occurs.

\textbf{Memory-efficient replay remains an open challenge.} These findings suggest that while experience replay is a valuable mechanism for improving LLM agent performance, it introduces significant memory and inference costs. Designing more efficient retrieval and summarization strategies for lifelong learning remains an important avenue for future research.

\subsection{Effect of Model Backbone and Task Difficulty}

\begin{table}[!t]
\tiny
  \centering
  \caption{The result when using different backbone LLM. The environment is DB. The best result for each backbone LLM is bold. "OOM" represents out of memory.}
  \resizebox{0.99\linewidth}{!}{
    \begin{tabular}{lcccccc}
    \toprule
    \multicolumn{1}{c}{\textbf{Model}} & \textbf{Exp=0} & \textbf{Exp=1} & \textbf{Exp=4} & \textbf{Exp=16} & \textbf{Exp=32} & \textbf{Exp=64} \\
    \midrule
    DeepSeek-R1-Distill-Llama-8B & 0.07  & \textbf{0.13 } & 0.35  & \textcolor[rgb]{ .502,  .502,  .502}{OOM} & \textcolor[rgb]{ .502,  .502,  .502}{OOM} & \textcolor[rgb]{ .502,  .502,  .502}{OOM} \\
    \rowcolor[rgb]{ .949,  .949,  .949} DeepSeek-R1-Distill-Qwen-7B & 0.10  & 0.12  & \textbf{0.18 } & \textcolor[rgb]{ .502,  .502,  .502}{OOM} & \textcolor[rgb]{ .502,  .502,  .502}{OOM} & \textcolor[rgb]{ .502,  .502,  .502}{OOM} \\
    QwQ-32B & \textbf{0.29 } & 0.23  & 0.21  & 0.25  & 0.31  & \textcolor[rgb]{ .502,  .502,  .502}{OOM} \\
    \rowcolor[rgb]{ .949,  .949,  .949} Qwen2.5-7B-Instruct & 0.74  & 0.71  & \textbf{0.76 } & 0.74  & \textcolor[rgb]{ .502,  .502,  .502}{OOM} & \textcolor[rgb]{ .502,  .502,  .502}{OOM} \\
    Qwen2.5-32B-Instruct & \textbf{0.82 } & 0.77  & 0.71  & 0.72  & 0.74  & \textcolor[rgb]{ .502,  .502,  .502}{OOM} \\
    \rowcolor[rgb]{ .949,  .949,  .949} Llama-3.1-8B-Instruct & 0.19  & 0.41  & 0.73  & 0.75  & 0.77  & \textbf{0.78 } \\
    Llama-3.1-70B-Instruct & 0.81  & 0.83  & 0.86  & 0.88  & 0.88  & \textbf{0.90 } \\
    \bottomrule
    \end{tabular}%
    }
  \label{tab:backbone_result}%
\end{table}%

\textbf{Backbone LLMs.} We evaluate a range of open-source LLMs across different architectures and model scales. The results in Table~\ref{tab:backbone_result} reveal notable differences in how backbone choice influences the effectiveness of experience replay. For strong base models such as Qwen2.5-7B-Instruct and Qwen2.5-32B-Instruct, the performance gain from adding prior experience is small or even negative. In particular, Qwen2.5-32B-Instruct achieves 0.82 accuracy without replay, with subsequent replay settings showing no clear improvement. In contrast, the Llama-3.1 series shows consistent and stable gains as more experience is provided, with Llama-3.1-70B-Instruct reaching 0.90 accuracy with 64 examples. This indicates that model architecture may substantially impact the utility of experience replay: some models may inherently learn well from single episodes, while others benefit more from historical trajectories.

\textbf{Reasoning LLMs vs. non-Reasoning LLMs.} Models designed for complex reasoning, such as DeepSeek-R1-Distill-Llama-8B and DeepSeek-R1-Distill-Qwen-7B, perform significantly worse and are prone to OOM failures at large replay sizes. These reasoning-optimized models tend to generate verbose thought chains and redundant intermediate outputs, which increase input length and can confuse the execution environment \cite{chen2024not}. This highlights a key difference between \LifelongAgentBench and prior benchmarks such as LiveCodeBench \cite{jain2024livecodebench} and GPQA \cite{rein2024gpqa}, where complex multi-hop reasoning is the primary objective. In contrast, \LifelongAgentBench emphasizes efficient skill acquisition and knowledge reuse from past interactions.

\textbf{Model Size.} We observe a scaling trend where larger backbones consistently outperform their smaller counterparts across most replay settings. Llama-3.1-70B-Instruct demonstrates superior robustness, achieving the highest accuracy (0.90) without encountering OOM even at 64 replay examples. Interestingly, medium-scale models such as Llama-3.1-8B-Instruct achieve comparable performance (0.78 at 64 examples), suggesting that with careful experience replay and model tuning, smaller models can approach the performance of much larger ones while offering substantially lower computational cost.

\subsection{Effect of Task Difficulty}

\textbf{Experience replay helps most on complex tasks.} We first analyze the DB environment, where task difficulty is manually categorized as Easy, Medium, or Hard based on required SQL skill combinations. As shown in Table~\ref{tab:db_difficulty}, experience replay provides marginal gains on Easy tasks (70\% to 76\%) but leads to substantial improvements on Hard tasks (49\% to 62\%). This suggests that replay is particularly valuable when agents face complex, multi-skill reasoning where prior examples offer useful reference points.

\begin{wraptable}{r}{5cm}
% \begin{table}[!t]
\tiny
\centering
\caption{Performance on different difficulty levels in DB.}
\label{tab:db_difficulty}
\resizebox{\linewidth}{!}{
    \begin{tabular}{ccccc}
    \toprule
    Difficulty & Exp=0 & Exp=1 & Exp=2 & Exp=8 \\
    \midrule
    Easy  & 0.70  & \textbf{0.76 } & 0.71  & 0.75  \\
    Medium & 0.56  & 0.53  & \textbf{0.59 } & 0.54  \\
    Hard  & 0.49  & 0.59  & 0.57  & \textbf{0.62 } \\
    \bottomrule
\end{tabular}
}
% \end{table}
\end{wraptable}

\textbf{Task length strongly correlates with replay benefit in KG.} In the KG environment, task difficulty is naturally reflected by the length of the ground-truth action sequence. Table~\ref{tab:kg_difficulty} shows that short tasks (length 2--4) benefit significantly from replay (e.g., length 2 improves from 48\% to 84\%), while longer tasks (length 7--9) see minimal or no improvement. As trajectory length increases, the added experience creates longer input sequences, which reduces the effective signal-to-noise ratio and increases the risk of context overflow or degraded performance.

\begin{wraptable}{r}{7cm}
% \begin{table}[!t]
\tiny
\centering
\caption{Performance by action sequence length in KG under different replay sizes.}
\label{tab:kg_difficulty}
\resizebox{0.99\linewidth}{!}{
    \begin{tabular}{c|c|cccc}
    \toprule
    \makecell{\# \textbf{Ground-Truth}\\ \textbf{Actions}} & \# \textbf{Task} & \textbf{Exp=0} & \textbf{Exp=1} & \textbf{Exp=4} & \textbf{Exp=16} \\
    \midrule
    \textbf{2} & 19114 & 0.48  & 0.72  & 0.78  & \textbf{0.84 } \\
    \rowcolor[rgb]{ .949,  .949,  .949} \textbf{3} & 1481  & 0.52  & 0.70  & \textbf{0.72 } & \textbf{0.72 } \\
    \textbf{4} & 7067  & 0.56  & 0.56  & 0.72  & \textbf{0.78 } \\
    \rowcolor[rgb]{ .949,  .949,  .949} \textbf{5} & 1740  & 0.16  & 0.30  & 0.30  & \textbf{0.44 } \\
    \textbf{6} & 626   & \textbf{0.14 } & 0.30  & 0.10  & \textbf{0.14 } \\
    \rowcolor[rgb]{ .949,  .949,  .949} \textbf{7} & 592   & 0.04  & 0.04  & \textbf{0.08 } & \textbf{0.08 } \\
    \textbf{8} & 63    & 0.08  & \textbf{0.12 } & 0.08  & 0.08  \\
    \rowcolor[rgb]{ .949,  .949,  .949} \textbf{9} & 46    & 0.11  & 0.13  & 0.13  & \textbf{0.17 } \\
    \bottomrule
\end{tabular}
}
% \end{table}
\end{wraptable}

\textbf{\LifelongAgentBench sensitively captures replay–difficulty interactions.} Overall, these findings demonstrate that \LifelongAgentBench provides a fine-grained benchmark to study how prior experience impacts learning under varying task difficulty. While experience replay is highly beneficial for short, well-bounded tasks, it poses scalability challenges for long-horizon tasks. Designing more effective memory compression, filtering, or retrieval strategies to handle these cases remains an important direction for future research.

\subsection{Scaling Experience with Group Self-Consistency}

% \begin{wrapfigure}{r}{0.5\textwidth}
\begin{figure}[!t]
\centering
\includegraphics[width=\linewidth]{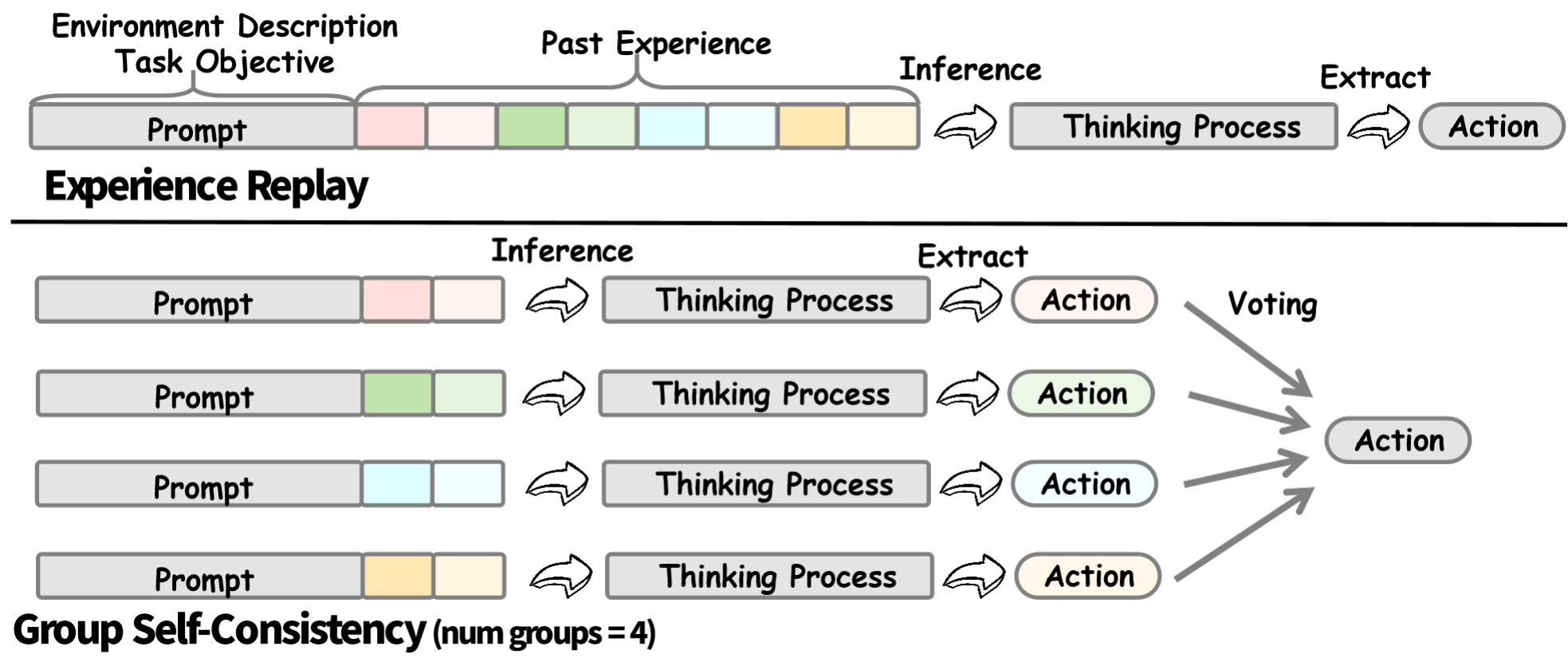}
\caption{Illustration of group self-consistency.}
\label{fig:group_self_consistency}
\end{figure}
% \end{wrapfigure}

\textbf{Group self-consistency reduces memory and stabilizes performance.} To mitigate the memory and inference overhead of large-scale experience replay, we propose group self-consistency, which splits retrieved experiences into smaller groups and aggregates their predictions using self-consistency voting \cite{wang2022self}. An illustration is provided in Figure~\ref{fig:group_self_consistency}. The experimental results are summarized in Tables~\ref{tab:gsc_accuracy_token}.

\textbf{Significant accuracy gains in DB.} In DB, group self-consistency provides clear performance improvements as replay size increases. Llama-3.1-8B-Instruct achieves 0.75 accuracy with 16 groups (16 examples), compared to only 0.61 without grouping. Qwen2.5-7B-Instruct similarly benefits, improving from 0.72 to 0.77 accuracy. Smaller models such as DeepSeek-R1-Distill variants show limited gains, likely constrained by reduced model capacity.

\textbf{Drastic memory savings in KG tasks.} In KG, where experience trajectories are much longer, group self-consistency dramatically reduces input token lengths. For Llama-3.1-8B-Instruct, token usage at 16 examples drops from 56,409 tokens (no grouping) to 11,002 tokens (16 groups), while maintaining stable accuracy. Qwen2.5-7B-Instruct shows similar trends, decreasing from 59,950 to 11,339 tokens with minimal accuracy loss.

\textbf{Group self-consistency offers a scalable replay strategy.} Overall, group self-consistency consistently stabilizes performance and alleviates memory bottlenecks across environments and models. Its simplicity and effectiveness make it a promising technique to enhance lifelong learning scalability for LLM agents under heavy replay loads. Future work may explore dynamic or adaptive grouping strategies that optimize the trade-off between experience diversity, inference cost, and available context window.

\begin{table}[!t]
  \centering
  \caption{Comparison of the accuracy (average input tokens) under different group self-consistency settings.}
  \resizebox{\linewidth}{!}{
    \begin{tabular}{cl|c|c|ccc|ccc}
    \toprule
    \multirow{2}[4]{*}{\textbf{Environment}} & \multicolumn{1}{c}{\# \textbf{Experience}} & \textbf{0} & \textbf{1} & \multicolumn{3}{c|}{\textbf{4}} & \multicolumn{3}{c}{\textbf{16}} \\
\cmidrule{2-10}          & \multicolumn{1}{c}{\# \textbf{Groups}} & \textbf{/} & \textbf{1} & \textbf{1} & \textbf{2} & \textbf{4} & \textbf{1} & \textbf{4} & \textbf{16} \\
    \midrule
    \multirow{4}[2]{*}{\textbf{DB}} & \textbf{Llama-3.1-8B-Instruct} & 0.16 (1128) & 0.51 (1773) & 0.63 (4189) & 0.57 (2750) & 0.59 (2512) & 0.61 (17874) & 0.70 (6008) & \textbf{0.75 (2888)} \\
          & \textbf{DeepSeek-R1-Distill-Llama-8B} & 0.06 (1213) & 0.11 (2706) & 0.13 (6542) & 0.12 (4645) & 0.13 (3469) & 0.12 (27737) & 0.13 (10365) & \textbf{0.18 (4932)} \\
          & \textbf{Qwen2.5-7B-Instruct} & 0.76 (951) & 0.69 (1567) & 0.68 (3952) & 0.73 (2445) & 0.73 (2292) & 0.72 (16383) & 0.71 (5443) & \textbf{0.77 (2685)} \\
          & \textbf{DeepSeek-R1-Distill-Qwen-7B} & 0.11 (1232) & 0.09 (2353) & 0.16 (5830) & 0.26 (3861) & 0.12 (2292) & 0.15 (18845) & \textbf{0.23 (6113)} & 0.15 (2744) \\
    \midrule
    \multirow{2}[2]{*}{\textbf{KG}} & \textbf{Llama-3.1-8B-Instruct} & 0.27 (2978) & 0.32 (9390) & 0.26 (16420) & 0.33 (12175) & \textbf{0.36 (8842)} & 0.32 (56409) & 0.34 (22191) & 0.34 (11002) \\
          & \textbf{Qwen2.5-7B-Instruct} & 0.22 (2858) & 0.27 (7150) & 0.19 (18436) & 0.29 (10998) & 0.30 (7522) & 0.19 (59950) & 0.19 (18539) & \textbf{0.36 (11339)} \\
    \bottomrule
    \end{tabular}%
    }
  \label{tab:gsc_accuracy_token}
\end{table}%

\subsection{Failure Mode Analysis}

To understand the failure patterns of LLM agents in \LifelongAgentBenchNoSpace, we classify all task outcomes based on agent behavior and system status. We observe four common failure modes: 
(1) \textbf{Incorrect final submission}: the agent outputs an answer in the correct format but with wrong content (\textit{completed}, Appendix \ref{sec:appendix_error_cases:complete}). 
(2) \textbf{Failure to commit}: the agent completes multiple operations but never explicitly submits the final answer (\textit{task\_limit\_reached}, Appendix \ref{sec:appendix_error_cases:task_limit_reached}).
(3) \textbf{Format violation}: the agent violates the required output format or instruction pattern (\textit{agent\_validation\_failed}, Appendix \ref{sec:appendix_error_cases:agent_validation_failed}).
(4) \textbf{Context overflow}: the agent exceeds the LLM context window due to excessive interactions or large intermediate outputs (\textit{agent\_context\_limit}, Appendix \ref{sec:appendix_error_cases:agent_context_limit}).
These results reveal key limitations of current LLM agents in multi-step interactive tasks: unstable reasoning, poor instruction adherence, and context management issues. Detailed examples of typical cases are provided in Appendix~\ref{sec:appendix_error_cases}.

\section{Conclusion, Limitation, and Future Works}
\label{sec:conclusion}

We present \LifelongAgentBenchNoSpace, the first unified benchmark specifically designed to evaluate the lifelong learning capabilities of LLM-based agents. Unlike prior benchmarks that treat agents as static systems, \LifelongAgentBench systematically measures agents’ ability to accumulate, retain, and transfer knowledge across diverse interactive environments. Our experiments demonstrate the potential of experience replay and group self-consistency to improve agent performance, while also revealing critical challenges.

Despite these advances, limitations remain. Experience replay introduces substantial memory and context length overhead, especially on long-horizon tasks. Performance also varies across model architectures, with smaller or reasoning-optimized models benefiting less from replay.

\LifelongAgentBench establishes a standardized platform for studying continual adaptation in agents, providing clear baselines and diagnostic tools to facilitate further research. We hope this work will inspire the development of more scalable, robust, and memory-efficient lifelong learning agents. Promising directions include more efficient memory retrieval strategies, dynamic experience selection, and extending the benchmark to multi-modal and real-world agent tasks.

\bibliography{reference}

\begin{thebibliography}{10}

\bibitem{achiam2023gpt}
Josh Achiam, Steven Adler, Sandhini Agarwal, Lama Ahmad, Ilge Akkaya, Florencia~Leoni Aleman, Diogo Almeida, Janko Altenschmidt, Sam Altman, Shyamal Anadkat, et~al.
\newblock Gpt-4 technical report.
\newblock {\em ArXiv preprint}, abs/2303.08774, 2023.

\bibitem{chen2024not}
Xingyu Chen, Jiahao Xu, Tian Liang, Zhiwei He, Jianhui Pang, Dian Yu, Linfeng Song, Qiuzhi Liu, Mengfei Zhou, Zhuosheng Zhang, et~al.
\newblock Do not think that much for 2+ 3=? on the overthinking of o1-like llms.
\newblock {\em ArXiv preprint}, abs/2412.21187, 2024.

\bibitem{zhu2023minecraft}
Linxi Fan, Guanzhi Wang, Yunfan Jiang, Ajay Mandlekar, Yuncong Yang, Haoyi Zhu, Andrew Tang, De{-}An Huang, Yuke Zhu, and Anima Anandkumar.
\newblock Minedojo: Building open-ended embodied agents with internet-scale knowledge.
\newblock In Sanmi Koyejo, S.~Mohamed, A.~Agarwal, Danielle Belgrave, K.~Cho, and A.~Oh, editors, {\em Advances in Neural Information Processing Systems 35: Annual Conference on Neural Information Processing Systems 2022, NeurIPS 2022, New Orleans, LA, USA, November 28 - December 9, 2022}, 2022.

\bibitem{french1999catastrophic}
Robert~M French.
\newblock Catastrophic forgetting in connectionist networks.
\newblock {\em Trends in cognitive sciences}, 3(4):128--135, 1999.

\bibitem{grattafiori2024llama}
Aaron Grattafiori, Abhimanyu Dubey, Abhinav Jauhri, Abhinav Pandey, Abhishek Kadian, Ahmad Al-Dahle, Aiesha Letman, Akhil Mathur, Alan Schelten, Alex Vaughan, et~al.
\newblock The llama 3 herd of models.
\newblock {\em ArXiv preprint}, abs/2407.21783, 2024.

\bibitem{gu2021beyond}
Yu~Gu, Sue Kase, Michelle Vanni, Brian~M. Sadler, Percy Liang, Xifeng Yan, and Yu~Su.
\newblock Beyond {I.I.D.:} three levels of generalization for question answering on knowledge bases.
\newblock In Jure Leskovec, Marko Grobelnik, Marc Najork, Jie Tang, and Leila Zia, editors, {\em {WWW} '21: The Web Conference 2021, Virtual Event / Ljubljana, Slovenia, April 19-23, 2021}, pages 3477--3488. {ACM} / {IW3C2}, 2021.

\bibitem{guo2025deepseek}
Daya Guo, Dejian Yang, Haowei Zhang, Junxiao Song, Ruoyu Zhang, Runxin Xu, Qihao Zhu, Shirong Ma, Peiyi Wang, Xiao Bi, et~al.
\newblock Deepseek-r1: Incentivizing reasoning capability in llms via reinforcement learning.
\newblock {\em ArXiv preprint}, abs/2501.12948, 2025.

\bibitem{jain2024livecodebench}
Naman Jain, King Han, Alex Gu, Wen-Ding Li, Fanjia Yan, Tianjun Zhang, Sida Wang, Armando Solar-Lezama, Koushik Sen, and Ion Stoica.
\newblock Livecodebench: Holistic and contamination free evaluation of large language models for code.
\newblock {\em ArXiv preprint}, abs/2403.07974, 2024.

\bibitem{koh2024tree}
Jing~Yu Koh, Stephen McAleer, Daniel Fried, and Ruslan Salakhutdinov.
\newblock Tree search for language model agents.
\newblock {\em ArXiv preprint}, abs/2407.01476, 2024.

\bibitem{Koh2024VisualWebArena}
Jingxiang Koh et~al.
\newblock Visualwebarena: Benchmarking vision-language agents for web interaction.
\newblock {\em ArXiv preprint}, abs/2402.09556, 2024.

\bibitem{li2025system}
Zhong-Zhi Li, Duzhen Zhang, Ming-Liang Zhang, Jiaxin Zhang, Zengyan Liu, Yuxuan Yao, Haotian Xu, Junhao Zheng, Pei-Jie Wang, Xiuyi Chen, et~al.
\newblock From system 1 to system 2: A survey of reasoning large language models.
\newblock {\em arXiv preprint arXiv:2502.17419}, 2025.

\bibitem{Liu2023AgentBench}
Han Liu et~al.
\newblock Agentbench: Benchmarking llms as agents.
\newblock {\em ArXiv preprint}, abs/2308.04035, 2023.

\bibitem{liu2024visualagentbench}
Xiao Liu, Tianjie Zhang, Yu~Gu, Iat~Long Iong, Yifan Xu, Xixuan Song, Shudan Zhang, Hanyu Lai, Xinyi Liu, Hanlin Zhao, et~al.
\newblock Visualagentbench: Towards large multimodal models as visual foundation agents.
\newblock {\em ArXiv preprint}, abs/2408.06327, 2024.

\bibitem{qin2023toolllm}
Libo Qin et~al.
\newblock Toolllm: Facilitating language model reasoning with tool understanding.
\newblock {\em ArXiv preprint}, abs/2309.03409, 2023.

\bibitem{rein2024gpqa}
David Rein, Betty~Li Hou, Asa~Cooper Stickland, Jackson Petty, Richard~Yuanzhe Pang, Julien Dirani, Julian Michael, and Samuel~R Bowman.
\newblock Gpqa: A graduate-level google-proof q\&a benchmark.
\newblock In {\em First Conference on Language Modeling}, 2024.

\bibitem{tian2024debugbench}
Runchu Tian, Yining Ye, Yujia Qin, Xin Cong, Yankai Lin, Yinxu Pan, Yesai Wu, Haotian Hui, Weichuan Liu, Zhiyuan Liu, et~al.
\newblock Debugbench: Evaluating debugging capability of large language models.
\newblock {\em ArXiv preprint}, abs/2401.04621, 2024.

\bibitem{wang2022self}
Xuezhi Wang, Jason Wei, Dale Schuurmans, Quoc~V. Le, Ed~H. Chi, Sharan Narang, Aakanksha Chowdhery, and Denny Zhou.
\newblock Self-consistency improves chain of thought reasoning in language models.
\newblock In {\em The Eleventh International Conference on Learning Representations, {ICLR} 2023, Kigali, Rwanda, May 1-5, 2023}. OpenReview.net, 2023.

\bibitem{yang2024qwen2}
An~Yang, Baosong Yang, Beichen Zhang, Binyuan Hui, Bo~Zheng, Bowen Yu, Chengyuan Li, Dayiheng Liu, Fei Huang, Haoran Wei, et~al.
\newblock Qwen2. 5 technical report.
\newblock {\em ArXiv preprint}, abs/2412.15115, 2024.

\bibitem{Yang2024AgentOccam}
Ke~Yang, Yao Liu, Sapana Chaudhary, Rasool Fakoor, Pratik Chaudhari, George Karypis, and Huzefa Rangwala.
\newblock Agentoccam: A simple yet strong baseline for llm-based web agents.
\newblock In {\em The Thirteenth International Conference on Learning Representations}, 2025.

\bibitem{zheng2025towards}
Junhao Zheng, Shengjie Qiu, Chengming Shi, and Qianli Ma.
\newblock Towards lifelong learning of large language models: A survey.
\newblock {\em ACM Computing Surveys}, 57(8):1--35, 2025.

\bibitem{Zheng2025Roadmap}
Junhao Zheng, Chengming Shi, Xidi Cai, Qiuke Li, Duzhen Zhang, Chenxing Li, Dong Yu, and Qianli Ma.
\newblock Lifelong learning of large language model based agents: A roadmap.
\newblock {\em ArXiv preprint}, abs/2501.07278, 2025.

\bibitem{zhouwebarena}
Shuyan Zhou, Frank~F Xu, Hao Zhu, Xuhui Zhou, Robert Lo, Abishek Sridhar, Xianyi Cheng, Tianyue Ou, Yonatan Bisk, Daniel Fried, et~al.
\newblock Webarena: A realistic web environment for building autonomous agents.
\newblock In {\em The Twelfth International Conference on Learning Representations}, 2024.

\end{thebibliography}
\bibliographystyle{plain}

%%%%%%%%%%%%%%%%%%%%%%%%%%%%%%%%%%%%%%%%%%%%%%%%%%%%%%%%%%%%

\appendix

\renewcommand*\contentsname{Appendix}
\clearpage
\addtocontents{toc}{\protect\setcounter{tocdepth}{2}}
\begin{center}
    \tableofcontents
\end{center}

\section{Lifelong Learning Environment Design}
\label{sec:lifelong_learning_environment_design}

\begin{table}[htbp]
\tiny
  \centering
  \caption{Skill set in each environment.}
  \begin{tabularx}{\linewidth}{>{\raggedright\arraybackslash}l >{\raggedright\arraybackslash}X}
    \toprule
    \textbf{Environment} & \textbf{Skills} \\
    \midrule
    DB &
    column\_alias, delete, group\_by\_multiple\_columns, group\_by\_single\_column, having\_aggregate\_calculation, having\_multiple\_conditions\_with\_aggregate, having\_single\_condition\_with\_aggregate, insert, limit\_and\_offset, limit\_only, order\_by\_multiple\_columns\_different\_directions, order\_by\_multiple\_columns\_same\_direction, order\_by\_single\_column, select, subquery\_multiple, subquery\_nested, ubquery\_single, table\_alias, update, where\_multiple\_conditions, where\_nested\_conditions, where\_single\_condition \\
    \midrule
    OS &
    addgroup, awk, cat, cd, chage, chgrp, chmod, chown, chsh, cp, echo, exit, find, gpasswd, grep, groupadd, ln, ls, mkdir, mv, rm, sed, sleep, tee, touch, useradd, usermod, vi, wc \\
    \midrule
    KG &
    get\_relations, get\_neighbors, intersection, get\_attributes, argmax, argmax, count \\
    \bottomrule
  \end{tabularx}
  \label{tab:skill_set}
\end{table}

In this section, we introduce the design principle and data construction process of each lifelong learning environment.
\subsection{Database}
\label{sec:lifelong_learning_environment_design:database}
\subsubsection{Environment Introduction}
\label{sec:lifelong_learning_environment_design:database:environment_introduction}
To ensure experiment reproducibility and reduce deployment complexity in the database environment, we encapsulate the MySQL database using Docker. At the start of each evaluation, the framework initializes a new Docker container from a predefined image. For each task, new tables are created based on task-specific information prior to execution and deleted upon task completion. This guarantees that modifications made during a task do not interfere with subsequent tasks. To optimize efficiency, the Docker container is launched only once at the beginning of the evaluation, rather than being recreated for each task, as the initialization of a MySQL container is time-consuming and frequent restarts would significantly increase overall evaluation time.

During each task, the agent interacts with the database based on the given goal $g$, performing operations such as querying, inserting, deleting, or updating records. For query tasks, the agent must submit the result at the end of the task, and the evaluation framework determines task success by comparing the submitted result with the ground truth. For other task types, the framework calculates a hash of the database state at the end of the task and compares it with a ground truth hash stored in the database to assess correctness.

In total, we define 22 skills within the database environment. The names and descriptions of these skills are as follows:

\begin{itemize}
  \item \textbf{column\_alias}: Assigning an alias to a column in a \texttt{SELECT} statement, e.g., \texttt{SELECT name AS employee\_name}.
  \item \textbf{delete}: Deleting data using the \texttt{DELETE} statement.
  \item \textbf{group\_by\_multiple\_columns}: Using the \texttt{GROUP BY} clause to group results by multiple columns.
  \item \textbf{group\_by\_single\_column}: Using the \texttt{GROUP BY} clause to group results by a single column.
  \item \textbf{having\_aggregate\_calculation}: Including calculations on aggregate function results in the \texttt{HAVING} clause, e.g., \texttt{MAX(salary) - MIN(salary)}.
  \item \textbf{having\_multiple\_conditions\_with\_aggregate}: Including multiple conditions involving aggregate functions in the \texttt{HAVING} clause.
  \item \textbf{having\_single\_condition\_with\_aggregate}: Including a single condition based on an aggregate function in the \texttt{HAVING} clause.
  \item \textbf{insert}: Inserting data using the \texttt{INSERT} statement.
  \item \textbf{limit\_and\_offset}: Using both \texttt{LIMIT} and \texttt{OFFSET} clauses together.
  \item \textbf{limit\_only}: Using only the \texttt{LIMIT} clause to restrict the number of returned rows, without using \texttt{OFFSET}.
  \item \textbf{order\_by\_multiple\_columns\_different\_directions}: Sorting results by multiple columns with different sorting directions in the \texttt{ORDER BY} clause.
  \item \textbf{order\_by\_multiple\_columns\_same\_direction}: Sorting results by multiple columns with the same sorting direction in the \texttt{ORDER BY} clause.
  \item \textbf{order\_by\_single\_column}: Sorting results by a single column in the \texttt{ORDER BY} clause.
  \item \textbf{select}: Querying data using the \texttt{SELECT} statement.
  \item \textbf{subquery\_multiple}: Having multiple subqueries in the main query.
  \item \textbf{subquery\_nested}: Nesting one subquery inside another subquery.
  \item \textbf{subquery\_single}: Including a single subquery in the \texttt{SELECT}, \texttt{WHERE}, or \texttt{HAVING} clause that involves only one table.
  \item \textbf{table\_alias}: Assigning an alias to a table in the \texttt{FROM} clause, e.g., \texttt{FROM employees AS e}.
  \item \textbf{update}: Updating data using the \texttt{UPDATE} statement.
  \item \textbf{where\_multiple\_conditions}: Including multiple conditions connected by \texttt{AND} or \texttt{OR} in the \texttt{WHERE} clause.
  \item \textbf{where\_nested\_conditions}: Including nested logical conditions in the \texttt{WHERE} clause, e.g., \texttt{WHERE (A AND B) OR C}.
  \item \textbf{where\_single\_condition}: Including a single condition in the \texttt{WHERE} clause, e.g., \texttt{WHERE id = 5}.
\end{itemize}

\subsubsection{Data Construction Process}
\label{sec:lifelong_learning_environment_design:database:data_construction}
\begin{figure}[!t]
    \centering
        \includegraphics[width=0.99\linewidth]{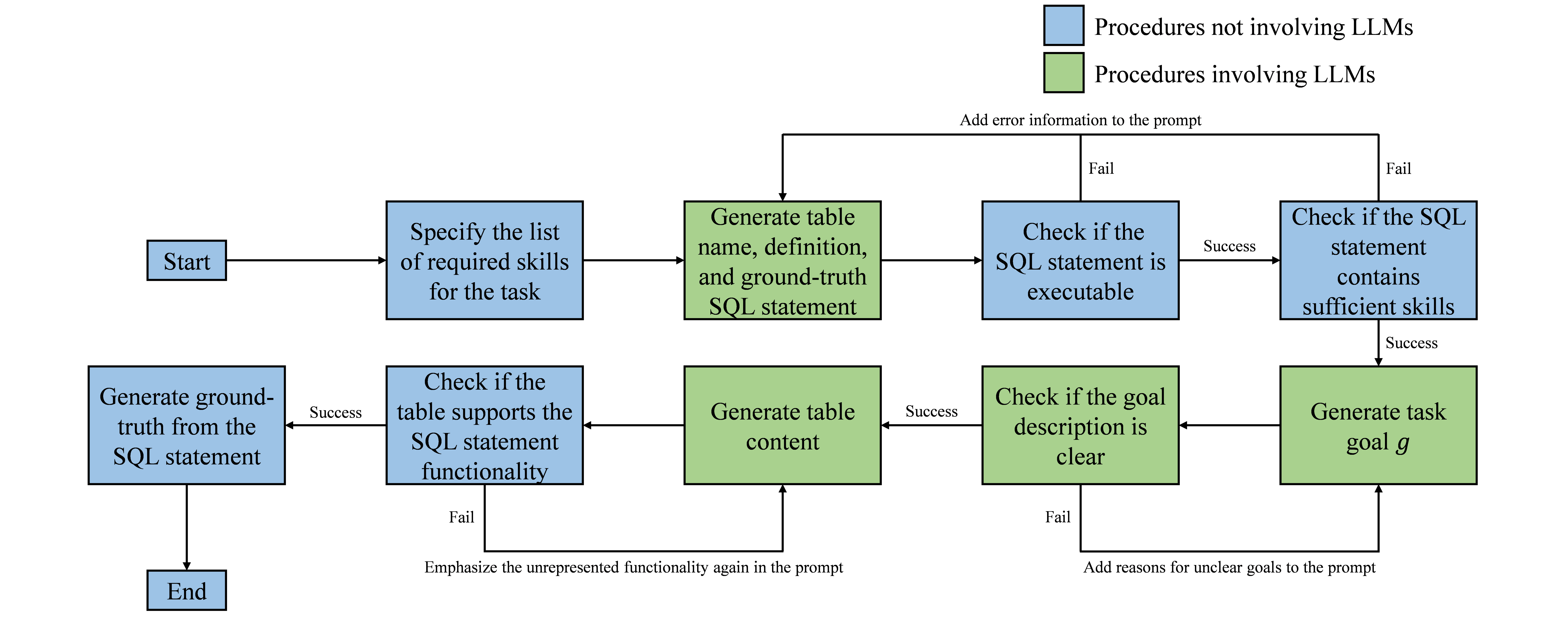}
    \caption{The data construction process in database environment.}
    \label{fig:data_construction_procedure_db}
\end{figure}

The task construction process in the database environment is illustrated in Figure \ref{fig:data_construction_procedure_db}. To prevent data leakage, we adopt the methodology proposed in \cite{tian2024debugbench}, leveraging a large language model to generate the benchmark data.

During data construction, we begin by specifying a list of skills that each task should incorporate. To ensure balanced coverage, skills with lower occurrence frequencies are assigned higher sampling probabilities. Additionally, we enforce a minimum occurrence threshold for each skill across the dataset to avoid the presence of isolated tasks.

\subsection{Operating System}
\label{sec:lifelong_learning_environment_design:operating_system}
\subsubsection{Environment Introduction}
\label{sec:lifelong_learning_environment_design:operating_system:environment_introduction}
In the operating system environment, we also use Docker containers to encapsulate the Ubuntu system with which the agent interacts. Due to the wide variety and broad impact of operations that the agent can perform in this environment, it is difficult to fully roll back the system state to its initial condition after each task. To address this, we destroy the current container instance after each task and instantiate a new one from the original image before the next task begins. This approach ensures both experimental reproducibility and isolation between tasks.

In the operating system environment, the agent is required to generate shell scripts composed of Bash commands based on the given goal $g$, in order to interact with the system. After each operation, the agent can choose either to continue issuing shell commands or to terminate the interaction by outputting a string that matches a predefined pattern.

Similarly, if the number of interactions between the agent and the environment reaches a predefined limit, the evaluation framework automatically terminates the interaction. Regardless of the reason for termination, the framework then determines whether the agent has successfully completed the task. Each task is associated with an evaluation script consisting of Bash commands. If the agent completes the task successfully, the script returns 0; otherwise, it returns a non-zero value. The success of the task is thus determined based on the return value of the evaluation script.

In the operating system environment, each task is associated with a specific skill, determined based on the ground truth Bash command required to solve the task. To ensure controlled and consistent evaluation, we restrict the agent to a predefined set of 29 Bash commands during task execution, with each command corresponding to a distinct skill. The complete list of these 29 skills is presented in Table \ref{tab:skill_set}.

\subsubsection{Data Construction Process}
\label{sec:lifelong_learning_environment_design:operating_system:data_construction}
\begin{figure}[!t]
    \centering
        \includegraphics[width=0.99\linewidth]{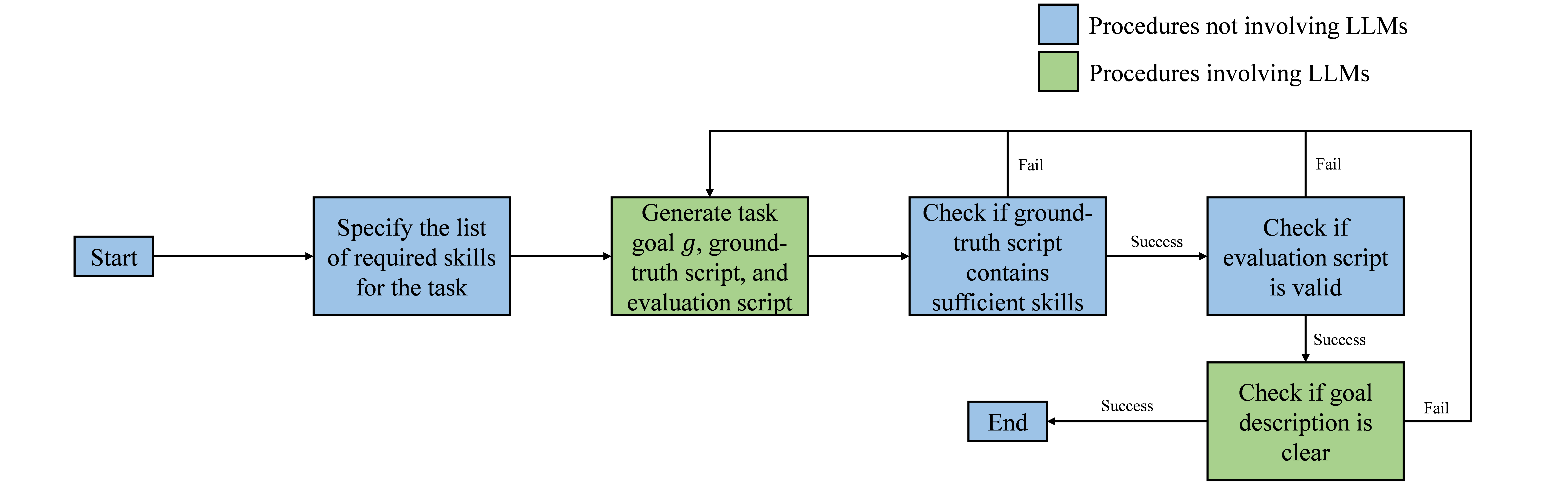}
    \caption{The data construction process in operating system environment.}
    \label{fig:data_construction_procedure_os}
\end{figure}

The task construction process in the operating system environment is illustrated in Figure \ref{fig:data_construction_procedure_os}. As in the database environment, we use a large language model to generate the tasks in the dataset.

To reduce token costs, however, we minimize the number of calls to the language model during data construction in the operating system environment compared to the database environment.

\subsection{Knowledge Graph}
\label{sec:lifelong_learning_environment_design:knowledge_graph}
\subsubsection{Environment Introduction}
\label{sec:lifelong_learning_environment_design:knowledge_graph:environment_introduction}
In the knowledge graph environment, the agent is tasked with achieving the goal $g$ by interacting with the knowledge graph, given both the goal and the entities involved. While humans typically query a knowledge graph using SPARQL statements, generating a complete SPARQL query in a single step is relatively challenging for an agent. This difficulty arises from the agent’s lack of basic knowledge about the graph structure—such as entity types, outgoing relations, and incoming relations. Directly including all this information in the prompt results in excessively long inputs, which increases the risk of the agent overlooking critical details \cite{Yang2024AgentOccam}. Moreover, such a direct approach to knowledge graph access is impractical for real-world applications.

Following the approach proposed in \cite{Liu2023AgentBench}, we use a constrained set of actions that the agent can execute. This design enables the agent to progressively and autonomously explore relevant information within the knowledge graph and achieve the goal $g$ through a sequence of actions.

During the interaction between the agent and the knowledge graph environment, the environment returns sets of relations and entities based on the actions executed by the agent. Because the agent must reference previously retrieved relations to construct new actions in subsequent steps, all queried relations are returned in full. For entity sets, we use the concept of \textit{variables} \cite{Liu2023AgentBench} to represent the results. Each variable is associated with the SPARQL statement that produced it, and instead of returning the set of queried entities directly, the environment returns the corresponding variable name to the agent. This mechanism allows the agent to conveniently reference previously queried entity sets in future actions, thereby enabling multi-hop reasoning over the knowledge graph.

In the knowledge graph environment, the set of available actions and their corresponding parameters is defined as follows. To complete a given task, the agent must execute a sequence of actions to achieve the goal $g$. Accordingly, we treat the actions appearing in the ground truth action sequence of each task as the task’s skills.

\begin{itemize}
  \item \textbf{get\_relations}: Given an entity or variable, returns all outgoing relations connected to it.
  \item \textbf{get\_neighbors}: Given an entity or variable and an outgoing relation, returns the set of entities reachable via that relation, represented as a new variable.
  \item \textbf{intersection}: Given two variables, returns a new variable representing the intersection of entities contained in both.
  \item \textbf{get\_attributes}: Given a variable, returns all attributes associated with the entities in that variable.
  \item \textbf{argmax}: Given a variable and an attribute, returns a new variable containing the entity (or entities) with the maximum value for the specified attribute.
  \item \textbf{argmin}: Given a variable and an attribute, returns a new variable containing the entity (or entities) with the minimum value for the specified attribute.
  \item \textbf{count}: Given a variable, returns the number of entities it contains.
\end{itemize}

\subsubsection{Data Construction Process}
\label{sec:lifelong_learning_environment_design:knowledge_graph:data_construction}
In the knowledge graph environment, tasks generated solely by large language models often lack practical relevance. To address this, we construct tasks based on existing knowledge graph question answering datasets. Specifically, we adopt the dataset introduced in \cite{gu2021beyond} and convert the S-expressions representing graph queries into ground truth action sequences for each task.
\section{Evaluation Framework Design}
\label{sec:appendix_evaluation_framework}
After constructing the environment and tasks, it is essential to develop a robust evaluation framework to facilitate the interaction between LLM agents and the environment. In designing our framework, we drew inspiration from widely-used web-based agent benchmarks such as WebArena\cite{zhouwebarena}, VisualWebArena\cite{Koh2024VisualWebArena}, as well as non-lifelong learning evaluation frameworks like AgentBench\cite{Liu2023AgentBench} and VisualAgentBench\cite{liu2024visualagentbench}. However, these frameworks commonly suffer from several limitations, including long evaluation times, incompatibility with lifelong learning scenarios, and the lack of a unified development interface.

First, the evaluation time for a single sample is often prolonged. Take WebArena as an example. WebArena is a benchmark designed to assess the performance of LLM agents in real-world web environments, encapsulating each deployed website within a Docker container. However, due to the complexity of the webpages used, instantiating each Docker container takes approximately one minute. To mitigate this issue, WebArena claims to employ a carefully orchestrated task sequence such that the execution of a preceding task does not affect subsequent ones. It therefore recommends initializing Docker containers only once before evaluation begins, rather than prior to each task. However, given that agent behaviors during evaluation are inherently unpredictable, determining whether tasks interfere with each other solely based on their goals \textit{g} is insufficient. Additionally, WebArena uses the accessibility tree of webpages as the environment observation. During the conversion to the accessibility tree, the evaluation framework traverses every HTML element in the DOM tree and maps it to a corresponding node. Even when using proprietary LLMs, the interaction time per task can reach up to ten minutes; for locally deployed open-source models, this duration increases further due to hardware constraints.

Second, these frameworks typically suggest evaluating agent performance via parallel execution of tasks, which is unsuitable for lifelong learning scenarios. For example, WebArena recommends partitioning the task sequence and running different sequences concurrently on separate servers. AgentBench manages all tasks using a central list and evaluates agent performance in parallel by distributing tasks across multiple processes in a process pool. When one process completes its assigned task, the framework automatically pulls the next one from the list. While these strategies reduce evaluation time, they fail to preserve task execution order—an essential factor in lifelong learning, where task sequencing determines the availability of transferable experience. Although both frameworks technically allow sequential task execution, doing so significantly increases overall evaluation time.

Third, these frameworks are primarily designed to assess the capabilities of different LLM agents, but do not provide a standardized interface for integrating or comparing various capability enhancement methods. This hampers fair comparisons. For instance, in WebArena, repeatedly resets webpage states within a single task using the built-in reset function to facilitate tree search. This practice is unrealistic, as real-world environments generally lack such reset mechanisms, and agent actions are often irreversible. Consequently, the comparison between the method\cite{koh2024tree} and those proposed in other studies is inherently unfair. This stems from the framework’s failure to provide a unified interface for the development of capability enhancement methods. Similarly, AgentBench separates the environment, controller, and agent into three independent processes that communicate via HTTP. Even if researchers have sufficient computational resources to deploy all components on a single machine, multi-process debugging introduces substantial development overhead. This complexity may explain why, compared to WebArena, fewer studies leverage AgentBench for investigating agent enhancement techniques.

To address these issues, we propose a new evaluation framework tailored to lifelong learning scenarios, aiming to support and accelerate the development of LLM-based agents and methods for continual capability improvement.

\subsection{Framework Architecture}
\label{sec:appendix_evaluation_framework:framework_architecture}
The evaluation framework consists of six key components: the language model pool, the agent, the environment, the chat history item factory, the controller, and the callbacks. The callback module is further composed of a callback handler and a list of callback functions.

\begin{figure}[!t]
    \centering
        \includegraphics[width=0.99\linewidth]{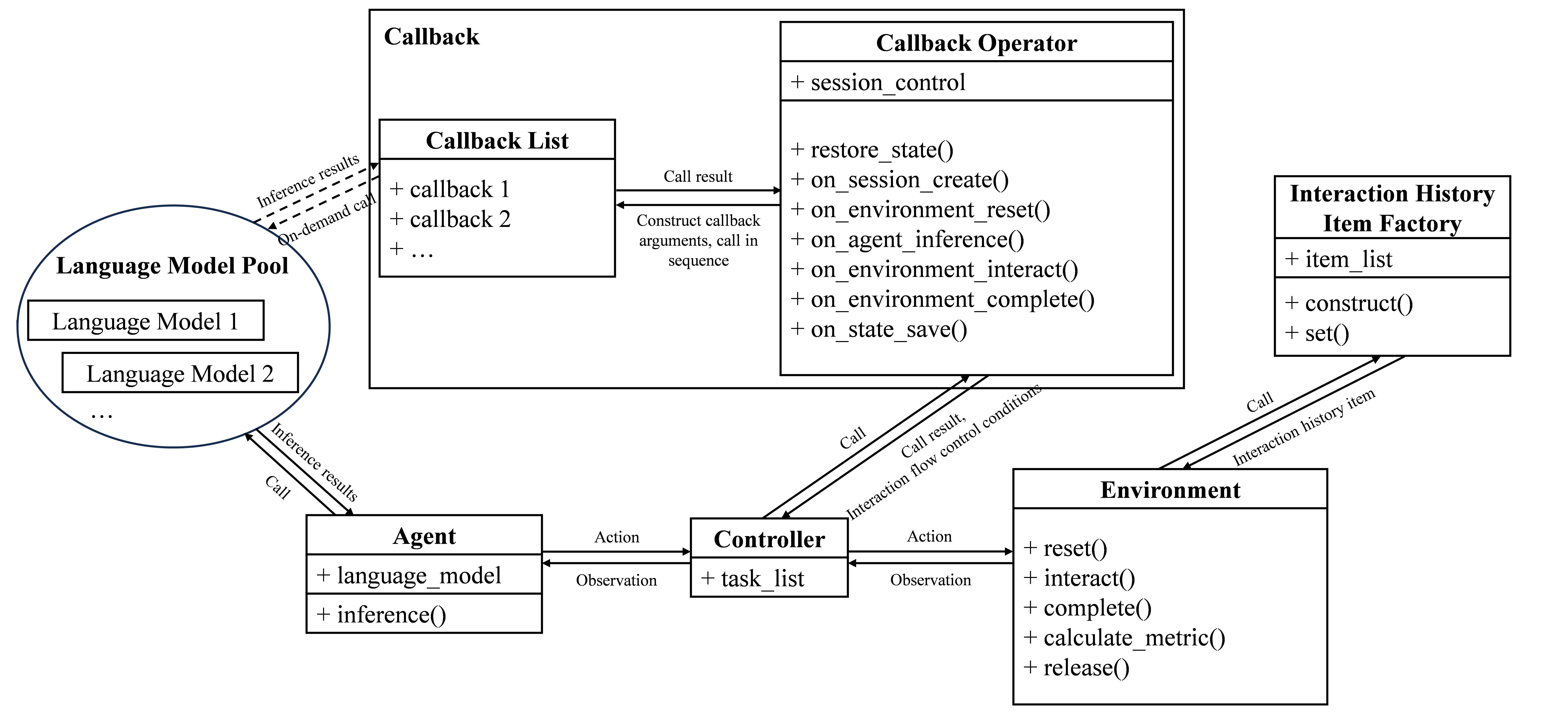}
    \caption{Evaluation Framework Architecture}
    \label{evaluation framework}
\end{figure}

\subsubsection{Language Model Pool}
\label{sec:appendix_evaluation_framework:framework_architecture:model_pool}
In the framework, the language model pool is implemented as a one-to-one mapping from string identifiers to language model objects. These model objects may belong to different language model classes, allowing for flexible integration of diverse models.

\subsubsection{Agent}
\label{sec:appendix_evaluation_framework:framework_architecture:agent}
The agent component is responsible for converting the historical actions taken by the agent, as well as past observations received from the environment, into a textual input sequence. This sequence is then fed into the language model to generate the agent’s next action through inference.

\subsubsection{Environment}
\label{sec:appendix_evaluation_framework:framework_architecture:environment}

The environment component is responsible for executing the agent’s actions and returning the corresponding responses to the controller. It provides five core methods: \texttt{reset}, \texttt{interact}, \texttt{complete}, \texttt{calculate\_metric}, and \texttt{release}. Specifically:

\begin{itemize}
  \item \textbf{\texttt{reset}}: Initializes the environment to the initial state $s_0$ at the beginning of each task and returns the task goal $g$. In environments that utilize Docker containers (e.g., the operating system environment), this method also creates a new container for agent interaction.

  \item \textbf{\texttt{interact}}: Executes the agent’s action and returns the resulting observation from the environment to the controller.

  \item \textbf{\texttt{complete}}: Evaluates whether the agent has successfully completed the task based on the goal-task reward function $R$. In environments where a new Docker container is created for each task (e.g., the operating system environment), this method also handles the destruction of the container. In environments that reuse a single container across tasks (e.g., the database environment), it restores the container to its initial state.

  \item \textbf{\texttt{calculate\_metric}}: Assesses the agent’s overall performance within a lifelong learning scenario, based on the cumulative rewards obtained across all tasks.

  \item \textbf{\texttt{release}}: Used in environments that do not require per-task container creation. It performs a unified cleanup by destroying the container to free system resources.
\end{itemize}

All test environments introduced in Section \ref{sec:environment_implementation} share a common abstract parent class. Each environment must implement all abstract methods defined by this parent class and may only expose the public methods specified therein to other components. This design not only facilitates the extension of the evaluation framework to support new environments, but also encourages the development of methods for enhancing the capabilities of large language model-based agents that are broadly applicable across various lifelong learning environments.

\subsubsection{Interaction History Item Factory}
\label{sec:appendix_evaluation_framework:framework_architecture:history_item_factory}
The interaction history item factory is responsible for initializing the interaction history based on the environment description and task goal $g$. During the environment's reset process, the construct method of the factory is called to generate the initial interaction history by combining the task goal $g$ with the initial observation $O_0$

Recent research has shown that incorporating historical experience into prompts or dynamically selecting prompt strategies based on environmental observations can significantly enhance the performance of LLM-based agents. To promote modular system design and decouple the logic of interaction history construction from environment-specific logic, we abstract this functionality into an independent interaction history item factory component. This design improves both the clarity and extensibility of the framework’s codebase.

\subsubsection{Controller}
\label{Controller}
\label{sec:appendix_evaluation_framework:framework_architecture:controller}

The controller component is primarily responsible for creating sessions, executing tasks, and storing the execution history of previously completed tasks.

A session is a structured object that encapsulates a task. In addition to recording key information such as the task index, interaction history, and the reward obtained by the agent, each session also maintains the session state that indicates its current status and the reason for task termination. As the task progresses, the interaction history grows incrementally. Both the agent and environment components can update the interaction history to exchange newly generated actions and observations, and they can also set the session state to notify the controller of any exceptional conditions encountered during task execution. Upon completion, each session is saved to a history list for future reference.

LLM-based experiments often span several hours or even days. To address this, our framework incorporates a snapshot mechanism that periodically persists critical data to disk at predefined intervals. This allows for efficient recovery and resumption in the event of unexpected interruptions or failures. Upon experiment startup, the controller attempts to restore any previously interrupted experiment. If the experiment is recoverable, the controller reads the necessary data from disk and skips already completed tasks, continuing with the remaining ones. After each task is completed and its corresponding session is added to the history list, the controller immediately saves the session data to disk. With this mechanism, an unexpected interruption results in the loss of, at most, a single task’s interaction data.

The controller is also responsible for calling callbacks at pre-defined events during task execution. We detail the callback mechanism in section \ref{sec:appendix_evaluation_framework:framework_architecture:callback}.

\subsubsection{Callback}
\label{sec:appendix_evaluation_framework:framework_architecture:callback}

Inspired by the \texttt{Trainer} module in the \textit{transformers} library, we designed a callback component for the evaluation framework. This component can manage multiple callbacks, each of which can modularly implement a specific function.

We manage all callbacks through a centralized list. Key points in task execution are defined as \textit{events}, and when an event occurs, all callbacks in the list are invoked sequentially. Each callback inherits from a common parent class, which provides event handler functions named after each event. By default, these handler functions do not affect task execution. However, researchers can inherit from the base callback class and override the event functions of interest. The customized callbacks can then be added to the callback list to influence task behavior and implement methods for enhancing agent capabilities.

This design ensures that all lifelong learning methods are developed and evaluated under consistent assumptions, thereby promoting comparability and fairness across experiments. Furthermore, since callbacks in the list are isolated and unaware of each other, researchers can combine multiple callbacks from different methods within a single experiment to explore the synergistic effects of various approaches on agent performance.

Each event handler function receives five arguments: the agent, the environment, the current session, the session history list, and the session controller. The agent and environment refer to the components described in Appendix \ref{sec:appendix_evaluation_framework:framework_architecture:agent} and Appendix \ref{sec:appendix_evaluation_framework:framework_architecture:environment}. The current session encapsulates a task, as detailed in Appendix \ref{sec:appendix_evaluation_framework:framework_architecture:controller}. 
The history session list is a key parameter for lifelong learning. Event handler functions are expected to extract useful experience from this list and enhance the large language model-based agent’s performance in lifelong learning scenarios by modifying the interaction history of the current task.

The session controller, a data member accessible to callbacks, consists of a set of boolean flags that can be used to skip specific steps in the interaction between the agent and the environment. For example, if an event handler detects that the agent’s response does not contain a valid action that the environment can interpret, it can use the session controller to skip the step of sending the action to the environment and instead prompt the agent to regenerate its response.

Notably, our framework allows each callback to optionally hold a reference to the language model during initialization and call it when needed. In contrast, most existing LLM-agent evaluation frameworks (e.g., AgentBench) do not decouple the LLM from the agent; instead, they treat the LLM-based agent as a single monolithic component. This design choice hinders the development of methods aimed at enhancing agent capabilities. For example, if a single LLM $L_1$ is used by an agent and two different methods, each requiring LLM $L_2$ access, are applied simultaneously via separate callbacks, and no model pool exists, both methods would redundantly load identical model arguments into GPU memory, as the callbacks are unaware of each other. 

Additionally, if a researcher wants to perform multiple forward passes within a callback to improve response quality, calling the agent’s inference API directly may be inappropriate, since it typically only supports inference for a single dialogue history, not for batched parallel queries. 

With the introduction of a shared language model pool, identical language model parameters used by both callbacks and the agent are loaded into GPU memory only once. This significantly reduces hardware requirements for developing agent capability enhancement methods and supports better modularity in the framework design.

We define a total of seven events for the callback mechanism. These events are as follows. During a single run of the evaluation framework, the \textit{restore\_state} event occurs at most once. Within the execution of a single task, \textit{on\_session\_create}, \textit{on\_environment\_reset}, \textit{on\_environment\_interact}, and \textit{on\_state\_save} each occur once, In contrast, \textit{on\_agent\_inference} and \textit{on\_environment\_interact} may be triggered multiple times throughout the task.
     
\begin{itemize}
    \item \texttt{restore\_state}:
    During the execution of an experiment, not only the controller's state but also the internal states of the callbacks (i.e., the values of their data members) may change. Callbacks should save their internal state to disk when the \texttt{on\_state\_save} event occurs and reload it when the \texttt{restore\_state} event is triggered.
    \item \texttt{on\_session\_create}:
    This event is triggered after a session is initialized. At this point, the session only contains the index of the assigned task.
    \item \texttt{on\_environment\_reset}:
    This event is triggered after the \texttt{reset} method of the environment is called.
    \item \texttt{on\_agent\_inference}:
    This event is triggered after the agent’s \texttt{inference} method is called.
    \item \texttt{on\_environment\_interact}:
    This event is triggered after the environment's \texttt{interact} method is called.
    \item \texttt{on\_environment\_complete}:
    This event is triggered after the environment's \texttt{complete} method is called.
    \item \texttt{on\_state\_save}: 
    This event occurs when the controller writes its state to disk.
\end{itemize}

\subsection{Differences Between Distributed Deployment and Single-Machine Deployment}
\label{sec:appendix_evaluation_framework:difference_single_distributed_deployment}
Nearly all LLM-based agent evaluation frameworks, including the one we propose, encapsulate the interactive environments for agents using Docker containers. However, due to the large number of parameters in LLM agents, they typically need to be deployed on high-performance computing (HPC) nodes. Most HPC nodes do not allow users to run Docker applications. As a result, most existing evaluation frameworks for LLM-based agents allow researchers to deploy the environment and the LLM on separate servers. WebArena \cite{zhouwebarena} and VisualWebArena \cite{Koh2024VisualWebArena} , which only consider environments in the form of webpages, deploy Docker containers containing the webpages on external servers, and then use headless browsers running on the LLM-hosting server to access those webpages. However, this approach clearly cannot be generalized to other types of environments. AgentBench \cite{Liu2023AgentBench} and VisualAgentBench \cite{liu2024visualagentbench} support distributed deployment via custom-built Remote Procedure Call (RPC) toolkit. Although their RPC toolkit can handle diverse environments such as web pages and databases, the introduction of RPC brings additional cognitive load to developers. Even when researchers use servers capable of running Docker and opt for small-parameter LLMs to develop agent enhancement methods, the use of RPC forces them to run different components of the evaluation framework in separate processes, significantly complicating debugging.

To address these issues, our evaluation framework is designed to support both single-machine and distributed deployment modes. During development, researchers can deploy all components on the same server and within a single process, enabling rapid prototyping and easier debugging. During evaluation, however, the framework also allows different components to be deployed across multiple servers and processes, making it possible to evaluate agents powered by large-scale language models in lifelong learning scenarios.

Crucially, the use of RPC should be transparent to researchers during development. The transition from local to distributed deployment should require only the execution of a small amount of additional, highly reusable code, allowing the framework to switch deployment modes seamlessly without introducing unnecessary complexity.

To achieve these goals, we designed and implemented a plug-and-play RPC toolkit based on a client-server architecture. During development, researchers can choose not to use this toolkit and debug all components of the framework within the same process. They can also use the toolkit to split components that need to be deployed on other servers into a client and a server, and convert component calls into remote procedure calls to enable distributed deployment of the evaluation framework. Throughout the development process, researchers can operate under the assumption that all components run within the same process, without needing to understand the internal details of the toolkit or RPC mechanisms. The plug-and-play nature of the toolkit allows effortless switching between single-machine and distributed deployments. The plug-and-play RPC toolkit helps reduce the cognitive load on researchers during development, shortens debugging time, and promotes the proposal of new agent enhancement methods in lifelong learning scenarios.  

We first introduce the RPC toolkit we implemented, and then introduce the structure of the framework under distributed deployment.

\begin{figure}[!t]
    \centering
        \includegraphics[width=0.99\linewidth]{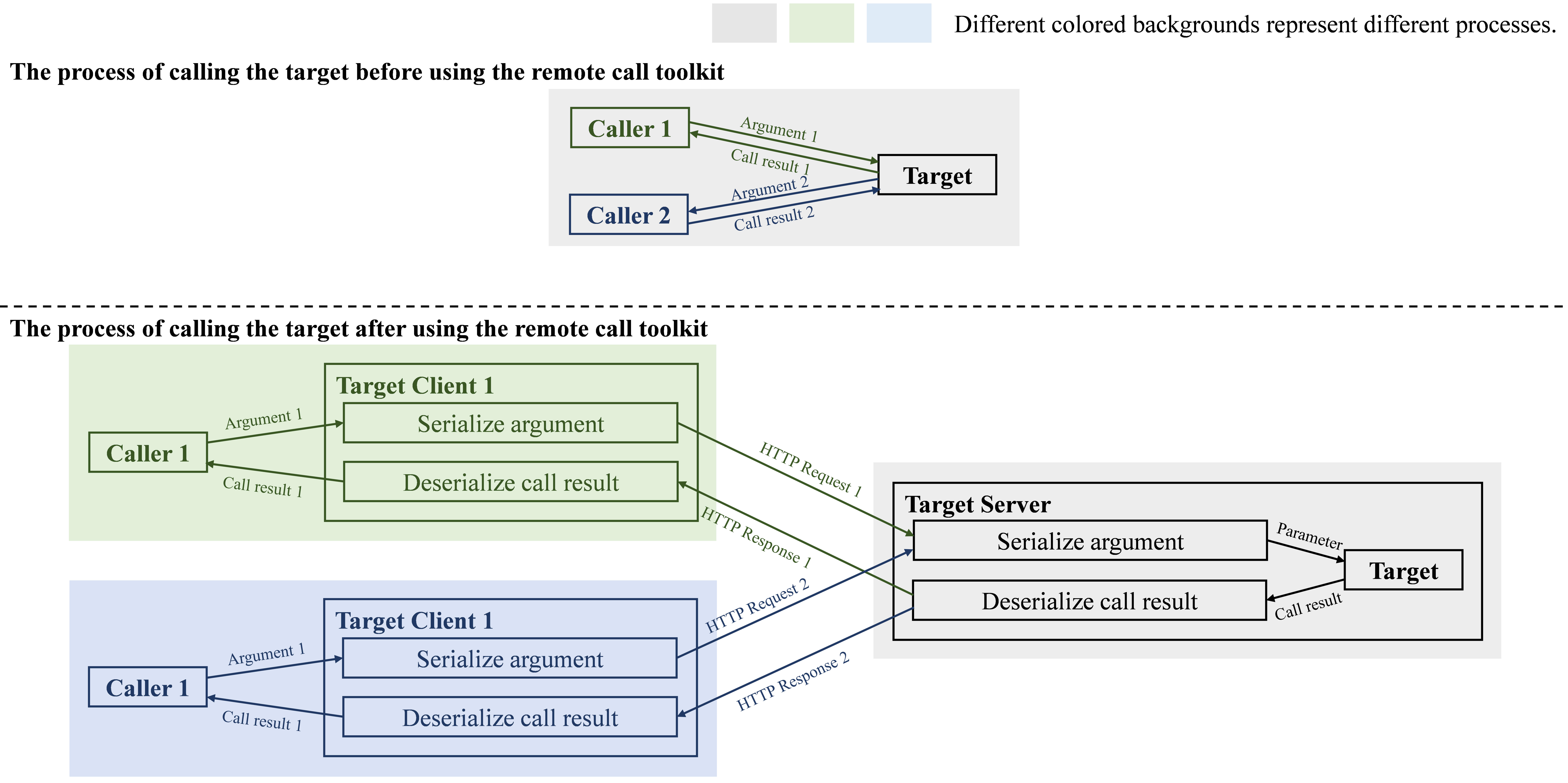}
    \caption{The Implementation of our RPC toolkit}
    \label{implementation_RPCT}
\end{figure}

\subsubsection{Remote Procedure Call Toolkit}
\label{sec:appendix_evaluation_framework:remote_procedure_call_toolkit}
The implementation of RPC toolkit is shown in the Figure \ref{implementation_RPCT}. In the figure, "client," "server," and "target" all refer to instantiated objects, but for brevity, we omit the word "object." We refer to the class that needs to be remotely called as the "target class." The target class can provide methods for other components of the evaluation framework to call. For each object of the target class (i.e., the target object), we wrap it with an object of the target server class (i.e., the target server object). Callers within the evaluation framework can access the target through the corresponding target client class object (i.e., the target client object). All target server classes and client classes are subclasses of the server and client base classes provided by our RPC toolkit. To implement RPC functionality, each target class must define its corresponding target server subclass and client subclass.

The target server continuously listens on a specific port on the server. When a caller calls a method of the target via the target client, the client first serializes the arguments and sends them via HTTP to the port that the target server is listening on. Upon receiving the HTTP request, the server deserializes the arguments into a format that the target can directly use and then performs the method call. After the call is completed, the result is returned to the target server, which then serializes the result and returns it via HTTP to the target client. The client then deserializes the result to obtain the correctly formatted output. The serialization and deserialization processes are mainly handled by the server and client base classes. The toolkit also supports reading and modifying data members in the same manner.

When implementing the server and client subclasses for a target, researchers only need to ensure that the target client provides the same interface as the target and defines the parameter types and return types of each interface using the \texttt{BaseModel} class from the \textit{Pydantic} library. This makes it possible to implement remote invocation of the target without needing to understand the specifics of the serialization and deserialization processes. Through this design, we decouple the process of data transfer between processes during remote invocation from the target’s data processing logic.

The RPC toolkit supports serialization and deserialization of various object types, including immutable built-in Python types, enumeration types, and any subclass of the client base class. We implement the client base class as a subclass of the \texttt{BaseModel}. Consequently, all concrete client classes are also subclasses of \texttt{BaseModel}. The \texttt{BaseModel} class requires developers to explicitly specify the types of data members via type annotations and automatically performs type validation during object construction. This ensures that objects maintain data integrity and type consistency after serialization and deserialization. 

By supporting serialization and deserialization of client subclass objects, our toolkit allow client subclass objects to be used as return values in remote procedure calls and passed back to the caller. This enables researchers to conveniently implement chained remote procedure calls, where the result obtained from one remote call can be directly used to initiate another remote procedure call, without the need to explicitly instantiate the client subclass object during development. The instantiation process is handled automatically by the toolkit, further simplifying the implementation.

\begin{figure}[!t]
    \centering
        \includegraphics[width=0.99\linewidth]{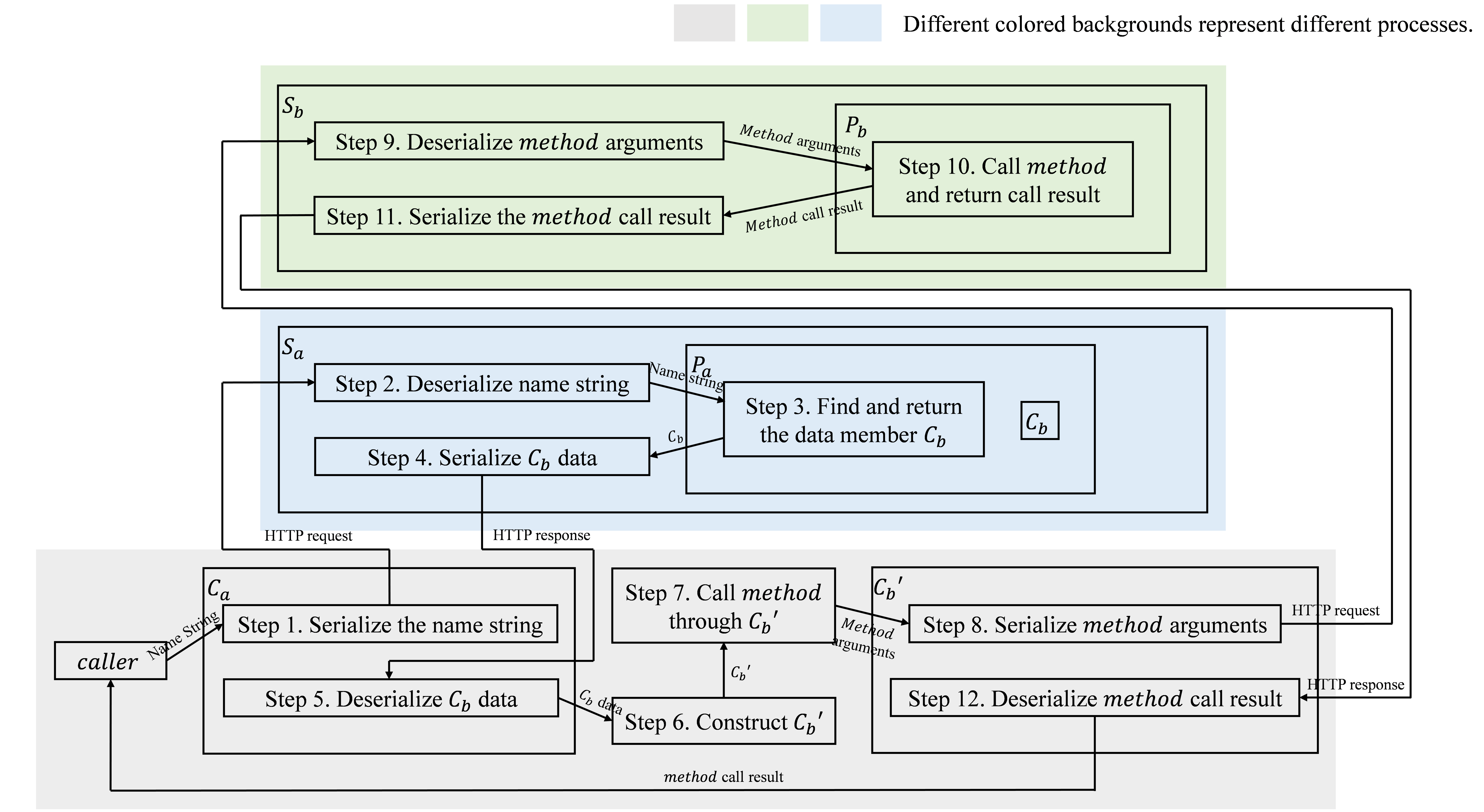}
    \caption{Example of Using our RPC Toolkit}
    \label{example_RPCT}
\end{figure}

The process of chained remote calls is illustrated in the Figure \ref{example_RPCT}. In the following example, there are two target objects, $P_a$ and $P_b$. $P_a$ contains $P_b$ as a data member. $P_a$ and $P_b$ can be two objects of the same target class or of different target classes. We assume the caller $caller$ wants to call the $method$ of $P_b$ via $P_a$, and $caller$, $P_a$, and $P_b$ are located in three different processes. Through the toolkit, what $caller$ actually accesses is the client object $C_a$ corresponding to $P_a$, and what $P_a$ actually stores is the client object $C_b$ corresponding to $P_b$.

Note that during development, researchers can still think of $caller$ as accessing $P_a$ rather than $C_a$, and $P_a$ as containing $P_b$ as a data member rather than $C_b$. When using the dot operator and calling $P_b$’s $method$, a remote procedure call is triggered. The dot operator takes a name string as input and returns the corresponding data member. In Figure \ref{example_RPCT}, the server objects corresponding to $P_a$ and $P_b$ are denoted as $S_a$ and $S_b$ respectively, and $C_b'$ is the internally instantiated client of $P_b$ created by the toolkit, which is transparent to the researcher and used to implement the chained remote call.

In the above remote procedure call process, all steps except for Step 3, Step 7, and Step 10 are implemented by the toolkit and are transparent to the researchers.

\subsubsection{Distributed Deployment Framework Structure}
\label{sec:appendix_evaluation_framework:distributed_deployment_framework}
\begin{figure}[!t]
    \centering
        \includegraphics[width=0.99\linewidth]{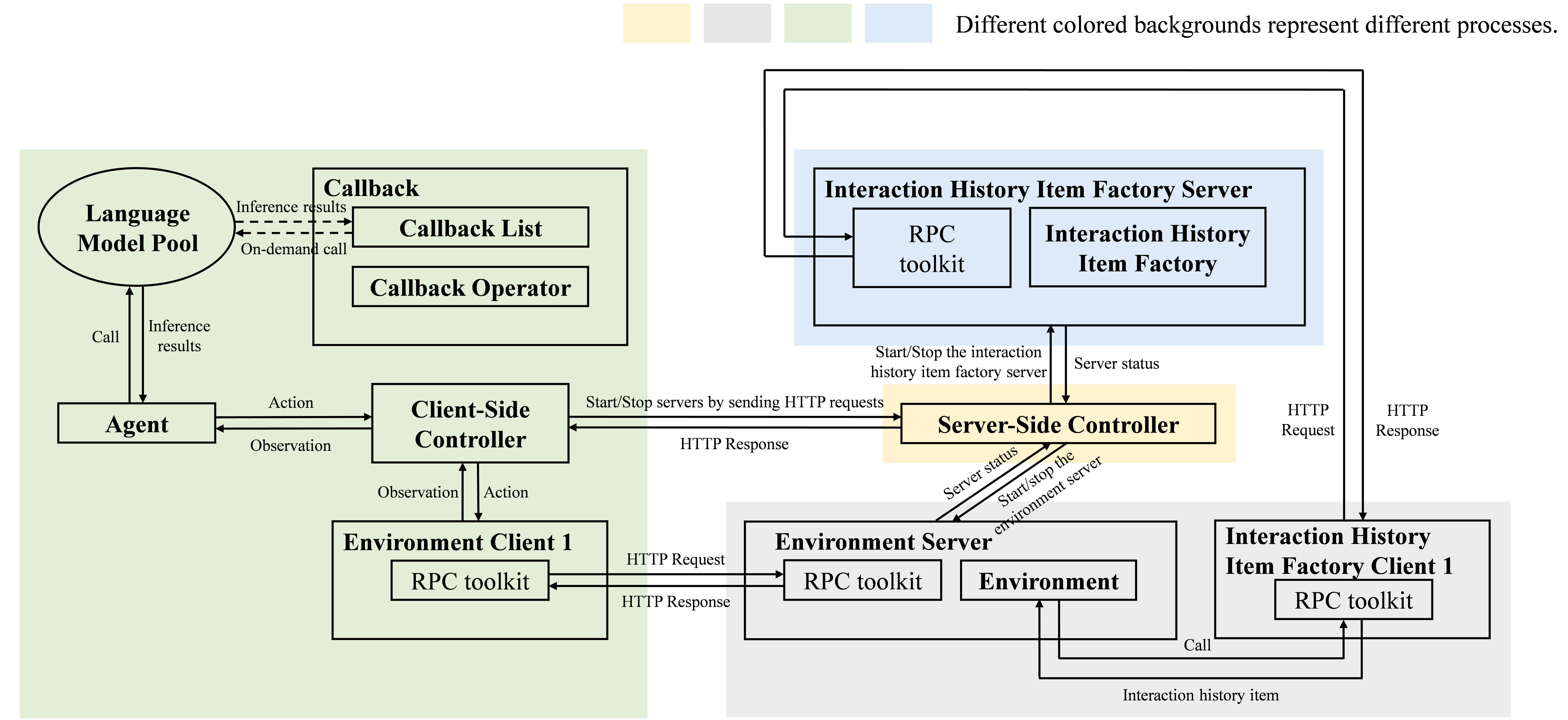}
    \caption{The Architecture of the Evaluation Framework in a Distributed Deployment}
    \label{architecture_deploy}
\end{figure}

The architecture of the evaluation framework under distributed deployment is illustrated in the Figure \ref{architecture_deploy}. For clarity, some internal details of the components and the RPC toolkit are omitted. During experiments, the client-side controller process typically runs on a high-performance computing node, while the controller server, interaction history item factory server, and environment server processes usually run on standard servers with Docker support. Communication between these distributed processes is handled by the RPC toolkit described in Appendix \ref{sec:appendix_evaluation_framework:remote_procedure_call_toolkit}.

\begin{figure}[!t]
    \centering
        \includegraphics[width=0.99\linewidth]{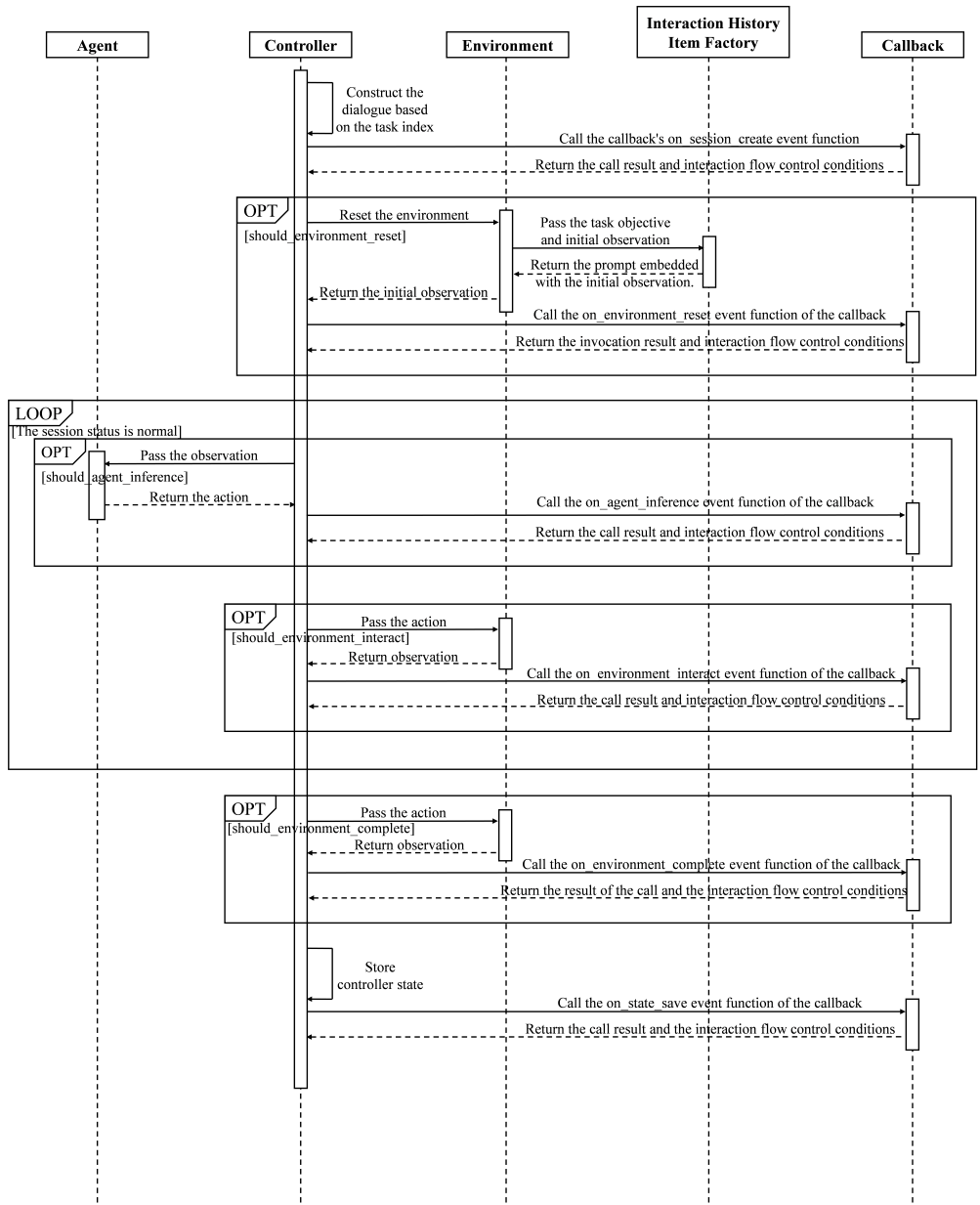}
    \caption{The execution process of an assessment task}
    \label{task_flow}
\end{figure}

Before the experiment begins, researchers first need to launch the server-side controller on a standard server, followed by launching the client-side controller on the high-performance computing node. The server-side controller continuously listens on a designated port. When the client-side controller starts, it sends an HTTP request to the server-side controller to start the environment server and the interaction history item factory server. Once the client-side controller receives confirmation of successful server initialization, it begins task scheduling. After all tasks are scheduled, the client-side controller sends another HTTP request to the server-side controller to shut down the interaction history item factory server and the environment server.

The end of an experiment corresponds to the termination of the client-side controller, interaction history item factory server, and environment server processes. In contrast, the server-side controller process continues to run. This design allows the server-side controller to continuously listen on a designated port and, upon the next launch of the client-side controller, reinitialize the environment server and interaction history item factory server based on the configuration file provided. In the distributed deployment, the client-side controller can be viewed as a wrapper around the single-machine controller, with additional functionality for starting and shutting down the necessary servers at the beginning and end of each experiment.

During the implementation of the evaluation framework, we found that if the interaction history item factory component is treated as a regular data member of the environment component and not wrapped in a server subclass from our RPC toolkit, modifications made to the interaction history item factory on the client side will not be synchronized with the environment server. Specifically, without running the interaction history item factory component in a separate process, the client-side controller's process can easily obtain a copy of the interaction history item factory that is identical to the one in the environment component. However, this copy is merely a local mutable variable on the frame of the client-side controller process, and any modifications made to its data members will not be synchronized with the interaction history item factory in the environment server. To resolve this issue, we run the interaction history item factory component in a separate process. Callers in the process where the environment server resides can access and modify the interaction history item factory through remote procedure calls. Meanwhile, callers in the process where the client-side controller resides can access and modify the interaction history item factory via chained remote procedure calls.

\subsection{The Execution Process of a Task}
\label{sec:appendix_evaluation_framework:excution_of_a_task}
To clearly demonstrate the execution process of the evaluation framework and show the specific timing of events during task execution, we present the execution process of a task in the evaluation process in Figure \ref{task_flow}. We assume that the evaluation framework is deployed on a single machine, so the components interact directly with each other without the need for the server or client in the RPC toolkit. In distributed deployment, for the researchers, the interaction methods and arguments passed between components remain the same as in a single-machine deployment. 

% \begin{figure}[!t]
%     \centering
%         \includegraphics[width=0.99\linewidth]{figures/appendix/task_flow.pdf}
%     \caption{The execution process of an assessment task}
%     \label{task_flow}
% \end{figure}

In the evaluation framework, the parameters of the callback component are passed by reference, so the callback does not explicitly return results to the controller. In the diagram, \texttt{should\_environment\_reset}, \texttt{should\_agent\_inference}, \texttt{should\_environment\_interact}, and \texttt{should\_environment\_complete} are the interaction flow control conditions determined by the callback component. All four interaction flow control conditions default to true. Callbacks can modify these control flags to influence the interaction process between the agent and the environment.
\section{Datasets Samples}
\label{sec:appendix_dataset_samples}

\subsection{Database Samples}

\begin{tcolorbox}[prompt_box_style, title={Environment: DB || Case 1}]
Instruction: Insert a new payment record for member ID 102 with a payment date of "2023-10-15", amount 75, and payment method "Credit Card" into the membership payments table. \textbackslash nThe name of this table is membership\_payments, and the headers of this table are payment\_id, member\_id, payment\_date, amount, payment\_method.

\bigskip

\begin{sloppypar}{\ttfamily Ground Truth Action: \textcolor{blue}{\texttt{INSERT INTO membership\_payments (member\_id, pay\\ment\_date, amount, payment\_method) VALUES (102, "2023-10-15", 75, "Cr\\edit Card")}}}\end{sloppypar}

\bigskip

Skill List: ["insert"]
\end{tcolorbox}

\begin{tcolorbox}[prompt_box_style, title={Environment: DB || Case 2}]
Instruction: Delete all entries where maintenance\_status is equal to 0.\textbackslash nThe name of this table is equipment\_service, and the headers of this table are equipment\_id, equipment\_type, installation\_date, maintenance\_status, service\_interval.

\bigskip

\begin{sloppypar}{\ttfamily Ground Truth Action: \textcolor{blue}{\texttt{DELETE FROM equipment\_service WHERE maintenance\_\\status = 0;}}}\end{sloppypar}

\bigskip

Skill List: ["delete", "where\_single\_condition"]

\end{tcolorbox}

\begin{tcolorbox}[prompt_box_style, title={Environment: DB || Case 3}]
Instruction: Update the result status to "completed" and hours spent to 48 for the experiment with ID equal to 105 in the experiment\_data table.\textbackslash nThe name of this table is experiment\_data, and the headers of this table are experiment\_id, researcher\_name, experiment\_date, result\_status, hours\_spent, samples\_used.

\bigskip

\begin{sloppypar}{\ttfamily Ground Truth Action: \textcolor{blue}{\texttt{UPDATE experiment\_data SET result\_status = "com\\pleted", hours\_spent = 48 WHERE experiment\_id = 105;}}}\end{sloppypar}

\bigskip

Skill List: ["update", "where\_single\_condition"]

\end{tcolorbox}

\begin{tcolorbox}[prompt_box_style, title={Environment: DB || Case 4}]
Instruction: What are the models, years, and prices of vehicles that are currently available? Return the model, year, and price, ordered by price from highest to lowest.\textbackslash nThe name of this table is vehicle\_inventory, and the headers of this table are vehicle\_id, model, year, price, status.

\bigskip

\begin{sloppypar}{\ttfamily Ground Truth Action: \textcolor{blue}{\texttt{SELECT model, year, price FROM vehicle\_inventory WHERE status = "available" ORDER BY price DESC;}}}\end{sloppypar}

\bigskip

Skill List: ["order\_by\_single\_column", "select", "where\_single\_condition"]

\end{tcolorbox}

\begin{tcolorbox}[prompt_box_style, title={Environment: DB || Case 5}]
Instruction: What are the property listings" id (listing ID), type (property type), price, and square feet? Return the results ordered by price from highest to lowest, then square feet from highest to lowest, and limit the output to 10 entries.\textbackslash nThe name of this table is property\_listings, and the headers of this table are listing\_id, property\_type, price, square\_feet, listed\_date, status, agent\_id.

\bigskip

\begin{sloppypar}{\ttfamily Ground Truth Action: \textcolor{blue} {\texttt{SELECT listing\_id AS id, property\_type AS typ\\e, price, square\_feet FROM property\_listings ORDER BY price DESC, square\_feet DESC LIMIT 10;}}}\end{sloppypar}

\bigskip

Skill List: ["column\_alias", "limit\_only", "order\_by\_multiple\_columns\_same\_direction", "select"]

\end{tcolorbox}

\begin{tcolorbox}[prompt_box_style, title={Environment: DB || Case 6}]
Instruction: Which routes have a distance greater than the average distance and less than the maximum distance? Return the route ID and driver name.\textbackslash nThe name of this table is delivery\_routes, and the headers of this table are route\_id, driver\_name, vehicle\_type, distance\_km, delivery\_status.

\bigskip

\begin{sloppypar}{\ttfamily Ground Truth Action: \textcolor{blue} {\texttt{SELECT route\_id, driver\_name FROM delivery\_rou\\tes WHERE distance\_km > (SELECT AVG(distance\_km) FROM delivery\_routes) AND distance\_km < (SELECT MAX(distance\_km) FROM delivery\_routes);}}}\end{sloppypar}

\bigskip

Skill List: ["select", "subquery\_multiple", "where\_multiple\_conditions"]

\end{tcolorbox}

\begin{tcolorbox}[prompt_box_style, title={Environment: DB || Case 7}]
Instruction: What are the owner IDs and the total number of vaccinations administered to dogs since January 1, 2023? Return the owner ID and corresponding total vaccinations, only including owners with more than 3 vaccinations.\textbackslash nThe name of this table is vaccination\_records, and the headers of this table are record\_id, owner\_id, pet\_type, vaccination\_type, vaccination\_date, veterinarian, dose\_number.

\bigskip

\begin{sloppypar}{\ttfamily Ground Truth Action: \textcolor{blue} {\texttt{SELECT owner\_id, COUNT(\*) AS total\_vaccinations FROM vaccination\_records WHERE pet\_type = "Dog" AND vaccination\_date >= "2023-01-01" GROUP BY owner\_id HAVING COUNT(\*) > 3;}}}\end{sloppypar}

\bigskip

Skill List: ["column\_alias", "group\_by\_single\_column",  "where\_multiple\_conditions", "having\_single\_condition\_with\_aggregate", "select"]

\end{tcolorbox}

% SSSSSSSSSSSSSSSSSSSSSSSSSSSSSSSSSSSSSSSSSSSSSSSSSSSSSSSSSSSSSSSSSSSSSSSSS

\begin{tcolorbox}[prompt_box_style, title={Environment: DB || Case 8}]
Instruction: What are the item IDs and names for items that are either: 1) in the "Electronics" category with stock quantity below the average stock quantity of \*\*all items\*\*, or 2) in the same category as the item with ID 100 (determined dynamically via its category)? Return the item ID and name.\textbackslash nThe name of this table is inventory\_management, and the headers of this table are item\_id, item\_name, category, stock\_quantity, last\_restock\_date.

\bigskip

\begin{sloppypar}{\ttfamily Ground Truth Action: \textcolor{blue} {\texttt{SELECT i.item\_id AS id, i.item\_name AS name FROM\\ inventory\_management AS i WHERE (i.category = "Electronics" AND i.stoc\\k\_quantity < (SELECT AVG(stock\_quantity) FROM inventory\_management))\\ OR i.category = (SELECT category FROM inventory\_management WHERE item\_\\id = 100);}}}\end{sloppypar}

\bigskip

Skill List: ["column\_alias", "subquery\_multiple", "table\_alias", "where\_nested\_conditions", "select"]

\end{tcolorbox}

\begin{tcolorbox}[prompt_box_style, title={Environment: DB || Case 9}]
Instruction: Which seat numbers have a price difference (maximum price minus minimum price) greater than 50? Return the seat number and price difference, ordered by seat number in descending order, and limit the results to 5 entries.\textbackslash nThe name of this table is ticket\_sales, and the headers of this table are ticket\_id, seat\_number, price, sale\_date, customer\_name, status.

\bigskip

\begin{sloppypar}{\ttfamily Ground Truth Action: \textcolor{blue} {\texttt{SELECT seat\_number, MAX(price)-MIN(price) AS pri\\ce\_diff FROM ticket\_sales GROUP BY seat\_number HAVING MAX(price)-MIN(p\\rice)>50 ORDER BY seat\_number DESC LIMIT 5;}}}\end{sloppypar}

\bigskip

Skill List: ["column\_alias", "group\_by\_single\_column", "having\_aggregate\_calculation", "having\_single\_condition\_with\_aggregate", "limit\_only", "order\_by\_single\_column", "select"]

\end{tcolorbox}

\begin{tcolorbox}[prompt_box_style, title={Environment: DB || Case 10}]
Instruction: What are the guest IDs, their average ratings, and total reviews for guests who have either given a 5-star rating or submitted a review before January 1, 2023, and have more than 5 total reviews with an average rating below 4.5? Return the results ordered by average rating in descending order and total reviews in ascending order.\textbackslash nThe name of this table is hotel\_reviews, and the headers of this table are review\_id, guest\_id, review\_text, rating, review\_date.

\bigskip

\begin{sloppypar}{\ttfamily Ground Truth Action: \textcolor{blue} {\texttt{SELECT guest\_id AS gid, AVG(rating) AS avg\_ratin\\g, COUNT(review\_id) AS total\_reviews FROM hotel\_reviews AS hr WHERE guest\_id IN (SELECT guest\_id FROM hotel\_reviews WHERE rating = 5) OR guest\_id IN (SELECT guest\_id FROM hotel\_reviews WHERE review\_date < "2\\023-01-01") GROUP BY guest\_id HAVING COUNT(review\_id) > 5 AND AVG(rati\\ng) < 4.5 ORDER BY avg\_rating DESC, total\_reviews ASC;}}}\end{sloppypar}

\bigskip

Skill List: ["where\_multiple\_conditions" "group\_by\_single\_column",  "select", "table\_alias", "subquery\_multiple", "column\_alias", "having\_multiple\_conditions\_with\_aggregate", "order\_by\_multiple\_columns\_different\_directions"]

\end{tcolorbox}

\subsection{Operating System Samples}

\begin{tcolorbox}[prompt_box_style, title={Environment: OS || Case 11}]
Instruction: Create three files "/tmp/file1", "/tmp/file2", and "/tmp/file3", setting each file"s permissions to 600 with a 1-second delay after creating each file. Use the "sleep" command between file operations.

\bigskip

\begin{sloppypar}{\ttfamily Ground Truth Action: \textcolor{blue} {"rm -f /tmp/file1 /tmp/file2 /tmp/file3"}}\end{sloppypar}

\bigskip

Skill List: ["chmod", "sleep", "touch"]

\end{tcolorbox}

% SSSSSSSSSSSSSSSSSSSSSSSSSSSSSSSSSSSSSSSSSSSSSSSSSSSSSSSSSSSSSSSSSSSSSSSSS

\begin{tcolorbox}[prompt_box_style, title={Environment: OS || Case 12}]
Instruction: Recursively change ownership of "/var/webapp" to user "webadmin", set directory permissions to 750, and file permissions to 640. Ensure hidden files in subdirectories are included.

\bigskip

\begin{sloppypar}{\ttfamily Ground Truth Action: \textcolor{blue} {"mkdir -p /var/webapp/\{public,private/.cache\} \&\& touch /var/webapp/public/index.html /var/webapp/private/.cache/temp.dat \&\& echo "config" > /var/webapp/private/.env"}}\end{sloppypar}

\bigskip

Skill List: ["chown", "find", "useradd"]

\end{tcolorbox}

\begin{tcolorbox}[prompt_box_style, title={Environment: OS || Case 13}]
Instruction: Modify the Nginx configuration file at /etc/nginx/sites-available/default with the following changes using sed: 1) Replace all "http://" with "https://", 2) Change the listen port from 80 to 8080, 3) Set server\_name to "myapp.com", 4) Disable server\_tokens by setting it to "off", 5) Add "client\_max\_body\_size 20M;" after the server block opening, 6) Insert "keepalive\_timeout 65;" before the location block, 7) Update the proxy\_pass directive to use HTTPS, 8) Add a comment "\# Security update" before server\_name, and 9) Ensure all changes are made in-place. Preserve the original configuration structure.

\bigskip

\begin{sloppypar}{\ttfamily Ground Truth Action: \textcolor{blue} {"mkdir -p /etc/nginx/sites-available \&\& printf\\ "server \{\textbackslash\textbackslash\textbackslash\textbackslash n    listen 80;\textbackslash\textbackslash\textbackslash\textbackslash n    server\_name example.com;\textbackslash\textbackslash\textbackslash\textbackslash n    server\_\\tokens on;\textbackslash\textbackslash\textbackslash\textbackslash n    location / \{\textbackslash\textbackslash\textbackslash\textbackslash n        proxy\_pass http://backend;\textbackslash\textbackslash\textbackslash\textbackslash n    \}\textbackslash\\\textbackslash\textbackslash\textbackslash n\}\textbackslash\textbackslash\textbackslash\textbackslash n" > /etc/nginx/sites-available/default"}}\end{sloppypar}

\bigskip

Skill List: ["cd", "cp", "echo", "sed"]

\end{tcolorbox}

% SSSSSSSSSSSSSSSSSSSSSSSSSSSSSSSSSSSSSSSSSSSSSSSSSSSSSSSSSSSSSSSSSSSSSSSSS

\begin{tcolorbox}[prompt_box_style, title={Environment: OS || Case 14}]
Instruction: 1. Create a group "testgroup", add "testuser" to it, set group ownership recursively on "/var/www/project" to "testgroup", set directory permissions to 775, file permissions to 660, and enable setgid on the project directory.

\bigskip

\begin{sloppypar}{\ttfamily Ground Truth Action: \textcolor{blue} {"useradd -m testuser \&\& mkdir -p /var/www/proje\\ct/{assets,logs} \&\& touch /var/www/project/assets/image1.jpg /var/www/p\\roject/logs/debug.log \&\& chmod 755 /var/www/project"}}\end{sloppypar}

\bigskip

Skill List: ["addgroup", "chgrp", "chmod", "find", "usermod"]

\end{tcolorbox}

\begin{tcolorbox}[prompt_box_style, title={Environment: OS || Case 15}]
Instruction: Remove /tmp/cleanup/file1.tmp and /tmp/cleanup/dir1/file2.tmp, delete all empty directories within /tmp/cleanup, and ensure the /tmp/cleanup directory exists.

\bigskip

\begin{sloppypar}{\ttfamily Ground Truth Action: \textcolor{blue} {"mkdir -p /tmp/cleanup/dir1 /tmp/cleanup/dir2 \&\& touch /tmp/cleanup/file1.tmp /tmp/cleanup/dir1/file2.tmp /tmp/cleanup\\/dir2/file3.log"}}\end{sloppypar}

\bigskip

Skill List: ["echo", "exit", "find", "ls", "mkdir", "rm"]

\end{tcolorbox}

% SSSSSSSSSSSSSSSSSSSSSSSSSSSSSSSSSSSSSSSSSSSSSSSSSSSSSSSSSSSSSSSSSSSSSSSSS
\begin{tcolorbox}[prompt_box_style, title={Environment: OS || Case 16}]
Instruction: Create a group "devteam", add users "user1", "user2", and "user3" to it, create a directory "/var/devteam\_projects" accessible only to the group, and generate a log file with group details.

\bigskip

\begin{sloppypar}{\ttfamily Ground Truth Action: \textcolor{blue} {"useradd -m user1 \&\& useradd -m user2 \&\& useradd -m user3"}}\end{sloppypar}

\bigskip

Skill List: ["addgroup", "chgrp", "chmod", "echo", "mkdir", "tee", "touch", "usermod"]

\end{tcolorbox}

\begin{tcolorbox}[prompt_box_style, title={Environment: OS || Case 17}]
Instruction: Create a backup directory "/backup", copy "/source/test.txt" to both "/backup" and "/destination", append "Backup completed" to "/logs/backup.log", set permissions of "/backup/test.txt" to 644, and ensure "/destination/test.txt" is owned by root.

\bigskip

\begin{sloppypar}{\ttfamily Ground Truth Action: \textcolor{blue} {"mkdir -p /source /destination /logs \&\& echo "Sa\\mple data" > /source/test.txt \&\& touch /logs/backup.log"}}\end{sloppypar}

\bigskip

Skill List: ["chmod", "chown", "cp", "echo", "find", "grep", "ls", "mkdir", "tee"]

\end{tcolorbox}

% SSSSSSSSSSSSSSSSSSSSSSSSSSSSSSSSSSSSSSSSSSSSSSSSSSSSSSSSSSSSSSSSSSSSSSSSS

\begin{tcolorbox}[prompt_box_style, title={Environment: OS || Case 18}]
Instruction: 1. Add "/bin/zsh" to /etc/shells. 2. Change "testuser" login shell to /bin/zsh. 3. Create group "devteam". 4. Add "testuser" to "devteam". 5. Create /shared directory with permissions 770 owned by "devteam" group.

\bigskip

\begin{sloppypar}{\ttfamily Ground Truth Action: \textcolor{blue} {"useradd -m -s /bin/bash testuser \&\& sed -i "/\textbackslash\textbackslash\\\textbackslash\textbackslash/bin\textbackslash\textbackslash\textbackslash\textbackslash/zsh/d" /etc/shells"}}\end{sloppypar}

\bigskip

Skill List: ["addgroup", "chage", "chgrp", "chmod", "chown", "chsh", "echo", "grep", "mkdir", "tee", "touch", "usermod"]

\end{tcolorbox}

\subsection{Knowledge Graph Samples}

% SSSSSSSSSSSSSSSSSSSSSSSSSSSSSSSSSSSSSSSSSSSSSSSSSSSSSSSSSSSSSSSSSSSSSSSSS
%2
\begin{tcolorbox}[prompt_box_style, title={Environment: KG || Case 19}]
Instruction: which institution has national wine centre of australia?, Entities: ["National Wine Centre of Australia"

\bigskip

\begin{sloppypar}{\ttfamily Ground Truth Action: \textcolor{blue} {["get\_relations(m.03hd1z)", "get\_neighbors(m.03\\hd1z,education.educational\_institution.parent\_institution)"]}}\end{sloppypar}

% Answer\_list: \textcolor{blue} {["m.01dbns"]}
\bigskip

Skill List: ["get\_neighbors"]

\end{tcolorbox}

% SSSSSSSSSSSSSSSSSSSSSSSSSSSSSSSSSSSSSSSSSSSSSSSSSSSSSSSSSSSSSSSSSSSSSSSSS
%3
\begin{tcolorbox}[prompt_box_style, title={Environment: KG || Case 20}]
Instruction: how many characters appear on the cover of tintin in the land of the soviets?

\bigskip

\begin{sloppypar}{\ttfamily Ground Truth Action: \textcolor{blue} {["get\_relations(m.02ll5h)", "get\_neighbors(m.02\\ll5h,comic\_books.comic\_book\_issue.characters\_on\_cover)", "count(\#0)"]}}\end{sloppypar}

% Answer\_list: \textcolor{blue} {["4"]}
\bigskip

Skill List: ["count", "get\_neighbors"]

\end{tcolorbox}

% SSSSSSSSSSSSSSSSSSSSSSSSSSSSSSSSSSSSSSSSSSSSSSSSSSSSSSSSSSSSSSSSSSSSSSSSS
%4
\begin{tcolorbox}[prompt_box_style, title={Environment: KG || Case 21}]
Instruction: Question: which short story of the sacred band of stepsons universe universe is know to have the earliest copyright date?, Entities: ["The Sacred Band of Stepsons universe"]

\bigskip

\begin{sloppypar}{\ttfamily Ground Truth Action: \textcolor{blue} {["get\_relations(m.0ch8hcq)", 
 "get\_neighbors(m.0c\\h8hcq,fictional\_universe.fictional\_universe.works\_set\_here)", "get\_att\\ributes(\#0)",  "argmin(\#0,book.written\_work.copyright\_date)"]}}\end{sloppypar}

% Answer\_list: \textcolor{blue} {["m.0cqztld"]}
\bigskip

Skill List: ["argmin", "get\_neighbors"]

\end{tcolorbox}

% SSSSSSSSSSSSSSSSSSSSSSSSSSSSSSSSSSSSSSSSSSSSSSSSSSSSSSSSSSSSSSSSSSSSSSSSS
%5
\begin{tcolorbox}[prompt_box_style, title={Environment: KG || Case 22}]
Instruction: which fictional character produced by marv wolfman did trevor von eeden create?, Entities: ["Trevor Von Eeden", "Marv Wolfman"]

\bigskip

\begin{sloppypar}{\ttfamily Ground Truth Action: \textcolor{blue} {["get\_relations(m.0279q8n)", "get\_neighbors(m.02\\79q8n,fictional\_universe.fictional\_character\_creator.fictional\_charact\\ers\_created)", "get\_relations(m.02gn9g)", "get\_neighbors(m.02gn9g,fict\\ional\_universe.fictional\_character\_creator.fictional\_characters\_create\\d)", "intersection(\#0,\#1)"]}}\end{sloppypar}

% Answer\_list: \textcolor{blue} {["4"]}
\bigskip

Skill List: ["get\_neighbors", "intersection"]

\end{tcolorbox}

% SSSSSSSSSSSSSSSSSSSSSSSSSSSSSSSSSSSSSSSSSSSSSSSSSSSSSSSSSSSSSSSSSSSSSSSSS
%6
\begin{tcolorbox}[prompt_box_style, title={Environment: KG || Case 23}]
Instruction: what was the most recently formed cyclone in the same category as hurricane dolly?, Entities: ["Hurricane Dolly"]

\bigskip

\begin{sloppypar}{\ttfamily Ground Truth Action: \textcolor{blue} {["get\_relations(m.04dn799)", "get\_neighbors(m.04\\dn799,meteorology.tropical\_cyclone.category)", "get\_relations(\#0)","ge\\t\_neighbors(\#0,meteorology.tropical\_cyclone\_category.
tropical\_cyclon\\es)", "get\_attributes(\#1)", "argmax(\#1,meteorology.tropical\_cyclone.fo\\rmed)"]}}\end{sloppypar}

% Answer\_list: \textcolor{blue} {["m.011v6mk8"]}
\bigskip

Skill List: ["argmax", "get\_neighbors"]

\end{tcolorbox}

% SSSSSSSSSSSSSSSSSSSSSSSSSSSSSSSSSSSSSSSSSSSSSSSSSSSSSSSSSSSSSSSSSSSSSSSSS
%7
\begin{tcolorbox}[prompt_box_style, title={Environment: KG || Case 24}]
Instruction: how much content about talk radio is produced by the person that produces weekend edition sunday?, Entities: ["Talk radio", "Weekend Edition Sunday"]

\bigskip

\begin{sloppypar}{\ttfamily Ground Truth Action: \textcolor{blue} {["get\_relations(m.07dn1)", "get\_neighbors(m.07dn\\1,broadcast.genre.content)", "get\_relations(m.0t4t10s)", "get\_neighbor\\s(m.0t4t10s,broadcast.content.producer)", "get\_relations(\#1)", "get\_ne\\ighbors(\#1,broadcast.producer.produces)", "intersection(\#0,\#2)", "count(\#3)"]}}\end{sloppypar}

% Answer\_list: \textcolor{blue} {["4"]}
\bigskip

Skill List: ["count", "get\_neighbors", "intersection"]

\end{tcolorbox}

% SSSSSSSSSSSSSSSSSSSSSSSSSSSSSSSSSSSSSSSSSSSSSSSSSSSSSSSSSSSSSSSSSSSSSSSSS
%8
\begin{tcolorbox}[prompt_box_style, title={Environment: KG || Case 25}]
Instruction: what semi-firm textured cheese is made from the products of lamb and goat?, Entities: ["lamb", "Goat", "semi-firm"]

\bigskip

\begin{sloppypar}{\ttfamily Ground Truth Action: \textcolor{blue} {["get\_relations(m.07bgp)", "get\_neighbors(m.07bg\\p,food.cheese\_milk\_source.cheeses)", "get\_relations(m.03fwl)", "get\_ne\\ighbors(m.03fwl,food.cheese\_milk\_source.cheeses)", "get\_relations(m.02\\h82t0)", "get\_neighbors(m.02h82t0,food.cheese\_texture.cheeses)", "inte\\rsection(\#1,\#2)", "intersection(\#0,\#3)"]}}\end{sloppypar}

% Answer\_list: \textcolor{blue} {["m.02h8b9t"]}
\bigskip

Skill List: ["get\_neighbors", "intersection"]

\end{tcolorbox}

% SSSSSSSSSSSSSSSSSSSSSSSSSSSSSSSSSSSSSSSSSSSSSSSSSSSSSSSSSSSSSSSSSSSSSSSSS
%9
\begin{tcolorbox}[prompt_box_style, title={Environment: KG || Case 26}]
Instruction: is a model of opel super 6 related to a eagle talon?, Entities: ["Opel Super 6", "Eagle Talon"]

\bigskip

\begin{sloppypar}{\ttfamily Ground Truth Action: \textcolor{blue} {["get\_relations(m.0gdk70)", "get\_neighbors(m.0gd\\k70,automotive.model.automotive\_class)", "get\_relations(\#0)", "get\_nei\\ghbors(\#0,automotive.automotive\_class.examples)", "get\_relations(m.02p\\04r)","get\_neighbors(m.02p04r,automotive.model.related\_models)", "get\_\\relations(\#2)", "get\_neighbors(\#2,automotive.similar\_automobile\_models\\.related\_model)", "intersection(\#1,\#3)"]}}\end{sloppypar}

% Answer\_list: \textcolor{blue} {["4"]}
\bigskip

Skill List: ["get\_neighbors", "intersection"]

\end{tcolorbox}
\section{Error Case Analysis}
\label{sec:appendix_error_cases}

To better understand the failure modes of LLM agents in \LifelongAgentBench, we analyze all task outcomes using two orthogonal attributes: \texttt{evaluation\_outcome} (correct vs.\ incorrect) and \texttt{sample\_status} (detailed agent or system status). This provides a fine-grained categorization of agent behavior and helps identify major bottlenecks.

\paragraph{Successful Cases (\texttt{evaluation\_outcome == "correct"})} We observe two typical success patterns:
\begin{itemize}
    \item \texttt{sample\_status == "completed"}: the agent explicitly submits a valid final answer following the required format (e.g., correctly querying customer feedback with SQL and submitting results).
    \item \texttt{sample\_status == "task\_limit\_reached"}: the agent stops without explicitly committing, but the final database state satisfies the evaluation criteria (e.g., correctly updating records within the step limit).
\end{itemize}

\paragraph{Error Cases (\texttt{evaluation\_outcome == "incorrect"})} We identify four common failure modes:
\begin{itemize}
    \item \texttt{sample\_status == "completed"}: the agent explicitly commits an incorrect answer despite following the required format (e.g., submitting an incomplete result in a multi-step SQL query).
    \item \texttt{sample\_status == "task\_limit\_reached"}: the agent fails to converge within the interaction limit, often due to redundant or looping operations (e.g., repeatedly adjusting query logic without submitting).
    \item \texttt{sample\_status == "agent\_validation\_failed"}: the agent violates output constraints, such as missing required keywords or formatting (e.g., omitting \texttt{Action: Answer} tag).
    \item \texttt{sample\_status == "agent\_context\_limit"}: the agent exceeds the LLM context window, especially in environments requiring long reasoning chains or large intermediate outputs.
\end{itemize}

\paragraph{System-Level Failures} We also observe rare non-agent errors:
\begin{itemize}
    \item \texttt{task\_environment\_error}: task-side infrastructure failures (e.g., Docker crashes).
    \item \texttt{task\_unknown\_error}: unknown exceptions in task execution.
    \item \texttt{agent\_out\_of\_memory}: GPU memory exhaustion on large models.
    \item \texttt{agent\_unknown\_error}: unexpected internal failures within the agent framework.
\end{itemize}

This taxonomy highlights that even when equipped with powerful LLMs, agents remain vulnerable to subtle evaluation failures, unstable reasoning loops, and external infrastructure issues. Our benchmark thus provides valuable diagnostic insights for developing more robust lifelong learning agents.

\subsection{Task Completed}
\label{sec:appendix_error_cases:complete}

\subsubsection{Error Case 1}

\textbf{Task Description:}
The agent was instructed to query the database to retrieve insurance policy details, specifically the policy IDs, holder names, and policy types from the \texttt{insurance\_policies} table. The results needed to be ordered by \texttt{effective\_date} from the most recent to the oldest, skipping the first five entries and returning the subsequent ten entries.

\textbf{Agent Behavior:}
Initially, the agent executed the following SQL query:
\begin{verbatim}
SELECT policy_id, holder_name, policy_type 
FROM insurance_policies 
ORDER BY effective_date DESC 
LIMIT 4 OFFSET 5;
\end{verbatim}

This query was incorrect due to mistakenly using \texttt{LIMIT 4} instead of the required \texttt{LIMIT 10}. Consequently, the query returned only four entries instead of the intended ten.

Recognizing the mistake after receiving the results, the agent correctly identified the error in the limit clause. It then adjusted and executed the corrected query:
\begin{verbatim}
SELECT policy_id, holder_name, policy_type 
FROM insurance_policies 
ORDER BY effective_date DESC 
LIMIT 10 OFFSET 5;
\end{verbatim}

The updated query returned the correct number of rows (ten entries), satisfying the task requirements.

\textbf{Reason for Failure:}
Despite eventually correcting the SQL query, the task outcome was marked as incorrect (status \texttt{completed} but evaluation incorrect). The failure arose from the initial misunderstanding and execution of an inaccurate query, causing unnecessary computational resources (an extra interaction) and illustrating unstable reasoning. Such errors highlight difficulties in accurately translating natural language constraints (e.g., ``starting from the 6th entry and limited to 10 entries'') into correct SQL clauses (\texttt{OFFSET} and \texttt{LIMIT}).

\textbf{Implications and Insights:}
\begin{itemize}
    \item \textbf{Unstable Reasoning:} The agent correctly explained the logic but initially failed to accurately apply it, demonstrating inconsistency between stated understanding and execution.
    \item \textbf{Inefficiency:} The error required additional agent-system interactions, revealing suboptimal reasoning efficiency.
    \item \textbf{Potential Solutions:} Future improvements could involve explicit verification of each SQL query component against the original natural language instruction or introducing intermediate verification steps before query execution.
\end{itemize}

This case exemplifies a common issue where an agent's conceptual interpretation of instructions is correct but its initial execution is flawed, underscoring the necessity for enhanced accuracy in natural language-to-code translation.

\begin{tcolorbox}[prompt_box_style, title={Error Case 1 || Environment: DB || Error Type: completed}]
User: I will ask you a question, then you should help me operate a MySQL database with SQL to answer the question.\textbackslash nYou have to explain the problem and your solution to me and write down your thoughts.\textbackslash nAfter thinking and explaining thoroughly, every round you can choose one of the two actions: Operation or Answer.\textbackslash n\textbackslash nTo do operation, the format should be like this:\textbackslash nAction: Operation\textbackslash n"sql\textbackslash nSELECT \* FROM table WHERE condition;\textbackslash n"\textbackslash nYou MUST put SQL in markdown format without any other comments. Your SQL should be in one line.\textbackslash nI will use \"Action: Operation\" literally to match your SQL.\textbackslash nEvery time you can only execute one SQL statement. I will only execute the statement in the first SQL code block. Every time you write a SQL, I will execute it for you and give you the output.\textbackslash nIf the SQL is not executed successfully, the response will be the error message.\textbackslash nOtherwise, the response will be the raw MySQL response.\textbackslash nFor SELECT queries, the response will be the result of the query, such as [(1, "John Doe", "HR"), (2, "Jane Smith", "IT"), ...], where each tuple represents a row and the elements are the values of the columns in the row.\textbackslash nFor SQL such as INSERT, UPDATE, and DELETE, the response will be an empty list [] indicating that the SQL was executed successfully.\textbackslash n\textbackslash nIf you have obtain the answer by interacting with the database, and you MUST commit your final answer using the format like this:\textbackslash nAction: Answer\textbackslash nFinal Answer: [(1, "John Doe", "HR"), (2, "Jane Smith", "IT"), ...]\textbackslash nDO NOT write this pattern unless you are sure about your answer. I expect an accurate and correct answer.\textbackslash nYour answer should be accurate. Your answer must be exactly the same as the correct answer.\textbackslash nIf the question is about modifying the database, then after done operation, your answer field can be anything.\textbackslash nIf the question is about querying the database, then after done operation, your answer field should be the result of the query.\textbackslash nWe note that the column names will not be displayed in the result, and you need to ensure both the orders of the columns and rows are correct.\textbackslash nIf your response cannot match any pattern I mentioned earlier, you will be judged as FAIL immediately.\textbackslash nOnce you commit your answer or the number of rounds reaches 3, the task will be finished and the system will judge whether you pass the task or not.\textbackslash n\textbackslash nNow, I will give you the question that you need to solve.

\bigskip

Agent: OK.

\bigskip

User: What are the policy IDs, holder names, and policy types of the insurance policies? Return the results ordered by effective date from most recent to oldest, starting from the 6th entry and limited to 10 entries.\textbackslash nThe name of this table is insurance\_policies, and the headers of this table are policy\_id, holder\_name, policy\_type, effective\_date, premium\_amount, coverage\_limit.

\bigskip

Agent: To solve this problem, we need to first understand the requirements. We need to return the policy IDs, holder names, and policy types of the insurance policies. We also need to order the results by effective date from most recent to oldest, starting from the 6th entry and limited to 10 entries.\textbackslash n\textbackslash nHere's my thought process:\textbackslash n\textbackslash n1. We need to select the policy ID, holder name, and policy type from the insurance\_policies table.\textbackslash n2. We need to order the results by effective date in descending order (most recent to oldest).\textbackslash n3. We need to start the result set from the 6th entry.\textbackslash n4. We need to limit the result set to 10 entries.\textbackslash n\textbackslash nBased on this, I will choose the \"Operation\" action to execute the SQL query.\textbackslash n\textbackslash nAction: Operation\textbackslash n"sql\textbackslash nSELECT policy\_id, holder\_name, policy\_type \textbackslash nFROM insurance\_policies \textbackslash nORDER BY effective\_date DESC \textbackslash nLIMIT 4 OFFSET 5;\textbackslash n"\textbackslash nPlease execute this SQL query.

\bigskip

User: [(6, "Alice Brown", "Health"), (7, "Bob Green", "Auto"), (8, "Charlie White", "Life"), (9, "Diana Black", "Home")]

\bigskip

Agent: The query I executed returned the first 4 rows from the 6th row onwards, ordered by effective date in descending order. However, this is not what we want. We want to return the first 10 rows from the 6th row onwards, ordered by effective date in descending order.\textbackslash n\textbackslash nTo achieve this, we need to modify the query to return the first 10 rows from the 6th row onwards, ordered by effective date in descending order.\textbackslash n\textbackslash nHere's my thought process:\textbackslash n\textbackslash n1. We need to select the policy ID, holder name, and policy type from the insurance\_policies table.\textbackslash n2. We need to order the results by effective date in descending order (most recent to oldest).\textbackslash n3. We need to start the result set from the 6th entry.\textbackslash n4. We need to limit the result set to 10 entries.\textbackslash n\textbackslash nBased on this, I will choose the \"Operation\" action to execute the SQL query.\textbackslash n\textbackslash nAction: Operation\textbackslash n"sql\textbackslash nSELECT policy\_id, holder\_name, policy\_type \textbackslash nFROM insurance\_policies \textbackslash nORDER BY effective\_date DESC \textbackslash nLIMIT 10 OFFSET 5;\textbackslash n"

\bigskip

User: [(6, "Alice Brown", "Health"), (7, "Bob Green", "Auto"), (8, "Charlie White", "Life"), (9, "Diana Black", "Home"), (10, "Eve Gray", "Travel"), (11, "Alice Brown", "Health"), (12, "Bob Green", "Auto"), (13, "Charlie White", "Life"), (14, "Diana Black", "Home"), (15, "Eve Gray", "Travel")]

\bigskip

Agent: The query I executed returned the first 10 rows from the 6th row onwards, ordered by effective date in descending order. This is what we want.\textbackslash n\textbackslash nNow, we can commit the final answer.\textbackslash n\textbackslash nAction: Answer\textbackslash nFinal Answer: [(6, "Alice Brown", "Health"), (7, "Bob Green", "Auto"), (8, "Charlie White", "Life"), (9, "Diana Black", "Home"), (10, "Eve Gray", "Travel"), (11, "Alice Brown", "Health"), (12, "Bob Green", "Auto"), (13, "Charlie White", "Life"), (14, "Diana Black", "Home"), (15, "Eve Gray", "Travel")]

\end{tcolorbox}

\subsubsection{Error Case 2}

\textbf{Task Description:}
The agent was tasked with querying the database to identify regions from the \texttt{customer\_usage} table that have a total data usage exceeding 1000 and an average call duration under 300. The results needed to include the region, total data usage, and average call duration, grouped by region, and limited to 5 entries starting from the third entry.

\textbf{Agent Behavior:}
Initially, the agent executed the following SQL query:
\begin{verbatim}
SELECT region, SUM(data_usage) AS total_data_usage, 
AVG(call_duration) AS average_call_duration 
FROM customer_usage 
GROUP BY region 
HAVING SUM(data_usage) > 1000 AND AVG(call_duration) < 300 
ORDER BY total_data_usage DESC 
LIMIT 2, 5;
\end{verbatim}

However, the agent misunderstood the usage of \texttt{LIMIT} and \texttt{OFFSET} syntax in SQL, using \texttt{LIMIT 2, 5} instead of the correct \texttt{LIMIT 5 OFFSET 2} syntax. After executing the initial query, the agent attempted to correct its misunderstanding by issuing a second query:
\begin{verbatim}
SELECT region, SUM(data_usage) AS total_data_usage, 
AVG(call_duration) AS average_call_duration 
FROM customer_usage 
GROUP BY region 
HAVING SUM(data_usage) > 1000 AND AVG(call_duration) < 300 
ORDER BY total_data_usage DESC 
LIMIT 3, 5;
\end{verbatim}

Yet, this second query also incorrectly interpreted the desired offset and limit, resulting in incorrect indexing.

\textbf{Reason for Failure:}
The agent's failure stemmed from confusion regarding the SQL syntax for specifying offsets and limits. Instead of correctly applying \texttt{OFFSET 2 LIMIT 5} to start from the third entry, the agent incorrectly applied \texttt{LIMIT 2,5} and subsequently \texttt{LIMIT 3,5}, leading to incorrect or shifted result sets.

\textbf{Implications and Insights:}
\begin{itemize}
    \item \textbf{Syntax Misinterpretation:} The agent demonstrated difficulty accurately applying SQL syntax for offset and limit clauses, highlighting issues in translating natural language queries to precise SQL statements.
    \item \textbf{Persistent Mistake:} Despite attempting to correct its mistake, the agent continued to misuse the syntax, illustrating repeated misunderstanding rather than one-off oversight.
    \item \textbf{Potential Solutions:} Improved SQL syntax awareness or explicit syntactic verification mechanisms during intermediate reasoning steps could significantly enhance accuracy.
\end{itemize}

This case highlights the importance of reinforcing SQL syntax knowledge and accuracy during the reasoning and execution phases for LLM agents.

\begin{tcolorbox}[prompt_box_style, title={Error Case 2 || Environment: DB || Error Type: completed}]

User: I will ask you a question, then you should help me operate a MySQL database with SQL to answer the question.\textbackslash nYou have to explain the problem and your solution to me and write down your thoughts.\textbackslash nAfter thinking and explaining thoroughly, every round you can choose one of the two actions: Operation or Answer.\textbackslash n\textbackslash nTo do operation, the format should be like this:\textbackslash nAction: Operation\textbackslash n"sql\textbackslash nSELECT \* FROM table WHERE condition;\textbackslash n"\textbackslash nYou MUST put SQL in markdown format without any other comments. Your SQL should be in one line.\textbackslash nI will use "Action: Operation" literally to match your SQL.\textbackslash nEvery time you can only execute one SQL statement. I will only execute the statement in the first SQL code block. Every time you write a SQL, I will execute it for you and give you the output.\textbackslash nIf the SQL is not executed successfully, the response will be the error message.\textbackslash nOtherwise, the response will be the raw MySQL response.\textbackslash nFor SELECT queries, the response will be the result of the query, such as [(1, "John Doe", "HR"), (2, "Jane Smith", "IT"), ...], where each tuple represents a row and the elements are the values of the columns in the row.\textbackslash nFor SQL such as INSERT, UPDATE, and DELETE, the response will be an empty list [] indicating that the SQL was executed successfully.\textbackslash n\textbackslash nIf you have obtain the answer by interacting with the database, and you MUST commit your final answer using the format like this:\textbackslash nAction: Answer\textbackslash nFinal Answer: [(1, "John Doe", "HR"), (2, "Jane Smith", "IT"), ...]\textbackslash nDO NOT write this pattern unless you are sure about your answer. I expect an accurate and correct answer.\textbackslash nYour answer should be accurate. Your answer must be exactly the same as the correct answer.\textbackslash nIf the question is about modifying the database, then after done operation, your answer field can be anything.\textbackslash nIf the question is about querying the database, then after done operation, your answer field should be the result of the query.\textbackslash nWe note that the column names will not be displayed in the result, and you need to ensure both the orders of the columns and rows are correct.\textbackslash nIf your response cannot match any pattern I mentioned earlier, you will be judged as FAIL immediately.\textbackslash nOnce you commit your answer or the number of rounds reaches 3, the task will be finished and the system will judge whether you pass the task or not.\textbackslash n\textbackslash nNow, I will give you the question that you need to solve.

\bigskip

Agent: OK.

\bigskip

User: Which regions have a total data usage exceeding 1000 and an average call duration under 300? Return the region, total data usage, and average call duration, grouped by region. Limit the results to 5 entries starting from the third entry.\textbackslash nThe name of this table is customer\_usage, and the headers of this table are customer\_id, region, data\_usage, call\_duration, month.

\bigskip

Agent: To solve this problem, we need to first calculate the total data usage and average call duration for each region. Then, we can filter the results to only include regions with a total data usage exceeding 1000 and an average call duration under 300. Finally, we can group the results by region and limit them to 5 entries starting from the third entry.\textbackslash n\textbackslash nAction: Operation\textbackslash n"sql\textbackslash nSELECT region, SUM(data\_usage) AS total\_data\_usage, AVG(call\_duration) AS average\_call\_duration \textbackslash nFROM customer\_usage \textbackslash nGROUP BY region \textbackslash nHAVING SUM(data\_usage) > 1000 AND AVG(call\_duration) < 300 \textbackslash nORDER BY total\_data\_usage DESC \textbackslash nLIMIT 2, 5;\textbackslash n"

\bigskip

User: [("South", Decimal("1200"), Decimal("250.0000")), ("Central", Decimal("1200"), Decimal("280.0000")), ("EastCoast", Decimal("1100"), Decimal("290.0000")), ("WestCoast", Decimal("1100"), Decimal("200.0000")), ("Southwest", Decimal("1050"), Decimal("290.0000"))]

\bigskip

Agent: The results show the top 5 regions with a total data usage exceeding 1000 and an average call duration under 300. However, we need to exclude the first two entries since we want to start from the third entry.\textbackslash n\textbackslash nAction: Operation\textbackslash n"sql\textbackslash nSELECT region, SUM(data\_usage) AS total\_data\_usage, AVG(call\_duration) AS average\_call\_duration \textbackslash nFROM customer\_usage \textbackslash nGROUP BY region \textbackslash nHAVING SUM(data\_usage) > 1000 AND AVG(call\_duration) < 300 \textbackslash nORDER BY total\_data\_usage DESC \textbackslash nLIMIT 3, 5;\textbackslash n"

\bigskip

User: [("Central", Decimal("1200"), Decimal("280.0000")), ("EastCoast", Decimal("1100"), Decimal("290.0000")), ("WestCoast", Decimal("1100"), Decimal("200.0000")), ("Southwest", Decimal("1050"), Decimal("290.0000"))]

\bigskip

Agent: The results now show the top 5 regions with a total data usage exceeding 1000 and an average call duration under 300, starting from the third entry.\textbackslash n\textbackslash nAction: Answer\textbackslash nFinal Answer: [("Central", Decimal("1200"), Decimal("280.0000")), ("EastCoast", Decimal("1100"), Decimal("290.0000")), ("WestCoast", Decimal("1100"), Decimal("200.0000")), ("Southwest", Decimal("1050"), Decimal("290.0000"))]

\bigskip

\end{tcolorbox}

\subsection{Task Limited Reached}
\label{sec:appendix_error_cases:task_limit_reached}

\subsubsection{Error Case 3}

\textbf{Task Description:}
The agent was tasked with updating the hourly rates for staff members in Security roles from the \texttt{staff\_shifts} table. Specifically, it was required to increase the hourly rate by 15\% for Security roles where the hours worked exceeded 30, and the current hourly rate was below the average hourly rate of all Security roles.

\textbf{Agent Behavior:}
Initially, the agent correctly calculated the average hourly rate for Security roles using the following SQL query:
\begin{verbatim}
SELECT AVG(hourly_rate) AS avg_rate 
FROM staff_shifts 
WHERE role = "Security";
\end{verbatim}

Upon identifying the average hourly rate (20.5), the agent correctly queried the eligible entries to be updated:
\begin{verbatim}
SELECT * FROM staff_shifts 
WHERE role = "Security" AND hours_worked > 30 AND hourly_rate < 20.5;
\end{verbatim}

Having successfully retrieved the targeted entries, the agent executed the update query:
\begin{verbatim}
UPDATE staff_shifts 
SET hourly_rate = hourly_rate * 1.15 
WHERE role = "Security" AND hours_worked > 30 
AND hourly_rate < 20.5;
\end{verbatim}

Finally, the agent attempted to verify the successful execution of the update by recalculating the average hourly rate. However, at this point, the agent reached the maximum allowed interaction steps, failing to explicitly commit a final answer.

\textbf{Reason for Failure:}
The agent failed due to exceeding the maximum interaction limit without committing a final answer explicitly. Although the operations performed were logically sound and correct, the agent did not conclude the task by confirming the success explicitly within the allowed steps.

\textbf{Implications and Insights:}
\begin{itemize}
    \item \textbf{Lack of Task Commitment:} The agent demonstrated proper initial reasoning and executed correct SQL statements. However, it failed to finalize the task promptly, reflecting inadequate task closure behavior.
    \item \textbf{Interaction Efficiency:} This case highlights the necessity for the agent to strategically plan and optimize interactions, especially when task constraints (such as interaction limits) are strict.
    \item \textbf{Potential Solutions:} Introducing intermediate commitment strategies or improved planning to manage interaction limits effectively could enhance overall task performance and success rates.
\end{itemize}

This case emphasizes the importance of explicit task completion strategies and efficient interaction management in multi-step tasks for LLM agents.

\begin{tcolorbox}[prompt_box_style, title={Error Case 3 || Environment: DB || Error Type: task limit reached}]
User: I will ask you a question, then you should help me operate a MySQL database with SQL to answer the question.\textbackslash nYou have to explain the problem and your solution to me and write down your thoughts.\textbackslash nAfter thinking and explaining thoroughly, every round you can choose one of the two actions: Operation or Answer.\textbackslash n\textbackslash nTo do operation, the format should be like this:\textbackslash nAction: Operation\textbackslash n"sql\textbackslash nSELECT \* FROM table WHERE condition;\textbackslash n"\textbackslash nYou MUST put SQL in markdown format without any other comments. Your SQL should be in one line.\textbackslash nI will use "Action: Operation" literally to match your SQL.\textbackslash nEvery time you can only execute one SQL statement. I will only execute the statement in the first SQL code block. Every time you write a SQL, I will execute it for you and give you the output.\textbackslash nIf the SQL is not executed successfully, the response will be the error message.\textbackslash nOtherwise, the response will be the raw MySQL response.\textbackslash nFor SELECT queries, the response will be the result of the query, such as [(1, "John Doe", "HR"), (2, "Jane Smith", "IT"), ...], where each tuple represents a row and the elements are the values of the columns in the row.\textbackslash nFor SQL such as INSERT, UPDATE, and DELETE, the response will be an empty list [] indicating that the SQL was executed successfully.\textbackslash n\textbackslash nIf you have obtain the answer by interacting with the database, and you MUST commit your final answer using the format like this:\textbackslash nAction: Answer\textbackslash nFinal Answer: [(1, "John Doe", "HR"), (2, "Jane Smith", "IT"), ...]\textbackslash nDO NOT write this pattern unless you are sure about your answer. I expect an accurate and correct answer.\textbackslash nYour answer should be accurate. Your answer must be exactly the same as the correct answer.\textbackslash nIf the question is about modifying the database, then after done operation, your answer field can be anything.\textbackslash nIf the question is about querying the database, then after done operation, your answer field should be the result of the query.\textbackslash nWe note that the column names will not be displayed in the result, and you need to ensure both the orders of the columns and rows are correct.\textbackslash nIf your response cannot match any pattern I mentioned earlier, you will be judged as FAIL immediately.\textbackslash nOnce you commit your answer or the number of rounds reaches 3, the task will be finished and the system will judge whether you pass the task or not.\textbackslash n\textbackslash nNow, I will give you the question that you need to solve.

\bigskip

Agent: OK.

\bigskip

User: Update the hourly rate by increasing it by 15\% for Security roles where hours worked exceed 30 and their current hourly rate is below the average hourly rate of all Security roles.\textbackslash nThe name of this table is staff\_shifts, and the headers of this table are shift\_id, staff\_name, role, shift\_date, hours\_worked, hourly\_rate, total\_pay.

\bigskip

Agent: To solve this problem, we need to first calculate the average hourly rate of all Security roles. Then, we need to find the Security roles where hours worked exceed 30 and their current hourly rate is below the average hourly rate of all Security roles. Finally, we can update the hourly rate of these roles by increasing it by 15\%.\textbackslash n\textbackslash nAction: Operation\textbackslash n"sql\textbackslash nSELECT AVG(hourly\_rate) AS avg\_rate FROM staff\_shifts WHERE role = "Security";\textbackslash n"

\bigskip

User: [(Decimal("20.5000"),)]

\bigskip

Agent: The average hourly rate of all Security roles is 20.5. Now, we need to find the Security roles where hours worked exceed 30 and their current hourly rate is below 20.5.\textbackslash n\textbackslash nAction: Operation\textbackslash n"sql\textbackslash nSELECT \* FROM staff\_shifts WHERE role = "Security" AND hours\_worked > 30 AND hourly\_rate < 20.5;\textbackslash n"

\bigskip

User: [(1, "Mike Johnson", "Security", "2023-01-01", 35, 18, 630), (2, "Anna Lee", "Security", "2023-01-02", 32, 20, 640), (5, "Evan Smith", "Security", "2023-01-05", 31, 19, 589), (7, "George Adams", "Security", "2023-01-07", 33, 17, 561), (10, "Jenna Hall", "Security", "2023-01-10", 34, 16, 544)]

\bigskip

Agent: We have found the Security roles that meet the conditions. Now, we can update the hourly rate of these roles by increasing it by 15\%.\textbackslash n\textbackslash nAction: Operation\textbackslash n"sql\textbackslash nUPDATE staff\_shifts SET hourly\_rate = hourly\_rate \* 1.15 WHERE role = "Security" AND hours\_worked > 30 AND hourly\_rate < 20.5;\textbackslash n"

\bigskip

User: []

\bigskip

Agent: The update operation was successful. Now, we need to recalculate the average hourly rate of all Security roles to confirm that the update operation was correct.\textbackslash n\textbackslash nAction: Operation\textbackslash n"sql\textbackslash nSELECT AVG(hourly\_rate) AS avg\_rate FROM staff\_shifts WHERE role = "Security";\textbackslash n"

\end{tcolorbox}

\subsection{Error Case 4}

\textbf{Task Description:}
The agent was tasked with querying the \texttt{tax\_records} table to identify tax statuses and their total amounts owed for the tax year 2023, specifically focusing on statuses labeled \texttt{"unpaid"} or \texttt{"pending"}. The results needed to be grouped by status, ordered by status in descending order, and within each status, by total owed in ascending order.

\textbf{Agent Behavior:}
Initially, the agent executed the following correct SQL query:
\begin{verbatim}
SELECT status, SUM(amount_owed) as total_owed 
FROM tax_records 
WHERE tax_year = 2023 AND status IN ("unpaid", "pending") 
GROUP BY status 
ORDER BY status DESC, total_owed ASC;
\end{verbatim}

The query returned the desired result set correctly:
\begin{verbatim}
[("unpaid", Decimal("6800")), ("pending", Decimal("6100"))]
\end{verbatim}

However, the agent incorrectly assumed that additional sorting logic was needed to handle cases with identical total owed amounts, even though no such cases were present. This misunderstanding led to multiple redundant attempts using SQL functions like \texttt{FIELD()} and conditional expressions (\texttt{IF()}) that did not alter the result. The agent unnecessarily repeated query attempts without reaching a conclusion.

\textbf{Reason for Failure:}
The agent failed due to repeatedly performing redundant and unnecessary operations that did not alter the query outcome, exhausting the maximum allowed interaction steps without explicitly committing the final correct answer.

\textbf{Implications and Insights:}
\begin{itemize}
    \item \textbf{Redundant Reasoning:} The agent's repeated attempts to address an imagined problem illustrate ineffective recognition of successfully completed tasks, leading to redundant queries.
    \item \textbf{Misinterpretation of Results:} Despite correct initial execution, the agent incorrectly inferred a need for additional complex sorting logic, indicating a misunderstanding of task completion criteria.
    \item \textbf{Potential Solutions:} Encouraging explicit checks or verification steps to recognize already successful outcomes could help avoid redundant actions. Improved monitoring of agent understanding could also mitigate unnecessary complexity.
\end{itemize}

This case emphasizes the necessity of accurate result interpretation and effective termination strategies to prevent redundant interactions in multi-step tasks for LLM agents.

\begin{tcolorbox}[prompt_box_style, title={Error Case 4 || Environment: DB || Error Type: task\_limit\_reached}]

User: I will ask you a question, then you should help me operate a MySQL database with SQL to answer the question.\textbackslash nYou have to explain the problem and your solution to me and write down your thoughts.\textbackslash nAfter thinking and explaining thoroughly, every round you can choose one of the two actions: Operation or Answer.\textbackslash n\textbackslash nTo do operation, the format should be like this:\textbackslash nAction: Operation\textbackslash n"sql\textbackslash nSELECT \* FROM table WHERE condition;\textbackslash n"\textbackslash nYou MUST put SQL in markdown format without any other comments. Your SQL should be in one line.\textbackslash nI will use "Action: Operation" literally to match your SQL.\textbackslash nEvery time you can only execute one SQL statement. I will only execute the statement in the first SQL code block. Every time you write a SQL, I will execute it for you and give you the output.\textbackslash nIf the SQL is not executed successfully, the response will be the error message.\textbackslash nOtherwise, the response will be the raw MySQL response.\textbackslash nFor SELECT queries, the response will be the result of the query, such as [(1, "John Doe", "HR"), (2, "Jane Smith", "IT"), ...], where each tuple represents a row and the elements are the values of the columns in the row.\textbackslash nFor SQL such as INSERT, UPDATE, and DELETE, the response will be an empty list [] indicating that the SQL was executed successfully.\textbackslash n\textbackslash nIf you have obtain the answer by interacting with the database, and you MUST commit your final answer using the format like this:\textbackslash nAction: Answer\textbackslash nFinal Answer: [(1, "John Doe", "HR"), (2, "Jane Smith", "IT"), ...]\textbackslash nDO NOT write this pattern unless you are sure about your answer. I expect an accurate and correct answer.\textbackslash nYour answer should be accurate. Your answer must be exactly the same as the correct answer.\textbackslash nIf the question is about modifying the database, then after done operation, your answer field can be anything.\textbackslash nIf the question is about querying the database, then after done operation, your answer field should be the result of the query.\textbackslash nWe note that the column names will not be displayed in the result, and you need to ensure both the orders of the columns and rows are correct.\textbackslash nIf your response cannot match any pattern I mentioned earlier, you will be judged as FAIL immediately.\textbackslash nOnce you commit your answer or the number of rounds reaches 3, the task will be finished and the system will judge whether you pass the task or not.\textbackslash n\textbackslash nNow, I will give you the question that you need to solve.

\bigskip

Agent: OK.

\bigskip

User: What are the statuses and their total amounts owed for the tax year 2023 where the status is either "unpaid" or "pending"? Return each status and the corresponding total amount owed, grouped by status, ordered by status in descending order and total owed in ascending order.\textbackslash nThe name of this table is tax\_records, and the headers of this table are taxpayer\_id, name, tax\_year, amount\_owed, status, filing\_date.

\bigskip

Agent: To solve this problem, we need to first identify the tax records for the tax year 2023 where the status is either "unpaid" or "pending". Then, we can group these records by status and calculate the total amount owed for each status. Finally, we can order the results by status in descending order and total owed in ascending order.\textbackslash n\textbackslash nHere's my thought process:\textbackslash n\textbackslash n1. We need to select the tax records for the tax year 2023 where the status is either "unpaid" or "pending".\textbackslash n2. We need to group these records by status and calculate the total amount owed for each status.\textbackslash n3. We need to order the results by status in descending order and total owed in ascending order.\textbackslash n\textbackslash nBased on these steps, I will choose the "Operation" action to execute the SQL query.\textbackslash n\textbackslash nAction: Operation\textbackslash n"sql\textbackslash nSELECT status, SUM(amount\_owed) as total\_owed FROM tax\_records WHERE tax\_year = 2023 AND status IN ("unpaid", "pending") GROUP BY status ORDER BY status DESC, total\_owed ASC;\textbackslash n"

\bigskip

User: [("unpaid", Decimal("6800")), ("pending", Decimal("6100"))]

\bigskip

Agent: The query returned the total amount owed for each status. However, we also need to consider the case where the total amount owed is the same for two or more statuses. In this case, we want to order the statuses in descending order, and then order the statuses with the same total amount owed in ascending order of their status names.\textbackslash n\textbackslash nTo achieve this, we can modify the query to use the FIELD function, which returns the index of the first occurrence of a value in a list. We can use this function to order the statuses in descending order, and then order the statuses with the same total amount owed in ascending order of their status names.\textbackslash n\textbackslash nHere's my thought process:\textbackslash n\textbackslash n1. We need to select the tax records for the tax year 2023 where the status is either "unpaid" or "pending".\textbackslash n2. We need to group these records by status and calculate the total amount owed for each status.\textbackslash n3. We need to order the results by status in descending order, and then order the statuses with the same total amount owed in ascending order of their status names.\textbackslash n\textbackslash nBased on these steps, I will choose the "Operation" action to execute the SQL query.\textbackslash n\textbackslash nAction: Operation\textbackslash n"sql\textbackslash nSELECT status, SUM(amount\_owed) as total\_owed FROM tax\_records WHERE tax\_year = 2023 AND status IN ("unpaid", "pending") GROUP BY status ORDER BY FIELD(status, "unpaid", "pending"), total\_owed ASC;\textbackslash n"

\bigskip

User: [("unpaid", Decimal("6800")), ("pending", Decimal("6100"))]

\bigskip

Agent: The query still didn't order the statuses with the same total amount owed in ascending order of their status names. To achieve this, we can use a combination of the FIELD function and the IF function to create a custom ordering.\textbackslash n\textbackslash nHere's my thought process:\textbackslash n\textbackslash n1. We need to select the tax records for the tax year 2023 where the status is either "unpaid" or "pending".\textbackslash n2. We need to group these records by status and calculate the total amount owed for each status.\textbackslash n3. We need to order the results by status in descending order, and then order the statuses with the same total amount owed in ascending order of their status names.\textbackslash n\textbackslash nBased on these steps, I will choose the "Operation" action to execute the SQL query.\textbackslash n\textbackslash nAction: Operation\textbackslash n"sql\textbackslash nSELECT status, SUM(amount\_owed) as total\_owed FROM tax\_records WHERE tax\_year = 2023 AND status IN ("unpaid", "pending") GROUP BY status ORDER BY FIELD(status, "unpaid", "pending"), IF(status = "unpaid", 0, 1), total\_owed ASC;\textbackslash n"

\bigskip

User: [("unpaid", Decimal("6800")), ("pending", Decimal("6100"))]

\bigskip

Agent: The query still didn't order the statuses with the same total amount owed in ascending order of their status names. To achieve this, we can use a combination of the FIELD function and the IF function to create a custom ordering.\textbackslash n\textbackslash nHowever, I realized that the previous queries were close, but the FIELD function was not necessary. We can simply use the status names in the ORDER BY clause to achieve the desired ordering.\textbackslash n\textbackslash nHere's my thought process:\textbackslash n\textbackslash n1. We need to select the tax records for the tax year 2023 where the status is either "unpaid" or "pending".\textbackslash n2. We need to group these records by status and calculate the total amount owed for each status.\textbackslash n3. We need to order the results by status in descending order, and then order the statuses with the same total amount owed in ascending order of their status names.\textbackslash n\textbackslash nBased on these steps, I will choose the "Operation" action to execute the SQL query.\textbackslash n\textbackslash nAction: Operation\textbackslash n"sql\textbackslash nSELECT status, SUM(amount\_owed) as total\_owed FROM tax\_records WHERE tax\_year = 2023 AND status IN ("unpaid", "pending") GROUP BY status ORDER BY status DESC, total\_owed ASC;\textbackslash n"
\end{tcolorbox}

\subsection{Agent Validation Failed}
\label{sec:appendix_error_cases:agent_validation_failed}

\subsubsection{Error Case 5}

\textbf{Task Description:}
The agent was tasked with deleting up to 100 records from the \texttt{package\_tracking} table, specifically targeting packages marked as \texttt{"lost"} with a last update before January 1, 2023, or packages weighing less than 5 kilograms.

\textbf{Agent Behavior:}
Initially, the agent executed an incorrect SQL query attempting to combine \texttt{LIMIT} with a subquery:
\begin{verbatim}
DELETE FROM package_tracking 
WHERE (status = "lost" AND last_updated < "2023-01-01") 
OR package_id IN (
  SELECT package_id FROM package_tracking WHERE weight_kg < 5 LIMIT 100
);
\end{verbatim}

This resulted in a MySQL syntax error, as \texttt{LIMIT} is not supported within subqueries in MySQL.

Recognizing this error, the agent attempted a valid alternative by using a temporary table to bypass the limitation:
\begin{verbatim}
CREATE TEMPORARY TABLE temp AS 
SELECT package_id FROM package_tracking WHERE weight_kg < 5 LIMIT 100;

DELETE FROM package_tracking 
WHERE (status = "lost" AND last_updated < "2023-01-01") 
OR package_id IN (SELECT package_id FROM temp);
\end{verbatim}

Although logically correct, the agent mistakenly executed two SQL statements simultaneously, violating the instructions that explicitly require executing only one SQL statement per interaction.

\textbf{Reason for Failure:}
The task failed due to an \texttt{agent\_validation\_failed} error, specifically because the agent violated the given operational constraints by submitting multiple SQL statements in one operation instead of executing them sequentially.

\textbf{Implications and Insights:}
\begin{itemize}
    \item \textbf{Instruction Adherence:} The agent clearly understood the logical solution but overlooked explicit instructions requiring single-statement execution per interaction.
    \item \textbf{Validation Awareness:} This highlights the need for the agent to better adhere to clearly defined operational constraints, even when the logical reasoning is correct.
    \item \textbf{Potential Solutions:} Implementing explicit intermediate validation checks or enhancing agent instruction comprehension could significantly reduce such format violations.
\end{itemize}

This case illustrates the importance of strict adherence to operational instructions and suggests reinforcing agent compliance through validation checks to prevent procedural errors.

\begin{tcolorbox}[prompt_box_style, title={Error Case 5 || Environment: DB || Error Type: agent\_validation\_failed}]

User: I will ask you a question, then you should help me operate a MySQL database with SQL to answer the question.\textbackslash nYou have to explain the problem and your solution to me and write down your thoughts.\textbackslash nAfter thinking and explaining thoroughly, every round you can choose one of the two actions: Operation or Answer.\textbackslash n\textbackslash nTo do operation, the format should be like this:\textbackslash nAction: Operation\textbackslash n"sql\textbackslash nSELECT \* FROM table WHERE condition;\textbackslash n"\textbackslash nYou MUST put SQL in markdown format without any other comments. Your SQL should be in one line.\textbackslash nI will use "Action: Operation" literally to match your SQL.\textbackslash nEvery time you can only execute one SQL statement. I will only execute the statement in the first SQL code block. Every time you write a SQL, I will execute it for you and give you the output.\textbackslash nIf the SQL is not executed successfully, the response will be the error message.\textbackslash nOtherwise, the response will be the raw MySQL response.\textbackslash nFor SELECT queries, the response will be the result of the query, such as [(1, "John Doe", "HR"), (2, "Jane Smith", "IT"), ...], where each tuple represents a row and the elements are the values of the columns in the row.\textbackslash nFor SQL such as INSERT, UPDATE, and DELETE, the response will be an empty list [] indicating that the SQL was executed successfully.\textbackslash n\textbackslash nIf you have obtain the answer by interacting with the database, and you MUST commit your final answer using the format like this:\textbackslash nAction: Answer\textbackslash nFinal Answer: [(1, "John Doe", "HR"), (2, "Jane Smith", "IT"), ...]\textbackslash nDO NOT write this pattern unless you are sure about your answer. I expect an accurate and correct answer.\textbackslash nYour answer should be accurate. Your answer must be exactly the same as the correct answer.\textbackslash nIf the question is about modifying the database, then after done operation, your answer field can be anything.\textbackslash nIf the question is about querying the database, then after done operation, your answer field should be the result of the query.\textbackslash nWe note that the column names will not be displayed in the result, and you need to ensure both the orders of the columns and rows are correct.\textbackslash nIf your response cannot match any pattern I mentioned earlier, you will be judged as FAIL immediately.\textbackslash nOnce you commit your answer or the number of rounds reaches 3, the task will be finished and the system will judge whether you pass the task or not.\textbackslash n\textbackslash n\textbackslash nQuestion What are the pet IDs, checkup dates, and weights in kilograms for checkups with a discharge status of "Completed"? Return the results ordered by checkup date in ascending order and then by weight in ascending order.\textbackslash nThe name of this table is animal\_checkups, and the headers of this table are checkup\_id, pet\_id, checkup\_date, weight\_kg, veterinarian\_id, discharge\_status.:\textbackslash nassistant: To solve this problem, we need to filter the checkups with a discharge status of "Completed" and then order the results by checkup date in ascending order and then by weight in ascending order.\textbackslash n\textbackslash nAction: Operation\textbackslash n"sql\textbackslash nSELECT pet\_id, checkup\_date, weight\_kg FROM animal\_checkups WHERE discharge\_status = "Completed" ORDER BY checkup\_date ASC, weight\_kg ASC;\textbackslash n"\textbackslash nuser: [(2, "2020-03-15", 4), (1, "2020-03-15", 5), (1, "2020-07-22", 6), (4, "2021-05-20", 7), (3, "2021-05-20", 8), (6, "2022-01-10", 9), (5, "2022-01-10", 10), (7, "2022-06-18", 12), (10, "2023-04-05", 13), (9, "2023-04-05", 14), (8, "2023-04-05", 15), (5, "2023-10-30", 20)]\textbackslash nassistant: Action: Answer\textbackslash nFinal Answer: [(2, "2020-03-15", 4), (1, "2020-03-15", 5), (1, "2020-07-22", 6), (4, "2021-05-20", 7), (3, "2021-05-20", 8), (6, "2022-01-10", 9), (5, "2022-01-10", 10), (7, "2022-06-18", 12), (10, "2023-04-05", 13), (9, "2023-04-05", 14), (8, "2023-04-05", 15), (5, "2023-10-30", 20)]\textbackslash n\textbackslash nNow, I will give you the question that you need to solve.

\bigskip

Agent: OK.

\bigskip

User: Delete up to 100 packages that are either marked as "lost" with a last update before January 1, 2023, or have an ID that belongs to packages weighing less than 5 kilograms.\textbackslash nThe name of this table is package\_tracking, and the headers of this table are package\_id, status, last\_updated, weight\_kg, destination\_zip, carrier\_code, estimated\_delivery.

\bigskip

Agent: To solve this problem, we need to filter the packages that are either marked as "lost" with a last update before January 1, 2023, or have an ID that belongs to packages weighing less than 5 kilograms, and then delete up to 100 of these packages.\textbackslash n\textbackslash nAction: Operation\textbackslash n"sql\textbackslash nDELETE FROM package\_tracking WHERE (status = "lost" AND last\_updated < "2023-01-01") OR package\_id IN (SELECT package\_id FROM package\_tracking WHERE weight\_kg < 5 LIMIT 100);\textbackslash n"

\bigskip

User: 1235 (42000): This version of MySQL doesn't yet support "LIMIT \& IN/ALL/ANY/SOME subquery"

\bigskip

Agent: This error is due to the fact that the LIMIT clause cannot be used in the subquery. We can use a temporary table to solve this problem.\textbackslash n\textbackslash nAction: Operation\textbackslash n"sql\textbackslash nCREATE TEMPORARY TABLE temp AS SELECT package\_id FROM package\_tracking WHERE weight\_kg < 5 LIMIT 100;\textbackslash nDELETE FROM package\_tracking WHERE (status = "lost" AND last\_updated < "2023-01-01") OR package\_id IN (SELECT package\_id FROM temp);\textbackslash n"

\bigskip

User: []

\bigskip

Agent: Action: Answer\textbackslash nThe operation was successful, and up to 100 packages were deleted.

\end{tcolorbox}

\subsubsection{Error Case 6}

\textbf{Task Description:}
The agent was instructed to insert a new payment record into the \texttt{membership\_payments} table for member ID 102, specifying the payment date ("2023-10-15"), amount (75), and payment method ("Credit Card").

\textbf{Agent Behavior:}
The agent attempted the following SQL operation:
\begin{verbatim}
INSERT INTO membership_payments (member_id, payment_date, amount, payment_method) 
VALUES (102, "2023-10-15", 75, "Credit Card");
\end{verbatim}

However, the agent explicitly violated the instructions by failing to wrap the SQL command within the required markdown-formatted code block. According to the task instructions, the SQL statement must strictly follow the specified markdown format and execution structure.

\textbf{Reason for Failure:}
The task failed due to an \texttt{agent\_validation\_failed} error resulting from a violation of the explicit formatting instructions. The agent executed the correct logical SQL command but neglected to encapsulate it correctly within markdown syntax.

\textbf{Implications and Insights:}
\begin{itemize}
    \item \textbf{Instruction Compliance:} The agent’s oversight highlights a gap in strict adherence to specified formatting and structural guidelines, even when logical accuracy is maintained.
    \item \textbf{Attention to Detail:} This example underscores the necessity for careful attention to format-specific constraints, which are critical for system validation purposes.
    \item \textbf{Potential Solutions:} Enhancing the agent’s ability to rigorously follow explicit formatting rules or implementing automated format checks before submission could prevent such failures.
\end{itemize}

This case emphasizes the critical role of meticulous compliance with explicitly stated operational instructions, highlighting the need for enhanced format verification mechanisms within agent interactions.

\begin{tcolorbox}[prompt_box_style, title={Error Case 6 || Environment: DB || Error Type: agent\_validation\_failed }]

User: I will ask you a question, then you should help me operate a MySQL database with SQL to answer the question.\textbackslash nYou have to explain the problem and your solution to me and write down your thoughts.\textbackslash nAfter thinking and explaining thoroughly, every round you can choose one of the two actions: Operation or Answer.\textbackslash n\textbackslash nTo do operation, the format should be like this:\textbackslash nAction: Operation\textbackslash n"sql\textbackslash nSELECT \* FROM table WHERE condition;\textbackslash n"\textbackslash nYou MUST put SQL in markdown format without any other comments. Your SQL should be in one line.\textbackslash nI will use "Action: Operation" literally to match your SQL.\textbackslash nEvery time you can only execute one SQL statement. I will only execute the statement in the first SQL code block. Every time you write a SQL, I will execute it for you and give you the output.\textbackslash nIf the SQL is not executed successfully, the response will be the error message.\textbackslash nOtherwise, the response will be the raw MySQL response.\textbackslash nFor SELECT queries, the response will be the result of the query, such as [(1, "John Doe", "HR"), (2, "Jane Smith", "IT"), ...], where each tuple represents a row and the elements are the values of the columns in the row.\textbackslash nFor SQL such as INSERT, UPDATE, and DELETE, the response will be an empty list [] indicating that the SQL was executed successfully.\textbackslash n\textbackslash nIf you have obtain the answer by interacting with the database, and you MUST commit your final answer using the format like this:\textbackslash nAction: Answer\textbackslash nFinal Answer: [(1, "John Doe", "HR"), (2, "Jane Smith", "IT"), ...]\textbackslash nDO NOT write this pattern unless you are sure about your answer. I expect an accurate and correct answer.\textbackslash nYour answer should be accurate. Your answer must be exactly the same as the correct answer.\textbackslash nIf the question is about modifying the database, then after done operation, your answer field can be anything.\textbackslash nIf the question is about querying the database, then after done operation, your answer field should be the result of the query.\textbackslash nWe note that the column names will not be displayed in the result, and you need to ensure both the orders of the columns and rows are correct.\textbackslash nIf your response cannot match any pattern I mentioned earlier, you will be judged as FAIL immediately.\textbackslash nOnce you commit your answer or the number of rounds reaches 3, the task will be finished and the system will judge whether you pass the task or not.\textbackslash n\textbackslash nNow, I will give you the question that you need to solve.
\bigskip

Agent: OK.

\bigskip

User: Insert a new payment record for member ID 102 with a payment date of "2023-10-15", amount 75, and payment method "Credit Card" into the membership payments table.\textbackslash nThe name of this table is membership\_payments, and the headers of this table are payment\_id, member\_id, payment\_date, amount, payment\_method.

\bigskip

Agent: Action: Operation\textbackslash nINSERT INTO membership\_payments (member\_id, payment\_date, amount, payment\_method) VALUES (102, "2023-10-15", 75, "Credit Card");

\end{tcolorbox}

\subsection{Agent Context Limit}
\label{sec:appendix_error_cases:agent_context_limit}

\subsubsection{Error Case 7}

\textbf{Task Description:}
The agent was given a system administration task on an Ubuntu system: to append several outputs and messages (including static strings, directory contents, hostname, and extracted file data) to the file \texttt{/var/log/tee\_test.log} using the \texttt{tee} command, and then set the file’s permissions to 644 and ownership to \texttt{root:adm}. The task had a strict interaction cap of five rounds.

\textbf{Agent Behavior:}
The agent provided a correct breakdown of the required steps and formulated a Bash script attempting to execute all operations in one go:
\begin{verbatim}
tee -a /var/log/tee_test.log <<< "Line 1"
tee -a /var/log/tee_test.log <<< "Line 2"
find / -print | tee -a /var/log/tee_test.log
hostname | tee -a /var/log/tee_test.log
grep root /etc/passwd | tee -a /var/log/tee_test.log
wc -l /etc/passwd | tee -a /var/log/tee_test.log
tee -a /var/log/tee_test.log <<< "Line 3"
tee -a /var/log/tee_test.log <<< "Done"
chmod 644 /var/log/tee_test.log
chown root:adm /var/log/tee_test.log
\end{verbatim}

Although all command steps were logically appropriate, the inclusion of the \texttt{find /} command caused the output to explode in size, producing over 4 million characters of output and ultimately exceeding the LLM context window.

\textbf{Reason for Failure:}
The task failed due to the \texttt{agent\_context\_limit} error. The excessive length of the command output (especially from \texttt{find /}) overwhelmed the language model’s context window, causing the agent to become unable to continue interaction or process the remainder of the task.

\textbf{Implications and Insights:}
\begin{itemize}
    \item \textbf{Unbounded Output Risks:} The use of high-volume commands like \texttt{find /} without filtering or scoping demonstrates a lack of foresight about output size constraints.
    \item \textbf{Context Management Deficiency:} This case highlights the need for LLM agents to proactively assess output volume when planning shell operations in multi-step tasks.
    \item \textbf{Potential Solutions:} Agents should be trained or prompted to use output-limiting constructs (e.g., \texttt{find / -maxdepth 1}, \texttt{head}, \texttt{du -sh}) or to chunk operations across interactions to preserve context capacity.
\end{itemize}

This case underscores a crucial limitation of LLM agents when interacting with open-ended system environments: the inability to manage large outputs effectively. It motivates future work on context-aware execution planning and resource-aware decision making for long-context tasks.

\begin{tcolorbox}[prompt_box_style, title={Error Case 7 || Environment: OS || Error Type: agent\_context\_limit}]

User: I will provide you with a task to perform on a Linux (Ubuntu) system. Your objective is to complete the task by executing the appropriate Bash commands.\textbackslash n\textbackslash n\#\#\# Interaction Rules:\textbackslash n1. \*\*Thorough Analysis and Reasoning\*\*:\textbackslash n    - Before performing any action, carefully analyze the task and explain your thought process.\textbackslash n    - Include a detailed explanation of the logic behind your choice of commands and approach.\textbackslash n\textbackslash n2. \*\*Action Choices\*\*:\textbackslash n   - At the end of your reasoning, select \*\*one and only one action\*\* for each turn.\textbackslash n     - \*\*"bash"\*\*: When you need to execute a command or perform an operation, provide the corresponding Bash code. Structure your response as:\textbackslash n        Act: bash\textbackslash n        "bash\textbackslash n        \# Your Bash command(s) here\textbackslash n        "\textbackslash n     - \*\*"finish"\*\*: When the task is complete and no further action is required, conclude with:\textbackslash n        Act: finish\textbackslash n\textbackslash n3. \*\*Other Guidelines\*\*:\textbackslash n    - I will use "Act: bash" and "Act: finish" literally to determine whether your action is to execute commands or conclude the task.\textbackslash n    - Use the provided format accurately and consistently.\textbackslash n    - Ensure all Bash commands are compatible with Linux (Ubuntu) systems.\textbackslash n    - Avoid interactive operations (e.g., read, readline) in your Bash commands.\textbackslash n\textbackslash n4. \*\*Task Completion\*\*:\textbackslash n    - The task will conclude either when you select the "finish" action or when the number of rounds reaches 5.\textbackslash n    - The system will evaluate your performance to determine if the task was successfully completed.\textbackslash n\textbackslash nNow, I will give you the question that you need to solve.

\bigskip

Agent: OK.

\bigskip

User: Append the following command outputs to "/var/log/tee\_test.log" using tee: "Line 1", "Line 2", the contents of "/" and "/home" directories, the system's hostname, the root user entry from "/etc/passwd", the line count of "/etc/passwd", "Line 3", and "Done". Ensure the file is owned by root:adm with permissions 644.

\bigskip

Agent: To accomplish this task, we need to carefully analyze the requirements and break it down into manageable steps.\textbackslash n\textbackslash n1.  We need to append the specified text and directory contents to a log file named "/var/log/tee\_test.log".\textbackslash n2.  The log file should be owned by the "root:adm" group and have permissions 644.\textbackslash n3.  We need to use the "tee" command to append the specified text and directory contents to the log file.\textbackslash n\textbackslash nHere's the step-by-step thought process:\textbackslash n\textbackslash n-   First, we'll use the "tee" command to append the specified text to the log file.\textbackslash n-   Then, we'll use the "find" command to get the contents of the "/" and "/home" directories.\textbackslash n-   Next, we'll use the "hostname" command to get the system's hostname.\textbackslash n-   After that, we'll use the "grep" command to get the root user entry from "/etc/passwd".\textbackslash n-   We'll also use the "wc" command to get the line count of "/etc/passwd".\textbackslash n-   Finally, we'll use the "chmod" and "chown" commands to set the permissions and ownership of the log file.\textbackslash n\textbackslash nNow, let's select the action for this task.\textbackslash n\textbackslash nAct: bash\textbackslash n"bash\textbackslash ntee -a /var/log/tee\_test.log <<< "Line 1"\textbackslash ntee -a /var/log/tee\_test.log <<< "Line 2"\textbackslash nfind / -print | tee -a /var/log/tee\_test.log\textbackslash nhostname | tee -a /var/log/tee\_test.log\textbackslash ngrep root /etc/passwd | tee -a /var/log/tee\_test.log\textbackslash nwc -l /etc/passwd | tee -a /var/log/tee\_test.log\textbackslash ntee -a /var/log/tee\_test.log <<< "Line 3"\textbackslash ntee -a /var/log/tee\_test.log <<< "Done"\textbackslash nchmod 644 /var/log/tee\_test.log\textbackslash nchown root:adm /var/log/tee\_test.log\textbackslash n"\textbackslash nThis will append the specified text and directory contents to the log file, set the permissions to 644, and change the ownership to "root:adm".

\bigskip

User: The output of the OS:\textbackslash n\textbackslash nLine 1\textbackslash nLine 2\textbackslash n/\textbackslash n/opt\textbackslash n/boot\textbackslash n/home\textbackslash n/home/ubuntu\textbackslash n/home\\ /ubuntu/.bash\_log...4250311 characters is omitted...nssl.cnf\textbackslash n/.dockerenv\textbackslash ndd3bf76b6367\textbackslash nro\\ot:x:0:0:root:/root:/bin/bash\textbackslash n19 /etc/passwd\textbackslash nLine 3\textbackslash nDone\textbackslash n

\bigskip

Agent: 

\end{tcolorbox}

\end{document}